\documentclass{article}
\usepackage[hypertexnames=false]{hyperref}       
\usepackage{arxiv}

\usepackage[utf8]{inputenc} 
\usepackage[T1]{fontenc}    
\usepackage{url}            
\usepackage{booktabs}       
\usepackage{amsfonts}       
\usepackage{nicefrac}       
\usepackage{microtype}      
\usepackage{lipsum}
\usepackage{amsmath}
\usepackage{graphicx}
\usepackage[superscript]{cite}
\usepackage{authblk}
\usepackage[pagewise]{lineno}
\usepackage{soul}
\usepackage{color}

\title{\textbf{Embodying Control in Soft Multistable Grippers from Morphofunctional Co-design}}

\author[1]{Juan C. Osorio}
\author[1]{Jhonatan S. Rincon}
\author[1]{Harith Morgan}
\author[1,*]{Andres F. Arrieta}
\affil[1]{School of Mechanical Engineering, Purdue University, West Lafayette, IN 47907 USA}
\affil[*]{corresponding authors: aarrieta@purdue.edu}

\begin{document}
\maketitle
\begin{abstract} \normalsize
Soft robots are distinguished by their flexibility and adaptability, allowing them to perform nearly impossible tasks for rigid robots. However, controlling their behavior is challenging due to their nonlinear material response and infinite degrees of freedom. A potential solution to these challenges is to discretize the infinite-dimensional configuration space into a finite but sufficiently large number of functional modes with programmed dynamics. We present a strategy for co-designing the desired tasks and morphology of pneumatically actuated soft robots with multiple encoded stable states and dynamic responses. Our approach introduces a general method to capture the soft robots’ response using an energy-based analytical model, the parameters of which are obtained using Recursive Feature Elimination. The resulting lumped-parameter model facilitates inverse co-design of the robot’s morphology and planned tasks by embodying specific dynamics upon actuation. We illustrate our approach’s ability to explore the configuration space by co-designing kinematics with optimized stiffnesses and time responses to obtain robots capable of classifying the size and weight of objects and displaying adaptable locomotion with minimal feedback control. This strategy offers a framework for simplifying the control of soft robots by exploiting the nonlinear mechanics of multistable structures and embodying mechanical intelligence into soft material systems.
\end{abstract}


\vfill

\section{Introduction}

Soft robots are characterized by their ability to interact with their environment, adapt to external stimuli, and protect against external disturbances~\cite{Polygerinos2017,Kim2013,Majidi2014}. The inherent safety of soft robots stems from using low modulus of  constituent materials allowing them to conduct tasks that are nearly impossible to their rigid counterparts~\cite {Rus2015}. The interplay between soft mechanics and controls intrinsic to soft robotics gives rise to innovative solutions for tasks ranging from simple grasping to complex robot locomotion~\cite{Laschi2016,Pfeifer2012,Trivedi2008,Tolley_2025}. However, this interplay also poses challenges complicating their modeling and control. These challenges include their infinite dimensionality, material nonlinearity, and large deformations that most soft robots exhibit~\cite{Rus_control_2023}. As a result, sensory systems~\cite{Truby2018} and complex predictive tools~\cite{Chin2020,Thuruthel2017} are required to implement open-loop control, leading to computationally demanding models to represent the robot's behavior. Embedding sensing and control in the robots' architecture~\cite{Drotman2021} offers a new strategy to address some of the soft robotics' main challenges, utilizing the robot morphology to reduce the use of complex algorithms and models by leveraging their intrinsically rich mechanical behavior. For example, granular material-based universal gripper~\cite{brown_universal_2010} can adapt to and grasp different object shapes without needing closed-loop control based on their mechanical response. Zoe et al.~\cite{zou_retrofit_2024} developed a methodology to use various soft actuators as pneumatic sensors, enabling distance and shape sensing, profile scanning, and stiffness determination. Other approaches focused on the use of mechanical instabilities for sequential programming and fast actuation. Yang et al.~\cite{Yang_2024} explore the sequential programming of a soft gripper to perform grasping and twisting motion driven solely by mechanical instabilities in the gripper's body. Another relevant example by Lue Y. et al.~\cite{Luo_2024_snap} demonstrates the use of snap-through instabilities for fast actuation and reconfiguration. These examples illustrate how the intelligent co-design of the robot's body morphology and functional behavior enables an approach to close the control loop for realizing desired tasks. 

Multistable structures offer an alternative path to achieve control of soft robotics with feedback via the programming of input-specific stable and defined shapes~\cite{Udani2021ProgrammableDomes,Yang2016,Jiang2019}. These types of structures display more than one energetically favorable configuration, enabling a soft system to reach different final shapes from specific actuation inputs. Multistable structures are often the result of assembling multiple classical bistable sub-structures, including constrained beams/trusses~\cite{Restrepo2015,Boston2022}, constrained dielectric elastomers~\cite{Zhao2016}, shells~\cite{Faber2020Dome-PatternedSheets, Udani2021ProgrammableDomes,Yang_grasping_origami_2021,Risso2022}, compliant mechanisms~\cite{Young_Synthesis2009}, or inflatable structures~\cite{Chi2022}. Presently, a wide variety of soft machines leverage mechanical instabilities to improve their design and performance~\cite{Pal2021,Jiang2023,Patel2023,Rothemund2018,Preston2019,van_Laake_2022,Choe2023} by programming different stable configurations and utilizing the rapid energy release during a snap-through instability~\cite{osorio_manta_2023,Gorissen2020,Tang2020,Chen2018,Luo_2024_snap}. Conrad et al.~\cite{conrad_3d-printed_2024} show an electronics-free controller for a complex pneumatic system based on soft logic gates that can be incorporated into different soft machines. Similarly, Peretz et al.~\cite{peretz_underactuated_2020} utilized mechanical instabilities to create fluidic logic and reproduce valving functionality using entirely soft elements. Melancon et al.~\cite{Melancon2022} used a bistable origami pattern to program different target points to its actuator, achieved by a single pressure source and predefined pressure path. Van Raemdonck et al.~\cite{Raemdonck2023} geometrically tuned an actuator to generate complex actuation sequences from a single input by delaying the unit's snap-through. These examples show the versatility of multistable structures in soft robotic systems, as the robot's final configuration can be predicted based solely on its structural response. Consequently, by transforming the infinite-dimensional deformation space into a finite number of stable states, the complexity of predicting the soft robot behavior is reduced, and its feedback can be expressed as a set of available stable configurations. This approach can simplify the complexity of predictive or data-driven algorithms currently needed to design and control soft robots, ultimately making soft robots more accessible and practical. However, the full potential of using multistability and programmed mechanical responses for soft robotics can only be realized by establishing computationally efficient methodologies for their morphological and functional co-design that circumvent the long runtime roadblocks imposed by the predominant use of conventional Finite Element (FE) packages to predict the geometry and response characteristics associated with the desired stable states.

\begin{figure}[!t]
  \centering
  \includegraphics[width=\textwidth]{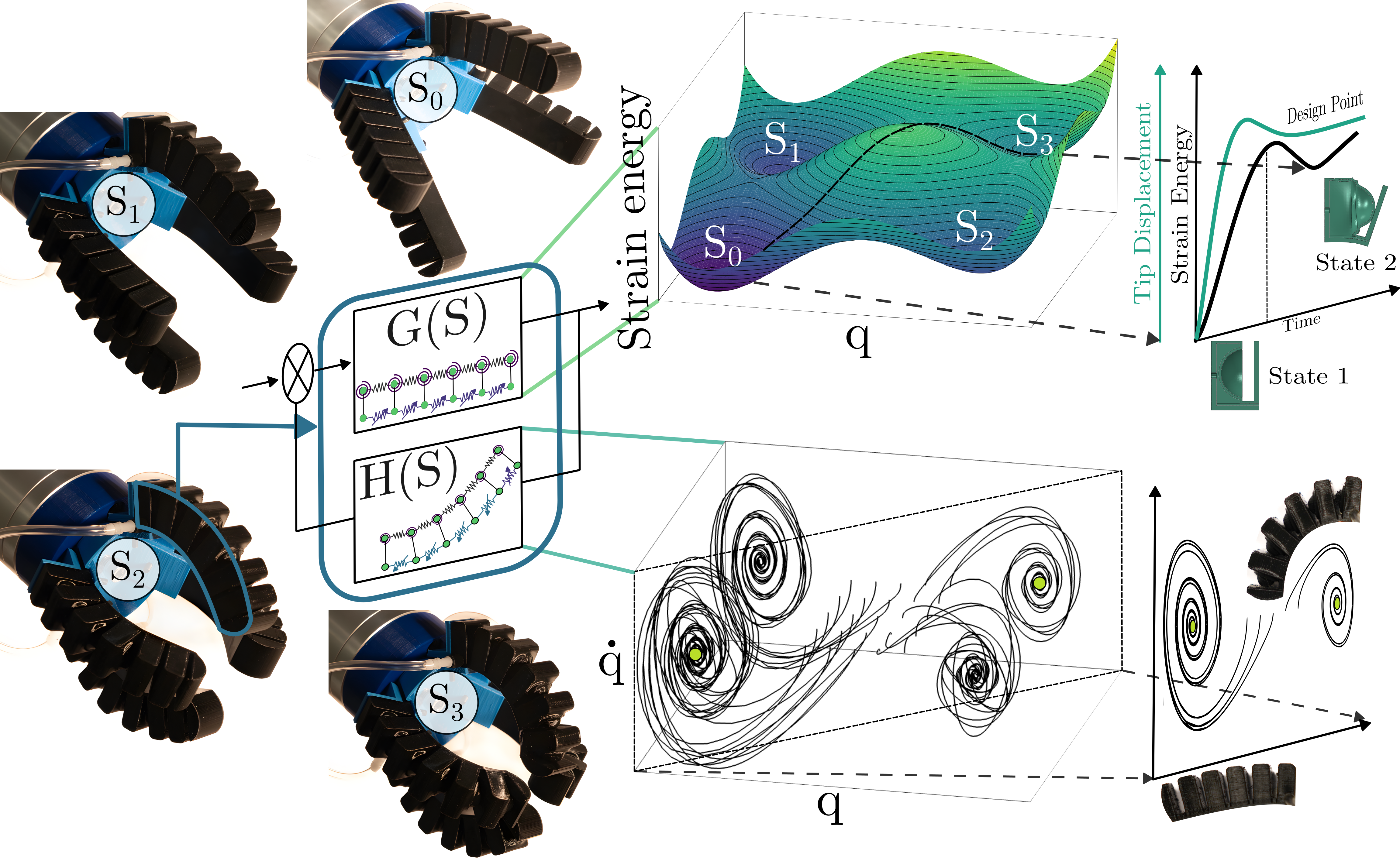}
  \caption{Multistable soft robot with pre-programmed dynamics for positioning and stiffness. Our robot displays four grasping set points ($s_0,s_1,s_3, s_4$) and embodied/morphological control. The robot's morphology and mechanical behavior encode its energy landscape's input and output dynamics (G(s)). Feedback H(s) is derived from the dynamic response of the robot, whereby designed attractors for each stable set point are encoded in the robot geometry. The system's output shows four set points as static minima in the energy landscape and attractors in the phase portrait, where $q$ and $\dot{q}$ are a schematic representation of the robot's generalized coordinates. Set points (energy minima) are accessed by activating/inverting the dome units: State 1 $\rightarrow$ Stress-free state (State 1) and  $\rightarrow$ Inverted state (State 2).}
  \label{fig:Intro_Gripper}
\end{figure}

We introduce a framework for the modeling and inverse design of soft robotics with encoded mechanical behaviors by mapping into a discretized space the desired kinematic configurations leveraging mechanical instabilities (see Figure~\ref{fig:Intro_Gripper}). We achieve this by incorporating different dome-shape programmable units addressable via pneumatic actuation featuring bistable, pseudo-bistable (metastable), and monostable~\cite{Faber2020Dome-PatternedSheets,Brinkmeyer2012} mechanical behaviors (see Figure~\ref{fig:si_dome_phalanx}). As a result, different set points in the form of specific kinematical shapes and time responses are embodied into the system morphology (Figure~\ref{fig:Intro_Gripper}), giving our soft robots the inherent capability of reaching different pre-programmed positions and performing specific dynamic behaviors (i.e., grasping timing and release, and locomotion) under open-loop inputs with almost zero error. This characteristic allows for robust task design of our soft robots, as the bistable units provide intrinsic kinematic control dictated by the embodied mechanical response. The position control inherent to the designed multistable morphology can serve as set points for triggering different control actions, decoupling the actuation from the robot's final configuration (shape). At the same time, specific dynamic responses can be programmed along each set point to trigger desiered behaviors. The discretized response afforded by the proposed topology enables the introduction of an energy-based analytical modeling framework, allowing the efficient shape and structural property prediction of different multistable soft robots that utilize the dome shape unit topology as a building block  (in our case, the Dome Phalanx Finger - DPF) that ultimately enables the inverse co-design of the robot's functionality and control via the morphology. Our modeling approach is based on lumped parameter elements that can capture different mechanical behaviors of the dome unit via mechanics-informed identification, mapping the geometry of our robot to its mechanical response. The collective interaction of these elastic elements dictates the soft robot's configuration, stiffness, and dynamic response. We demonstrate the versatility of our approach by performing the inverse co-design of DPF-based robots targeting specific target positions while maximizing the programmed states' stiffnesses. Finally, we show that desired dynamic behaviors can be encoded into the DPF morphology by leveraging the viscous effects of its constitutive material. The co-designed static and dynamics response is used to illustrate the task programming through an object classification task, pre-programmed pick-and-place tasks, and the controlled locomotion of six-legged soft walker. The presented modeling and inverse co-design approach leverages geometry and the resulting structural response to encode an electronics-free form of mechanical intelligence that simplifies the actuation and control of soft robots with implications on their accessibility, recyclability, security, and cost.

\section{Results}

\subsection{Embodied position and stiffness control}\label{sec:emb_pos_stiff}

\begin{figure}[!t]
  \centering
  \includegraphics[width=\textwidth]{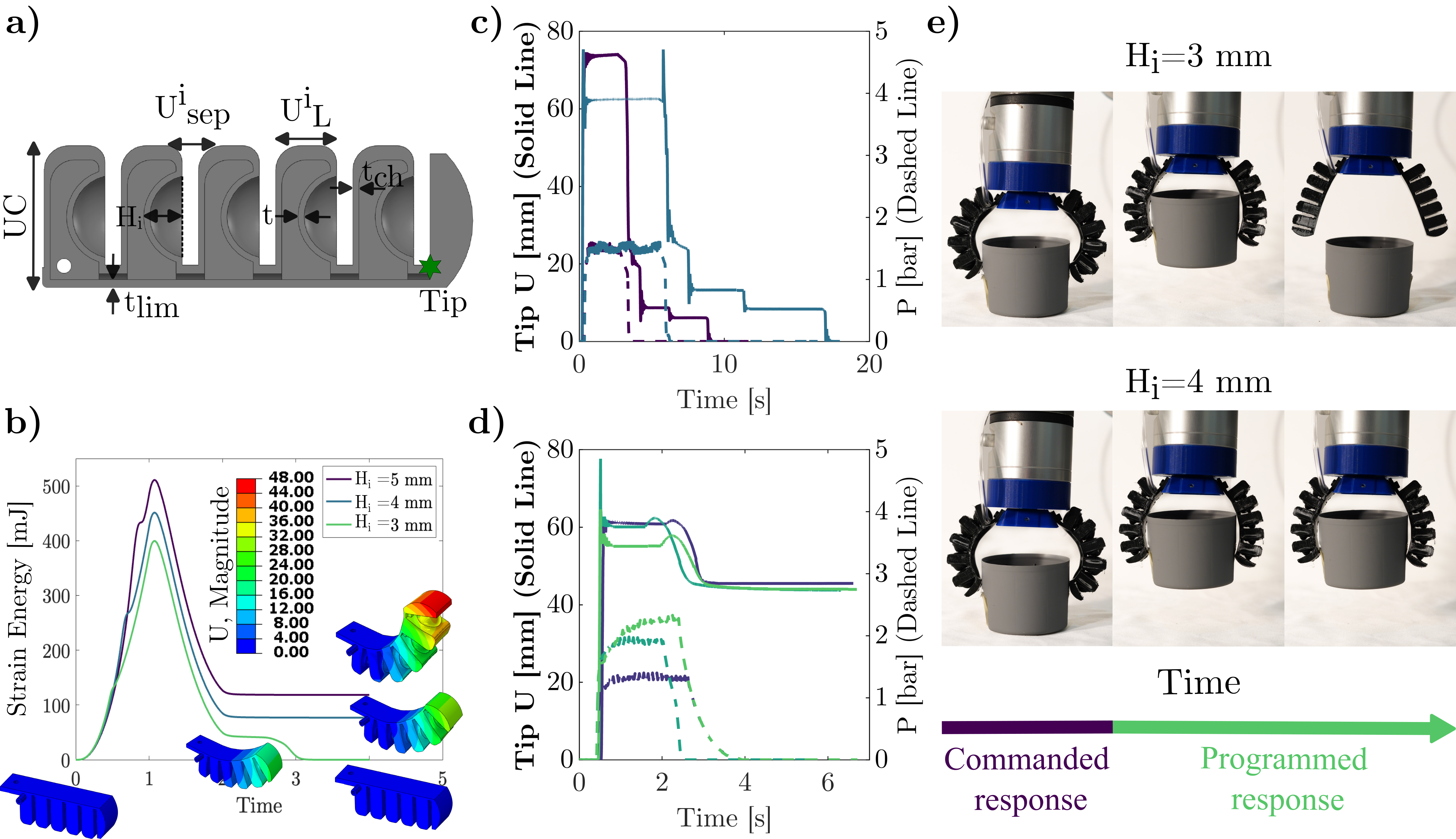}
  \caption{Dome Phalanx Finger (DPF) with different encoded mechanical responses and behaviors. a) 5-segment DPF geometry parameters ($i = 5$). b) Effect of dome height on finger response (Every segment with the same height). $H_i$ = 5 mm and $H_i$  = 4 mm show a bistable behavior, and $H_i$ = 3 mm shows a pseudo-bistable (metastable) response (self-resetting system). c) Metastable finger ($H_i = 3$ mm) Tip displacement and pressure over time. d) Bistable finger ($H_i = 5$ mm) Tip displacement and pressure over time. The same final position is achieved despite different actuation magnitudes (Input pressure). e) Time lapse of two gripper architectures showing both bistable and pseudo-bistable responses.}
  \label{fig:DPG_Geometry}
\end{figure}

We begin by modifying the common PneuNet bending actuator topology~\cite{mosadegh_pneumatic_2014} via the inclusion of dome-shaped units to create our Dome Phalanx Finger (DPF) topology (see Figure~\ref{fig:DPG_Geometry}a). The finger derives its behavior from the mechanical response of the dome-shaped shell elements and the interaction between each segment. As the units invert, the top section of the finger expands, while the limiting layer (see Figure~\ref{fig:si_Unit_Cell}d) retains its original length. This results in global curvature and different final stable shapes, depending on the number of inverted dome units (see Figure~\ref{fig:si_dome_phalanx}b). These dome-shaped units program different mechanical responses and stable shapes into the system, effectively discretizing the infinite-dimensional deformation space into a manageable number of kinematic configurations dependent on the dome unit geometry, each attainable via well-defined, open-loop inputs. The mechanical response and stable configurations of the DPF can be tuned by adjusting the dome height per unit ($H_i$), dome thickness ($t$), number of segments, dome unit size (UC), pneumatic chamber thickness ($t_{\text{ch}}$), limiting layer thickness ($t_{\text{lim}}$), dome unit length  ($U^i_{\text{L}}$), air chamber dimensions ($W_{\text{ch}}$ and $t_{\text{mid}}$), and the spacing between adjacent cells per unit ($U^i_{\text{sep}}$) (see Figure \ref{fig:DPG_Geometry}a and Figure \ref{fig:si_Unit_Cell}a for reference). The geometric parameters of the dome units \( H \) and \( t \) dictate the mechanical response, determining whether it is monostable (see Figure~\ref{fig:si_Unit_Cell}c(i)), metastable (see Figure~\ref{fig:si_Unit_Cell}c(ii) and Movie 1), or bistable (see Figure~\ref{fig:si_Unit_Cell}c(iii)). Furthermore, the transient response of the DPF can be controlled by the material properties and additional geometric parameters ($t_{\text{lim}}$, $t_{\text{ch}}$, $t_{\text{mid}}$), which significantly influence global curvature, system overshoot, and steady-state error (see Figure~\ref{fig:si_dp_param_effect}).

Given the broad range of geometrical configurations, each finger can be geometrically tuned to exhibit different mechanical responses: a bistable response, where the final shape is retained after dome inversion~\cite{Faber2020Dome-PatternedSheets,Udani2021ProgrammableDomes,Osorio2022ProgrammableGrippers} (Figure~\ref{fig:DPG_Geometry}b $H_i$ = 4 mm and $H_i$ = 5 mm). A metastable response, where each unit undergoes snap-through after which its shape is retained for a time $t = \tau$ (after the load is removed) before returning to its stress-free state as a consequence of the combined effects of the unit's viscoelastic material behavior and dome geometry~\cite{osorio_manta_2023,chen_spatiotemporally_2022,liu_effect_2021} (Figure~\ref{fig:DPG_Geometry}b, $H_i$ = 3 mm and Movie 1). Lastly, a monostable response, where the material behavior dominates over the geometric response and the finger responds similarly to a PneuNet bending actuator. As a result of the encoded mechanical responses, the actuation pressure required to invert the dome units is decoupled from the final shape, which is entirely determined by geometric parameters (see Figure \ref{fig:DPG_Geometry}d). Moreover, we leverage the material's viscous response and structural properties to control the reset dynamics of the DPF (see Figure~\ref{fig:DPG_Geometry}c), by controlling the unit metastable behavior (dependent on dome height, material properties) and the duration of the applied load. The resulting freedom to design the DPF's dynamics and set points allows for pre-programming a diverse set of tasks, sequences, and positions.

To showcase these unique capabilities, we developed a two DPF gripper architecture. By changing the dome height, we can program a gripper composed of two fingers to pick and hold an object after actuation (Figure~\ref{fig:DPG_Geometry}e, $H_i$ = 4 mm) or pick an object and release it as the dome units reset (Figure~\ref{fig:DPG_Geometry}e, $H_i$ = 3 mm), the latter of which is ideal for embodying different dynamic response dependent on the robot morphology. The number of stable states that can be encoded using the DPF morphology equals the number of constitutive units where each unit is tunable independently. Consequently, different unit geometries can be combined to achieve multiple stable states, enabling a diversity of behaviors within the same overall finger topology (Bistable + Metastable DPF). This allows for different energy minima and time responses to be embodied within the same finger, which can be accessed by a single pressure input and serve as discrete and functional robot configurations. 

\subsection{Model derivation from mechanics-informed parameter determination}
\label{sec:model_derivation}
\begin{figure}[!t]
  \centering
  \includegraphics[width=\textwidth]{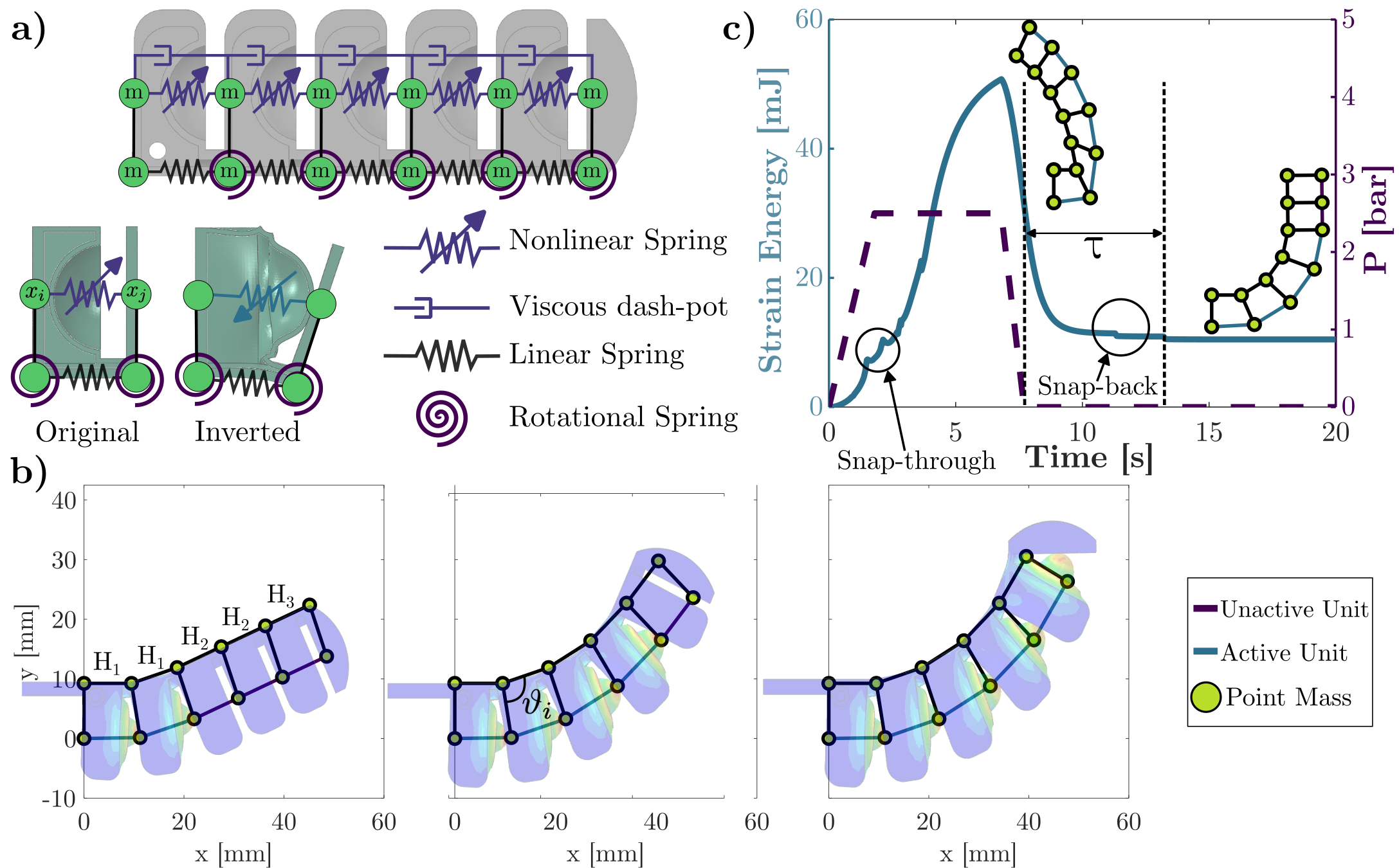}
  \caption{An energy-based model for static and dynamic analysis of multistable soft robots based on the DPF topology. a) Spring lattice model with linear, rotational, and nonlinear springs. b) Static stable states predicted by the lattice model for $H_1$ = $H_2$ = 4 mm, $H_3$ = $H_4$ = 4.5 mm, and $H_5$ = 5 mm. c) Dynamic response for the soft robot with four bistable units and two metastable units ($H_1 = H_2 = H_3 = H_4 = 5$ mm, and $H_5=H_6=3$ mm). The model captures each bistable unit's snap-through and metastable units' snap-back (See Movie 2).}
  \label{fig:DPG_Model}
\end{figure}

A simple yet reliable modeling framework is necessary to design, combine, and predict the finger's behavior based on its geometry. Given the broad design space and the computational cost of using FE models for our finger topology, we propose an energy-based spring model that captures the complex mechanical behaviors needed to design multistable soft robots with desired functional characteristics. Establishing an analytical model to capture the global behavior of the DPF requires consideration of both the unit cells' and the global geometric parameters. By capturing the contribution of each constitutive unit of the DPF and their interactions, we can construct an energy landscape with a simplified model, where each of the minima corresponds to a stable state and a final position of the system. In this way, the system's state ($x_{i}$) can be extracted from the resulting strain energy potential via a minimization process, whereby the DPF's geometrical and stiffness characteristics dictate the programmed stable shapes. The DPF geometry is represented by a lattice array comprising nonlinear, linear, and torsional springs for the static response and point masses and dash-pots for the dynamic response (Figure \ref{fig:DPG_Model}a). The energy for the springs used in the lattice can be written as follows \cite{Meaud2020}:

\begin{align}
&\text{\textbf{Nonlinear:} }E_{NL}(x_{ij})=\frac{1}{2} k_b \left(||x_{ij}||-s_{ij}\right)^2 \left(1+(1-\alpha)\left(\left(\frac{\left(||x_{ij}||-s_{ij}\right)}{d}\right)^2 - 2\frac{\left(||x_{ij}||-s_{ij}\right)}{d}\right)\right)\label{eq:Bistable_Spring}\\ 
&\text{\textbf{Linear:}  }E_{L}(x_{ij})=\frac{1}{2} k_{l}\left(||x_{ij}||-s_{ij}\right)^2\label{eq:Linear_Spring}\\
&\text{\textbf{Torsional:}  }E_{T}(\vartheta)=\frac{1}{2} k_{\vartheta}\left(\vartheta - \vartheta_0\right)^2\label{eq:Tor_Spring}\\
&E(x_{ij})=\sum_{i=1}^n E_{L}(x_{ij})+E_{NL}(x_{ij})+E_T(\vartheta).\label{eq:Tot_energy}
\end{align}

where $x_{ij} = x_i - x_j$, $\vartheta$ is the angle between the rigid link and the limiting layer (see Figure~\ref{fig:si_param_behavior}a and \ref{fig:si_static_dynamic}c), $\vartheta_0$ is the initial angle between the rigid link and the limiting layer, and $k_b$, $k_l$, $k_{\vartheta}$ are stiffness parameters that depend on the DPF geometry. The individual stiffness and spring connectivity allow us to capture the limiting layer and dome inversion effects. We use a nonlinear spring to represent different dome mechanics (Figure \ref{fig:si_Unit_Cell}c), which are described by the parameters $k_b$, $\alpha$, and $d$. Linear and rotational springs are utilized to capture the limiting layer of the finger by combining stretching and bending energies (see Figure~\ref{fig:DPG_Model}a). The connection between the limiting layer and the nonlinear springs is modeled as an infinitely rigid connection, given that the strain is negligible compared to the rest of the lattice elements. By minimizing the sum of all contributions to the system (Equation~\ref{eq:Tot_energy}), the static problem is solved, and all possible finger stable configurations are predicted (see Figure \ref{fig:DPG_Model}b).

A primary challenge of spring-lattice models is to accurately map the geometric parameters of the system to its stiffness constants. To address this challenge, we integrate FE simulations (see section \ref{sec:si_FE_simulation}) with Recursive Feature Elimination (RFE; see section~\ref{sec:model_and_constants})~\cite{guyon_gene_2002} and lasso regression to derive expressions for $k_b$, $\alpha$, and $d$—the nonlinear spring parameters in Equation~\ref{eq:Bistable_Spring}—as functions of dome height ($H$), dome thickness ($t$), dome curvature ($R$) and Young Modulus ($E$) of a generic constitutive unit (see Figure~\ref{fig:si_Unit_Cell}). We considered different dimensionless relations ($\pi_i$) based on the most relevant geometric interactions affecting the dome’s energy~\cite{Faber2020Dome-PatternedSheets}. These mechanics-informed interactions enhance the generality of the analysis by linking the mechanics to the unit’s geometry. For our model, we assume all spring constants can be expressed as functions of the shell’s load-carrying capacity ($\pi_1 = \frac{t}{H}$), curvature-to-thickness ratio ($\pi_2 = \frac{t}{R}$), and dome shallowness ($\pi_3 = \frac{H}{R}$). Consequently, each parameter can be formulated as:

\begin{equation}\label{param_relations}
\mathbf{k_b} = \mathbf{C}(\pi_1,\pi_2,\pi_3)\xi_{k} \hspace{0.5cm} \mathbf{\alpha} = \mathbf{C}(\pi_1,\pi_2,\pi_3)\xi_{\alpha}\hspace{0.5cm} \mathbf{d} = \mathbf{C}(\pi_1,\pi_2,\pi_3)\xi_{d}
\end{equation}

where $\mathbf{C}(\pi_1,\pi_2,\pi_3)$ is a matrix containing all possible candidates and interactions of the non-dimensional relations, as shown in Equation~\ref{eq:c_matrix}. Using RFE, the relevance of each feature can be automatically determined by iteratively removing one feature at a time and observing the model's coefficient of determination ($r^2$). This process yields the weights $\xi_{\alpha}$, $\xi_{k}$, and $\xi_{d}$ with fewer features while maintaining high accuracy. Notably, only the unit cell is used to determine the parameter features; consequently, the DPF's mechanical behavior is derived. A wide range of Young's modulus ($E$), dome height-to-dome base ratio $\left(\frac{H}{r_b}\right)$, dome base radius ($r_b$), and thickness ($t$) are simulated to capture the entire design space of the DPF, including various soft materials ($5 \text{ MPa} \leq E \leq 40$ MPa). As a result, we derive a general equation in terms of the three non-dimensional parameters (Equation~\ref{eq:kb_model}, Equation~\ref{eq:alpha_model}, and Equation~\ref{eq:d_model}, see Section~\ref{sec:parameter_tuning} for details). Using these derived expressions and a gradient-based optimization algorithm (see Section~\ref{sec:static_dynamic_model} for more information), the final shape of the finger can be predicted by minimizing the system's energy, which is now expressed as a function of the geometric parameters of the structure.

A comparison between FE simulations and the proposed model for different stable states of the finger (see Figure~\ref{fig:DPG_Model}b) and different design cases (see Figure~\ref{fig:si_model_validation}) shows good agreement with the energy-based lattice model while also reducing computational time by more than three orders of magnitude (see Table~\ref{tab:sim_time}).

The model is further extended to capture the structure’s dynamic response, allowing us to observe the material’s viscoelastic behavior and analyze metastable cases that energy minimization alone cannot predict. The dynamic response for different unit cell geometries is shown in Figure~\ref{fig:si_static_dynamic}b, and for different five-segmented fingers in Figure~\ref{fig:si_static_dynamic}d, where the dynamic model predicts the same final state of the system after reaching a steady state. To explore the combined behavior of different dome geometries, we modeled a five-segment DPF topology with bistable and metastable units (see Figure~\ref{fig:DPG_Model}c). We stacked four bistable units  ($H_1 = H_2 = H_3 = H_4 = 5$ mm) and two metastable units with the same dome height ($H_5 = H_6 = 3$ mm). This configuration allows us to achieve one unique set point: a stable state and a time-dependent state that holds its shape for a given predictable time $\tau$ (see Figure~\ref{fig:DPG_Model}c). The resulting dynamic response features five different snap-through events and two snap-backs (dome-unit resets), adequately representing the complex dynamics displayed by the DPF (Figure~\ref{fig:DPG_Model}c).

\subsection{Inverse co-design}\label{sec:inv_design}

Based on the model’s accuracy and agreement with finite element (FE) simulations, we formulate an inverse problem to co-design diverse DPF topologies capable of reaching a desired pre-programmed set point and dynamic behavior. Since the system’s stability is governed by its stable states, the optimization algorithm searches only within feasible stable configurations (i.e., the number of active units) and their neighborhoods, resulting in high computational efficiency. To this end, we employ Bayesian optimization~\cite{shahriari_taking_2016} to explore various design parameters, including the number of segments, dome height for each segment ($H_i$), dome thickness ($t$), unit separation for each segment ($\text{U}^i_{\text{sep}}$), and unit length for each segment ($\text{U}^i_{\text{L}}$) (See Figure~\ref{fig:DPG_Geometry}a for parameter definitions and Figure~\ref{fig:si_design_space} for the design space). We consider different objective functions to demonstrate the versatility of the DPF topology in encoding distinct final positions and dynamic responses with minimal actuation. First, we target a specific tip position while simultaneously maximizing stiffness (Position + Stiffness Design; see Section~\ref{sec:geo_opti}), thereby enabling embodied control with maximum stiffness. Second, we exploit the dynamic response of the structure to encode task-specific behaviors directly into the DPF topology, enabling task planning capabilities.

\begin{figure}[!t]
  \centering
  \includegraphics[width=\textwidth]{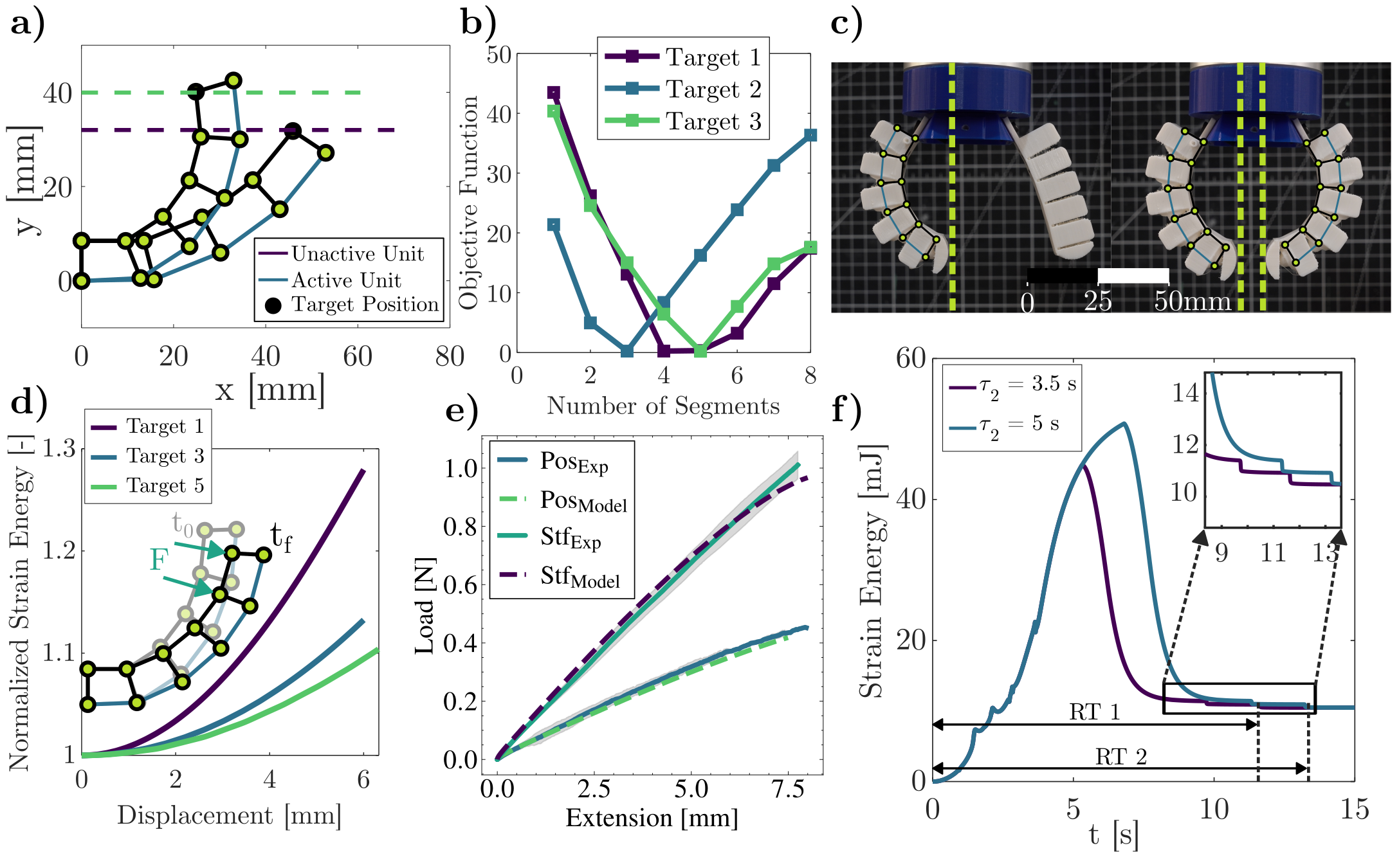}
  \caption{DPF inverse design and experimental validation. (a) Predicted results for Target 1 and 3 from the optimization model (Position Design). (b) Objective function values as a function of the number of segments, with the minimum value identified for each target position. (c) The gripper achieves multiple stable kinematical states predicted by the energy-based model, which are successfully transferred to 3D-printed prototypes. (d) Stiffness determination of the DPF using a follower force approach, with Normalized strain energy vs. displacement curves, is shown for Objectives 1, 3, and 5. (e) Comparison of the proposed model and experimental results for position-only (Pos) optimization and combined position + stiffness optimization (Stf). (f) Dynamic response for a six-section DPF under two different loading profiles.}
  \label{fig:Inv_design}
\end{figure}

\vspace{0.5cm}
\textbf{Position + stiffness design:} 

Given a target tip coordinate [x,y] ($\text{Target}_{xy}$), the inverse problem objective function can be written as:

\begin{align}\label{eq:opti_pos_stiff}
&\min_{H_i,t,\text{Unit}^i_L,\text{Unit}^i_s}w_1\left(\text{Target}_{xy}-\text{Tip}_{\text{ dis}}\right)^2 + w_{2}\left(\frac{1}{dF_{\text{in}}/dx}\right)^2\\
    &\text{s.t.} \quad H_{i+1} \leq H_i,\nonumber,\quad i = 1, \dots, N-1\\
    &\hspace{0.5cm}\quad U^{i+1}_{\text{sep}} = U^{i}_{\text{sep}},\quad i = 1, \dots, N\nonumber\\
    &\hspace{0.5cm}\quad U^{i+1}_{\text{L}} = U^{i}_{\text{L}},\quad i = 1, \dots, N\nonumber
\end{align}

where $\text{Tip}_{\text{ dis}}$ represents the tip point (see Tip in Figure \ref{fig:DPG_Geometry}a), which coincides with the top node of the finger's last unit. $w_1$ and $w_2$ are the weights assigned to each objective, and $dF_{\text{in}}/dx$ is the stiffness obtained from the model (see details in section \ref{sec:SI_Stiff_pos_Inv}). A domain constraint is imposed to guarantee that every dome unit is higher than the one in front ($H_{i+1}\leq H_i$), which in some instances can lead to up to N possible stable states (N being the number of segments). Moreover, we impose constraints for all unit separations and unit lengths to be equal for simplicity and design domain to guarantee bistability on all units, based on the phase diagram shown in Figure \ref{fig:si_Unit_Cell}b. However, different restrictions can be imposed depending on the purpose of the DPF design, and parameters can be individually tuned to achieve a large variety of positions and stable states (see Movie 3). Our algorithm was tested by defining five different tip coordinate targets using two materials with distinct properties (NinjaFlex 85A and Cheetah 95A) optimized to achieve the desired positions. For each case, the target coordinates in \( x \) and \( y \) were provided to the algorithm (Target Position in Figure~\ref{fig:Inv_design}a), and the optimization was performed by incrementally adding one unit per run and calculating the objective function. To reduce the computational cost, a maximum of 8 units was specified (Figure~\ref{fig:Inv_design}b), although additional units can be included if the objective requires a larger finger geometry. The geometry with the minimum objective function value (Obj in Tables~\ref{tab:position_results_Ninja} and \ref{tab:position_results_Cheetah}) was selected and plotted to verify whether the desired position was achieved (see Figure~\ref{fig:Inv_design}a). Geometric parameters, coordinate targets, and objective function values are reported in Table~\ref{tab:position_results_Ninja} and \ref{tab:position_results_Cheetah}. The final stable state and its comparison with each target can be observed in Figure~\ref{fig:si_inverse_position}. To validate our optimization algorithm, we 3D printed the five different geometries shown in Table~\ref{tab:experimental_val} using both tested materials (NinjaFlex 85A and Cheetah 95A) and measured the final tip position after commanded dome inversion. An average error of 7.3\% was measured across all samples, showing good agreement with the optimization algorithm and the lattice model. To evaluate the impact of including stiffness in the objective function, we experimentally tested (see Figure~\ref{fig:SI_Intron}) the five topologies generated via position-only optimization (Equation \ref{eq:opti_pos}) and position + stiffness optimization (Equation \ref{eq:opti_pos_stiff}). Using this co-design strategy, the DPF stiffness has increased on average 1.94 times for all the test cases, achieving maximum stiffnesses of 0.55 N/mm, 0.3 N/mm, and 0.15 N/mm for cases 2, 4, and  5, respectively (see Figure~\ref{fig:Inv_design}e for case 5), which consequently can enhances the achievable grasping force.

The topologies generated by this methodology exhibit deterministic behavior despite using polymeric materials, meaning that kinematic states are consistently attained as long as the gripper's geometric parameters remain unchanged (see Figure~\ref{fig:Inv_design}c and Movie 4). Specifically, we cycle each gripper topology between the initial and fully activated states, verifying the gripper's ability to achieve the same aperture and positioning across multiple cycles (Tests 1-3 in Figure~\ref{fig:Inv_design}c). Furthermore, we assessed material and time degradation effects by performing 300 cycles over three weeks (see Figure~\ref{fig:SI_durability}), which demonstrates that even after numerous activations and an extended period, the gripper's performance remains within 5\% of the initial design behavior. This characteristic enables robust robot design, as the bistable elements provide intrinsic kinematic control dictated by the gripper's mechanical response. It is worth mentioning, that the pressure for activating the units is not controlled using sensors and closed-loop control, illustrating the embodied self-regulation from the designed multistability. Consequently, the position control inherent to the designed multistable structure serves as control set points that can be embodied in the robot's morphology and achieved with a single, open-loop controlled actuator.

\vspace{0.5cm}
\textbf{Dynamic response design:} The unit cell morphology and co-design approach also allow us to encode desired dynamics responses in our soft robots. To this end, we utilized the model described in the previous section and added a term to account for the resetting time. Using the dynamic model (see section \ref{sec:static_dynamic_model}), the metastable units' resetting time is calculated, and the final position of the finger is determined. Given this, specific set points and dynamic behavior in the vicinity of that stable point can be designed. For this particular case, we selected a six-segment DPF with fixed dome heights for the first four units, and we utilized our optimization approach to maximize the resetting time under two different configurations. It is worth mentioning that the rest of the geometric parameters of the DPF ($U_{sep}$, $U_L$, $t_{lim}$) are kept constant during the optimization process, as their effect on the resetting time is negligible. At the same time, the optimization algorithm is utilized to determine the dome heights of the last two units and the remaining geometric parameters ($t$). Given this, the optimization problem can be posed as:

\begin{align}\label{eq:opti_pick_and_place}
&\min_{H_i,t} w_{j}\left(\frac{1}{dF_{\text{in}}/dx}\right)^2 + w_{j+M}\frac{1}{RT_2-RT_1}\\
&\text{s.t.} \quad H_{i} \in G_M,\nonumber,\quad i = 1 \dots, M\\
&\hspace{0.5cm} \quad H_{i+1} = H_i,\nonumber,\quad i = 1, \dots, M-1
\end{align}

where $M$ is the number of metastable units, $RT_2$ and $RT_1$ are the resetting time under two different pressure profiles (see Figure \ref{fig:Inv_design}f and Figure~\ref{fig:si_inverse_meta}a), $w_j$ are the optimization weights, and $G_M$ is the set that contains the metastable configurations. Notice that the last part of the objective function can be substituted by $w_{j+M}\left(\text{Target } RT-RT\right)$ (with $\text{Target } RT$ been the target resetting time) to design for a specific reset time of the units under a pressure profile. Despite setting most of the geometric parameters, we included the stiffness term in our objective function to maximize the grasping capacity for our pick-and-place application indirectly. The full dynamic response predicted by the model shows three different snap-through instability and two snap-backs in the two simulated cases (Figure~\ref{fig:Inv_design} f). As expected, to maximize the resetting time at two different conditions, the algorithm would converge to the highest possible metastable unit (see Table~\ref{tab:pick_place_geo}) for the given thickness (see Figure~\ref{fig:si_Unit_Cell}b).

\subsection{Dome Phalanx Robot (DPR)}
\begin{figure}[!t]
  \centering
  \includegraphics[width=\textwidth]{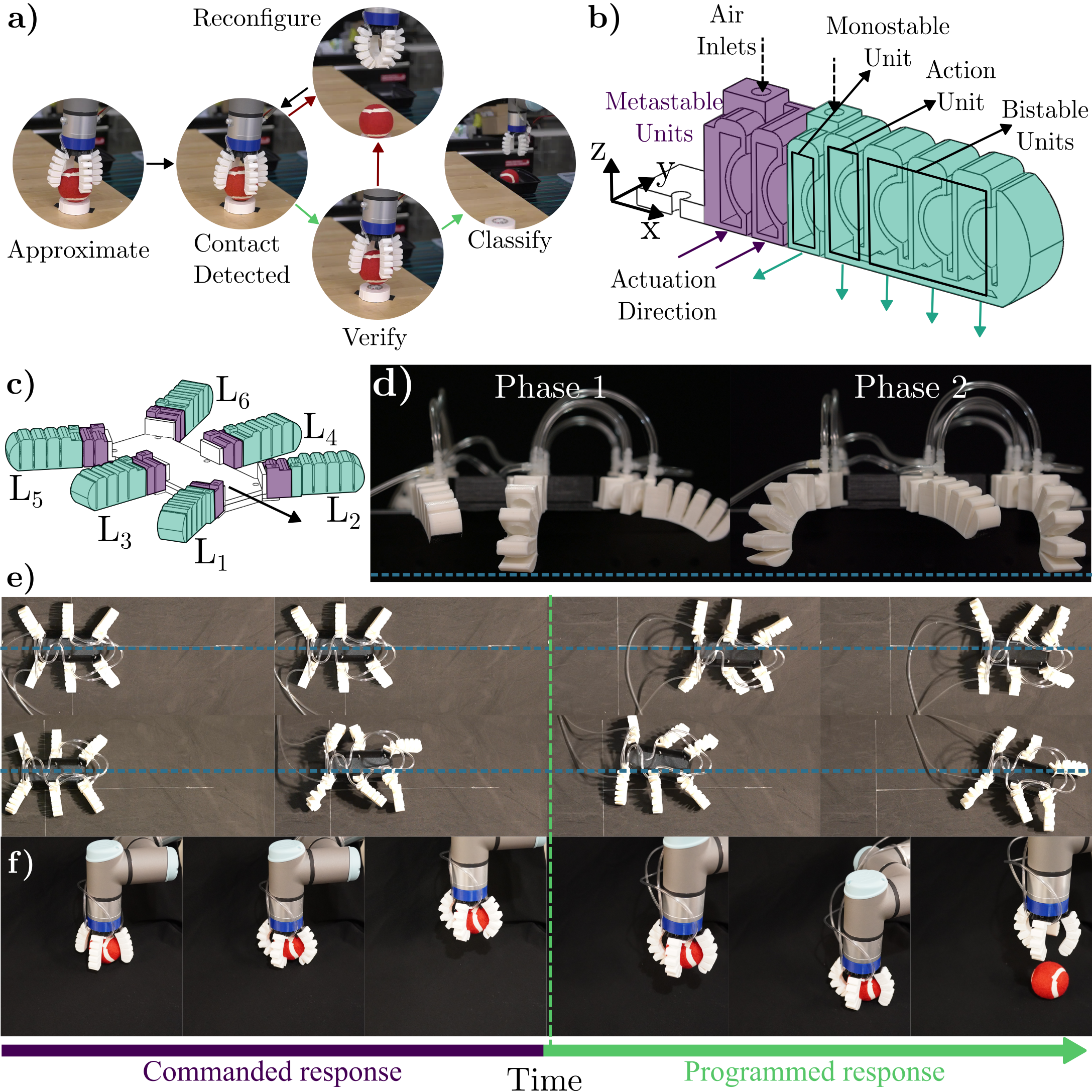}
  \caption{Dome Phalanx Robot (DPR) architectures and applications. a) Classification loop for the DPG. The gripper reconfigures each design's stable state to sort the objects by size and weight (see Movie 7). b) The DPW actuator has two different zones and three bending directions. The action unit is utilized to create turning at the given pressure command. d) DPW architecture connected to create the tripod gate ($L_1$-$L_4$-$L_5$ and $L_2$-$L_3$-$L_6$). Forward units are utilized to create turns. e) Oscillating phases of the DPW with the robot at 30 psi of pressure. f) Snapshots of embodied pick-and-place tasks: A single pressure input is given for the robot to pick the object (Commanded response) and then release it depending on the viscoelastic material response (Programmed response)(see Movie 6). g) Snapshots of the walking trajectory of the DPW. A command is given by increasing the pressure in the first cycle to enable turning in both directions (Programmed response) (see Movie 9).}
  \label{fig:gripper_performance}
\end{figure}

To explore the capabilities of our design approach and the advantages of leveraging multistability in soft robotics, we fabricated two soft robot topologies (Dome Phalanx Robot - DPR). We demonstrate different operation modes using the proposed design methodology and FDM 3D printing to create multiple robot architectures. Specifically, we focus on three types of applications: 1) A multistable soft gripper with four different stable configurations and a metastable configuration that leverages the viscoelastic response of the material to increase the grasping force. This architecture leverages the embodied states so the gripper can perform object size and weight sorting tasks informed by its pre-programmed morphology; 2) A hexapod soft robot that walks with alternating tripod gait~\cite{Tolley_2025} (Dome Phalanx Walker - DPW), by generating horizontal and vertical two-way bending functions combining bistable and metastable units on different directions (see Figure~\ref{fig:gripper_performance}b-d). We impose different actuator geometry and optimization constraints (detailed below) on the model for each operation mode to ensure that each actuator is accurately designed for each specific task. 3) A combination of bistable and metastable dome geometries, where the DPR’s constitutive material viscoelastic response of the material allows us to embody a pick and place robotic behavior into the geometry (see Figure \ref{fig:gripper_performance}f and Movie 6).

\vspace{0.5cm}
\textbf{1) Multistable soft robot:}
 We further leverage the characteristics of our multistable DPF by combining different static and dynamic configurations. We utilized our strategy to design a multistable Dome Phalanx Gripper - DPG with four different stable configurations and a variable dynamic behavior (See Figure ~\ref{fig:si_multi_obj} a). Given this, four different object sizes are encoded into the structure, which allows for object sorting within the range of the target configurations. A metastable unit is incorporated into the tip of each DPF to enhance its classification capabilities and allow the gripper to reconfigure momentarily into a stiffer configuration, allowing it to pick objects of the same size but with different masses. The control loop is closed by combining the morphological information of the robot with a contact sensor in one of the dome units (see section~\ref{sec:Multi_robot}  for details) to detect when the gripper is perturbed from its design stable position. For this application, the robot base length and DPF angle with respect to the base (see Figure~ \ref{fig:si_multi_obj} for reference) are incorporated, and four different apertures are programmed into the gripper morphology by minimizing for four discrete configurations (see more detail in section~\ref{sec:Multi_robot}). Given this, the objective function can be modified as follows:

\begin{align}\label{eq:opti_multi}
&\min_{H_i,t,\text{Unit}^i_L,\text{Unit}^i_s,\theta_B, Base_L}w_j\left(\text{Object Size}_j-\text{Tip}_{\text{dis y j}}\right)^2 + w_{M+N+j+1}\left(\frac{1}{dF_{\text{in}}/dx}\right)^2\\
    &\text{s.t.}\quad H_{i+1} \geq H_i,\nonumber,\quad i = 1, \dots, M-1\\
    &\hspace{0.5cm}\ \quad H_{i} \in G_M,\nonumber,\quad i = M, \dots, N\\
    &\hspace{0.5cm}\quad U^{i+1}_{\text{sep}} = U^{i}_{\text{sep}},\quad i = 1, \dots, N\nonumber\\
    &\hspace{0.5cm}\quad U^{i+1}_{\text{L}} = U^{i}_{\text{L}},\quad i = 1, \dots, N\nonumber
\end{align}

 where $N$ is the number of targeted stable states, $M$ is the number of metastable units,  and $j$ is the number of test objects ($1\leq j \leq M+N$). By doing this, we can optimize for four target positions and maximum stiffness in the fully activated state (four inverted units) and one metastable unit. It is worth noting that for this case, there is a different constraint for the dome height (i.e., $H_{i+1}>H_i$), which allows each of the states to be useful for grasping objects while being accessible in a sequential manner (increasing the input pressure). While the model could be expanded to modify the unit cell's dimensions further and determine the optimal number of units, we fixed the number of units at five to ensure at least one active dome per object size for simplicity. We tested our system by performing an inverse design with four different target positions (see section~\ref{sec:Multi_robot} for reference), using the same methodology described previously (section~\ref{sec:inv_design}). This methodology demonstrates the versatility of our DPF and provides an alternative approach to encoding multiple set points within the finger's body by modulating its unit cell geometry. Using this approach, we embed the classification task directly into the gripper's morphology by leveraging the encoded set points to infer object size. A contact sensor (see Section~\ref{sec:Multi_robot} for details) is integrated into the first unit of one DPF, enabling the detection of perturbations from the gripper's stable configuration. Based on this, we implement a classification loop (Figure~\ref{fig:gripper_performance}a): the gripper approaches the object, evaluates contact sensor feedback, and reconfigures if the measured resistance is below a programmed threshold. Upon contact detection, the object is gently lifted with low pressurization ($P < 5$ psi), and slippage is monitored by measuring the contact sensor a second time. If no slippage occurs, the object is classified into one of four size categories (see Movie 6). If the lift fails, the gripper reattempts by inverting the metastable unit at the tip, increasing stiffness to handle heavier objects of similar size (see Movie 7). This methodology enables simultaneous size and weight classification by combining the system's mechanical response with real-time feedback from a single sensor signal. More importantly, given that the robot's morphology purely drives the classification, there is robustness on the specific task, as it decouples from the actuation pressure.
 
 A key characteristic of our multistable soft robots is their intrinsic capacity to attain a desired shape and exert force without the need from external work, commonly from pressure. This allows our robots to work even after sustaining significant damage, which we induce here by piercing our DPF with several needles (see Movie 7). Notably, our robot maintains its embodied capability to classify objects even under the presence of leaks (illustrated by the oscillating tufts in Movie 9). This test demonstrates the robustness of the embodied control from our approach, as well as illustrating the damage tolerance afforded by the robot's multistability.

\vspace{0.5cm}
\textbf{2) Hexapod walker:} In the second application, we demonstrate the versatility of our dome unit to design a different topology, namely, soft robotic leg for a hexapod walker (see Figure~\ref{fig:gripper_performance}b and \ref{fig:gripper_performance}c). The presented walker utilizes a tripod gait, which means that three legs (two on one side and one on the opposite side, i.e., $L_1$, $L_5$ and $L_4$ in Figure~\ref{fig:gripper_performance}c) move simultaneously forward, while the other three stay in contact with the ground (See Figure~\ref{fig:gripper_performance}d). Each leg is created by rotating the dome unit into different planes to create a leg that moves in two planes (see Figure~\ref{fig:gripper_performance}b), which are connected by using a rigid base (see Figure~\ref{fig:si_walker_1}a). The leg is divided into two zones (highlighted in purple and green in Figure~\ref{fig:gripper_performance}b), which actuate in an alternating sequence to obtain the desired movement. More importantly, we leverage metastable units in the purple region to expand the range of the oscillating cycle on the phase where the legs are not touching the ground (see Figure~\ref{fig:si_walker_1}e and Movie 9 for a comparison between metastable and monostable cases), utilizing the resetting time to delay the reversion of the domes while the other phase is actuated. This amplifies the overall movement, as the leg would touch the ground at a farther point than its monostable counterpart (see Figure~\ref{fig:si_walker_1}e). Furthermore, the green zone is designed to move in the negative y axis (-y) and z-axis, allowing the leg to touch the ground and move backward to generate the oscillating cycle shown in Figure~\ref{fig:si_walker_1}c. The zone contains three different types of units: one monostable unit (which gives the -y movement), three bistable units that provide the negative z axis (-z) movement and can reconfigure to a second stable state, and a action unit that can be either monostable or metastable. The versatility of our robots' units allows us to have two different types of legs, front legs ($L_1$ and $L_2$) which have a metastable action unit to generate asymmetry (see Movie 9) in the movement and be able to change the walking direction (see section~\ref{sec:DPW_Walker} for details), and the middle and back legs, where we select the action units to be monostable as no asymmetry is needed. For the action unit on the front legs, the resetting time is designed to be larger than the one in the purple region, which results in a programmed response that only depends on the input pressure. As the pressure increases in one of the phases, the action unit is inverted (Commanded response in Figure~\ref{fig:gripper_performance}e) inducing asymmetric locomotion in the system (moving the center of mass of the robot), which would stay active as the cycle time of each phase is lower than the resetting time of the action unit (Programmed response Figure~\ref{fig:gripper_performance}e). Note that the robot can recover its forward movement by increasing the pressure on the opposite phase, restoring the symmetry in the system with the two action units inverted. Finally, the bistable units on the green zone can be inverted to morph the robot into an arch-like shape (see Figure~\ref{fig:si_walker_1}b), that increases its payload capacity as it avoids the friction with the ground. As a result, the morphology of the leg embodies a mechanical logic behavior that allows the DPW to move forward, turn in both directions and reconfigure just by controlling the input pressure (see Movie 9).

\vspace{0.5cm}
\textbf{3) Embodied soft robotic tasks:} In the second application, we utilize our ability to program our system's dynamic response. By leveraging our dome structures' viscoelastic response and geometry, we can pre-program the duration for which the structure remains activated before returning to its original state. Combining the meta- and bi-stability of the units enables us to design task planning operations (such as pick-and-place; see Movie 6), where the robot can pick an object by activating all units, move to a specific position, and release the object without additional interventions and with just one pressure input (i.e., eliminating the need for a release command; see Figure~\ref{fig:gripper_performance}e). We tested this operation mode by utilizing the architecture designed in section \ref{sec:inv_design} - Dynamic response design, to create a DPG architecture. The obtained results show how an object can be held while all units are inverted, even when the system releases pressure (Figure \ref{fig:si_pick_place}). However, the object is automatically released from the gripper after the metastable units reset, as desired from the programmed dynamics. Using this approach, we can program a simple pick-and-place operation by activating all domes or perform a simple pinch operation as demonstrated in Figure \ref{fig:si_pick_place} and Movie 6. Further flexibility in the attainable behavior of our robots can be obtained by including the actuation loading time, which results in the same DPG morphology adapting its response to pick an object and release it at two different times as demonstrated in Movie 6.

\section{Conclusion}

In this study, we introduced a modeling framework and inverse design methodology for soft robotics enabling the embodying of their control by discretizing their continuous response using multistable structures. We propose two robotic architectures, The Dome Phalanx Gripper and Dome Phalanx Walker, that leverage mechanical instabilities to discretize the robots’ configuration space into desired kinematics, which serve as control set points. Exploiting these characteristics, we embodied diverse static and dynamic responses, providing an alternative open-loop control strategy for realizing various robotic tasks typically requiring closed-loop approaches guided solely by the designed geometry and material response. Implementing an energy-based modeling approach was central to our methodology, enabling us to accurately predict and program the DPF’s behavior and units. This approach established a robust framework for modeling the finger’s static and dynamic responses with a mechanics-informed parameter discovery procedure that yields high-fidelity results consistent with experimental data. Significantly, this methodology extends beyond the DPF geometry and its lattice representation, as the combination of mechanics knowledge (in this case from shell theory), data, and recursive feature elimination can be further exploited for a wide variety of geometries and reduced-order models. Utilizing the predictive capabilities of our model, we successfully implement the inverse co-design of the DPFs’ morphology and controlled functionality, yielding desired target positions, accessible states, and programmable time responses based on the structure geometry and material viscoelasticity. More importantly, we show that we can perform simple classification tasks by combining the robot morphology and the simple feedback. We demonstrate that multistability enables for damage-tolerance in inflatable robots and shape and dynamics are less reliant on careful pressurization. Furthermore, the versatility of our approach is illustrated by enabling the controlled locomotion and steering of a walker by open-loop simple pressure modulation. As the robot behavior is transformed into a finite, tractable set of positions or set points, the implementation of closed-loop control strategies is simplified by tracking a finite number of states instead of the infinite number inherent to continuous systems. Combining our morphological control strategy with conventional sensors can create a streamlined closed-loop control, where sensors can be utilized to verify if the robot is in a specific state, thus reducing the need for constant feedback from the system. Our results show how these characteristics can be designed to allow for dynamic reconfiguration, grasping force modulation, and embodied logic in the robot morphology without needing closed-loop control. Our findings highlight the potential gains in performance when morphology and functionality are concurrently optimized (i.e., co-designed), a result only feasible via the simplified dynamics and modeling afforded by our approach. The implications of our results are far-reaching, suggesting a promising path toward developing robust soft robots. By encoding the control within the morphology and response of the system, we embody a form of mechanical intelligence offering a reduction in control complexity and forgoing the need for data-hungry models and expensive electronics. The resulting reduction in system complexity paves the way for more efficient, cost-accessible, sustainable, and adaptable soft robotics applications.


\section{Experimental Section}
\textit{Unit cell and DPF Fabrication--} The unit cells are 3D printed using fused deposition modeling (FDM) on a Raise3D Pro2 printer, utilizing Ninjatek Ninjaflex TPU filament \cite{NinjatekNinjaFlexFilament} and Ninjatek Cheetha TPU \cite{NinjatekCheetahFilament}. Each finger is post-processed to correct manufacturing errors and seal the air chambers. For the DPG, we printed a PLA base that secures each of the four fingers (see Movie 5 for assembly).

\vspace{0.5cm}

\textit{Stiffness Measurement tests--} The fingers were loaded onto a specially designed test rig and tested using a universal testing machine (3345 Instron) equipped with a 100N load cell (see Figure \ref{fig:SI_Intron}a). The test rig constrained all degrees of freedom at the finger’s base, with six different positions equally spaced along 90° to ensure that the initial load was always perpendicular to the tested geometry. We measured the force-displacement response five times per sample at a constant velocity of 10 mm/s. Raw data was evaluated and plotted using Python 3.12.

\vspace{0.5cm}

\textit{Finite element analysis--}
The unit cells are modeled in Abaqus using linear elastic material properties and a combination of S3R and S4R shell elements. A dynamic implicit quasi-static approach is utilized to capture the structure’s instabilities. Geometric nonlinear analysis (Nlgeom) is used, and snap-through is triggered using a pressure load on the dome while the edges are pinned. The unit cell is initially modeled in the stress-free state. Simulations were run for different material thicknesses ($t$) and dome heights ($H$). The full geometry of the DPF is also modeled in Abaqus using linear elastic material properties and C3D10 3D elements. Snap-through is triggered using a pressure load on every inner wall air chamber. As discussed in the analysis, a fixed boundary condition is applied to the first unit. After dome inversion, different relaxing steps are used to achieve the final stable state (see section \ref{sec:si_FE_simulation} for details).

\vspace{0.5cm}

\textit{Robot actuation--} Each DPF is actuated using a pressure input controlled by a ball valve. The results displayed in Movies 2 - 3 are produced by applying pressure as slowly as possible to observe the multistable phenomenon.

\vspace{0.5cm}

\textit{Pressure and displacement measurement--} The dynamic behavior of the DPF is captured in detail using a Photron Fastcam Mini UX100 high-speed camera and the data is processed using Tracker Video Analysis and Modeling Tool. Input pressure is measured using a HONEYWELL ABPDANN010BG2A3 pressure sensor coupled with an Arduino Uno (see section \ref{si:P_d_measure} for more details). Each experiment is performed 5 times and the average values are reported. 

\medskip
\textbf{Acknowledgements} \par 
The authors gratefully acknowledge the support from the DARPA Nature as Computer (NAC) program under the guidance of Dr. J. Zhou and NSF-CAREER award No. 1944597. 

\bibliographystyle{is-unsrt}
\bibliography{references}

\begin{thebibliography}{10}
\ifx \showCODEN  \undefined \def \showCODEN #1{CODEN #1}  \fi
\ifx \showISBN   \undefined \def \showISBN  #1{ISBN #1}   \fi
\ifx \showISSN   \undefined \def \showISSN  #1{ISSN #1}   \fi
\ifx \showLCCN   \undefined \def \showLCCN  #1{LCCN #1}   \fi
\ifx \showPRICE  \undefined \def \showPRICE #1{#1}        \fi
\ifx \showURL    \undefined \def \showURL {URL }          \fi
\ifx \path       \undefined \input path.sty               \fi
\ifx \ifshowURL \undefined
     \newif \ifshowURL
     \showURLtrue
\fi

\bibitem{Polygerinos2017}
Panagiotis Polygerinos, Nikolaus Correll, Stephen~A. Morin, Bobak Mosadegh,
  Cagdas~D. Onal, Kirstin Petersen, Matteo Cianchetti, Michael~T. Tolley, and
  Robert~F. Shepherd.
\newblock Soft robotics: Review of fluid-driven intrinsically soft devices;
  manufacturing, sensing, control, and applications in human-robot interaction.
\newblock {\em Advanced Engineering Materials}, 19:\penalty0 1700016, 12 2017.
\newblock \showISSN{1527-2648}.

\bibitem{Kim2013}
Sangbae Kim, Cecilia Laschi, and Barry Trimmer.
\newblock Soft robotics: a bioinspired evolution in robotics.
\newblock {\em Trends in Biotechnology}, 31:\penalty0 287--294, 5 2013.
\newblock \showISSN{0167-7799}.

\bibitem{Majidi2014}
Carmel Majidi.
\newblock Soft robotics: A perspective - current trends and prospects for the
  future.
\newblock {\em Soft Robotics}, 1:\penalty0 5--11, 3 2014.
\newblock \showISSN{21695180}.

\bibitem{Rus2015}
Daniela Rus and Michael~T. Tolley.
\newblock Design, fabrication and control of soft robots.
\newblock {\em Nature 2015 521:7553}, 521:\penalty0 467--475, 5 2015.
\newblock \showISSN{1476-4687}.

\bibitem{Laschi2016}
Cecilia Laschi, Barbara Mazzolai, and Matteo Cianchetti.
\newblock Soft robotics: Technologies and systems pushing the boundaries of
  robot abilities.
\newblock {\em Science Robotics}, 1, 12 2016.
\newblock \showISSN{24709476}.

\bibitem{Pfeifer2012}
Rolf Pfeifer, Max Lungarella, and Fumiya Iida.
\newblock The challenges ahead for bio-inspired 'soft' robotics.
\newblock {\em Communications turbyof the ACM}, 55:\penalty0 76--87, 11 2012.
\newblock \showISSN{00010782}.

\bibitem{Trivedi2008}
Deepak Trivedi, Christopher~D. Rahn, William~M. Kier, and Ian~D. Walker.
\newblock Soft robotics: Biological inspiration, state of the art, and future
  research.
\newblock {\em Applied Bionics and Biomechanics}, 5:\penalty0 99--117, 2008.
\newblock \showISSN{1176-2322}.

\bibitem{Tolley_2025}
Yichen Zhai, Jiayao Yan, Albert De~Boer, Martin Faber, Rohini Gupta, and
  Michael~T. Tolley.
\newblock Monolithic desktop digital fabrication of autonomous walking robots.
\newblock {\em Advanced Intelligent Systems}, n/a\penalty0 (n/a):\penalty0
  2400876, 2025.

\bibitem{Rus_control_2023}
Cosimo~Della Santina, Christian Duriez, and Daniela Rus.
\newblock Model-based control of soft robots: A survey of the state of the art
  and open challenges.
\newblock {\em IEEE Control Systems}, 43:\penalty0 30--65, 6 2023.
\newblock \showISSN{1941000X}.

\bibitem{Truby2018}
Ryan~L. Truby, Michael Wehner, Abigail~K. Grosskopf, Daniel~M. Vogt,
  Sebastien~G.M. Uzel, Robert~J. Wood, and Jennifer~A. Lewis.
\newblock Soft somatosensitive actuators via embedded 3d printing.
\newblock {\em Advanced Materials}, 30:\penalty0 1706383, 4 2018.
\newblock \showISSN{1521-4095}.

\bibitem{Chin2020}
Keene Chin, Tess Hellebrekers, and Carmel Majidi.
\newblock Machine learning for soft robotic sensing and control.
\newblock {\em Advanced Intelligent Systems}, 2:\penalty0 1900171, 6 2020.
\newblock \showISSN{2640-4567}.

\bibitem{Thuruthel2017}
Thomas~George Thuruthel, Egidio Falotico, Mariangela Manti, Andrea Pratesi,
  Matteo Cianchetti, and Cecilia Laschi.
\newblock Learning closed loop kinematic controllers for continuum manipulators
  in unstructured environments.
\newblock {\em Soft robotics}, 4, 9 2017.
\newblock \showISSN{2169-5180}.

\bibitem{Drotman2021}
Dylan Drotman, Saurabh Jadhav, David Sharp, Christian Chan, and Michael~T.
  Tolley.
\newblock Electronics-free pneumatic circuits for controlling soft-legged
  robots.
\newblock {\em Science Robotics}, 6:\penalty0 2627, 2 2021.
\newblock \showISSN{24709476}.

\bibitem{brown_universal_2010}
Eric Brown, Nicholas Rodenberg, John Amend, Annan Mozeika, Erik Steltz,
  Mitchell~R. Zakin, Hod Lipson, and Heinrich~M. Jaeger.
\newblock Universal robotic gripper based on the jamming of granular material.
\newblock {\em Proceedings of the National Academy of Sciences}, 107\penalty0
  (44):\penalty0 18809--18814, November 2010.
\newblock Publisher: Proceedings of the National Academy of Sciences.

\bibitem{zou_retrofit_2024}
Shibo Zou, Sergio Picella, Jelle De~Vries, Vera~G. Kortman, Aim{\'e}e Sakes,
  and Johannes T.~B. Overvelde.
\newblock A retrofit sensing strategy for soft fluidic robots.
\newblock {\em Nat Commun}, 15\penalty0 (1):\penalty0 539, January 2024.
\newblock \showISSN{2041-1723}.

\bibitem{Yang_2024}
Yi~Yang, Helen Read, Mohammed Sbai, Ahmad Zareei, Antonio~Elia Forte, David
  Melancon, and Katia Bertoldi.
\newblock Complex deformation in soft cylindrical structures via programmable
  sequential instabilities.
\newblock {\em Advanced Materials}, 36\penalty0 (46):\penalty0 2406611, 2024.

\bibitem{Luo_2024_snap}
Yichi Luo, Dinesh~K. Patel, Zefang Li, Yafeng Hu, Hao Luo, Lining Yao, and
  Carmel Majidi.
\newblock Intrinsically multistable soft actuator driven by mixed-mode
  snap-through instabilities.
\newblock {\em Advanced Science}, 11\penalty0 (18):\penalty0 2307391, 2024.

\bibitem{Udani2021ProgrammableDomes}
Janav~P. Udani and Andres~F. Arrieta.
\newblock {Programmable mechanical metastructures from locally bistable domes}.
\newblock {\em Extreme Mechanics Letters}, 42:\penalty0 101081, 1 2021.
\newblock \showISSN{2352-4316}.

\bibitem{Yang2016}
Dian Yang, Lihua Jin, Ramses~V. Martinez, Katia Bertoldi, George~M. Whitesides,
  and Zhigang Suo.
\newblock Phase-transforming and switchable metamaterials.
\newblock {\em Extreme Mechanics Letters}, 6:\penalty0 1--9, 3 2016.
\newblock \showISSN{2352-4316}.

\bibitem{Jiang2019}
Yijie Jiang, Lucia~M. Korpas, and Jordan~R. Raney.
\newblock Bifurcation-based embodied logic and autonomous actuation.
\newblock {\em Nature Communications 2019 10:1}, 10:\penalty0 1--10, 1 2019.
\newblock \showISSN{2041-1723}.

\bibitem{Restrepo2015}
David Restrepo, Nilesh~D. Mankame, and Pablo~D. Zavattieri.
\newblock Phase transforming cellular materials.
\newblock {\em Extreme Mechanics Letters}, 4:\penalty0 52--60, 9 2015.
\newblock \showISSN{2352-4316}.

\bibitem{Boston2022}
D.~Matthew Boston, Francis~R. Phillips, Todd~C. Henry, and Andres~F. Arrieta.
\newblock Spanwise wing morphing using multistable cellular metastructures.
\newblock {\em Extreme Mechanics Letters}, 53:\penalty0 101706, 5 2022.
\newblock \showISSN{2352-4316}.

\bibitem{Zhao2016}
Jianwen Zhao, Shu Wang, David McCoul, Zhiguang Xing, Bo~Huang, Liwu Liu, and
  Jinsong Leng.
\newblock Bistable dielectric elastomer minimum energy structures.
\newblock {\em Smart Materials and Structures}, 25, 6 2016.
\newblock \showISSN{1361665X}.

\bibitem{Faber2020Dome-PatternedSheets}
Jakob~A. Faber, Janav~P. Udani, Katherine~S. Riley, André~R. Studart, and
  Andres~F. Arrieta.
\newblock {Dome-Patterned Metamaterial Sheets}.
\newblock {\em Advanced Science}, 7\penalty0 (22), 11 2020.
\newblock \showISSN{21983844}.

\bibitem{Yang_grasping_origami_2021}
Yi~Yang, Katherine Vella, and Douglas~P. Holmes.
\newblock Grasping with kirigami shells.
\newblock {\em Science Robotics}, 6\penalty0 (54):\penalty0 eabd6426, 2021.

\bibitem{Risso2022}
Giada Risso, Maria Sakovsky, and Paolo Ermanni.
\newblock A highly multi-stable meta-structure via anisotropy for large and
  reversible shape transformation.
\newblock {\em Advanced Science}, 9, 9 2022.
\newblock \showISSN{21983844}.

\bibitem{Young_Synthesis2009}
Young~Seok Oh and Sridhar Kota.
\newblock {Synthesis of Multistable Equilibrium Compliant Mechanisms Using
  Combinations of Bistable Mechanisms}.
\newblock {\em Journal of Mechanical Design}, 131\penalty0 (2), 01 2009.
\newblock \showISSN{1050-0472}.
\newblock 021002.

\bibitem{Chi2022}
Yinding Chi, Yanbin Li, Yao Zhao, Yaoye Hong, Yichao Tang, and Jie Yin.
\newblock Bistable and multistable actuators for soft robots: Structures,
  materials, and functionalities.
\newblock {\em Advanced Materials}, 34, 2022.
\newblock \showISSN{15214095}.

\bibitem{Pal2021}
Aniket Pal, Vanessa Restrepo, Debkalpa Goswami, and Ramses~V. Martinez.
\newblock Exploiting mechanical instabilities in soft robotics: Control,
  sensing, and actuation.
\newblock {\em Advanced Materials}, 33:\penalty0 2006939, 5 2021.
\newblock \showISSN{1521-4095}.

\bibitem{Jiang2023}
Yongkang Jiang, Yingtian Li, Ke~Liu, Hongying Zhang, Xin Tong, Diansheng Chen,
  Lei Wang, and Jamie Paik.
\newblock Ultra-tunable bistable structures for universal robotic applications.
\newblock {\em Cell Reports Physical Science}, 4:\penalty0 101365, 5 2023.
\newblock \showISSN{2666-3864}.

\bibitem{Patel2023}
Dinesh~K Patel, Xiaonan Huang, Yichi Luo, Mrunmayi Mungekar, M~Khalid Jawed,
  Lining Yao, Carmel Majidi, D~K Patel, L~Yao, X~Huang, Y~Luo, C~Majidi,
  M~Mungekar, and M~K Jawed.
\newblock Highly dynamic bistable soft actuator for reconfigurable multimodal
  soft robots.
\newblock {\em Advanced Materials Technologies}, 8:\penalty0 2201259, 1 2023.
\newblock \showISSN{2365-709X}.

\bibitem{Rothemund2018}
Philipp Rothemund, Alar Ainla, Lee Belding, Daniel~J. Preston, Sarah Kurihara,
  Zhigang Suo, and George~M. Whitesides.
\newblock A soft, bistable valve for autonomous control of soft actuators.
\newblock {\em Science Robotics}, 3, 3 2018.
\newblock \showISSN{24709476}.

\bibitem{Preston2019}
Daniel~J. Preston, Philipp Rothemund, Haihui~Joy Jiang, Markus~P. Nemitz, Jeff
  Rawson, Zhigang Suo, and George~M. Whitesides.
\newblock Digital logic for soft devices.
\newblock {\em Proceedings of the National Academy of Sciences of the United
  States of America}, 116:\penalty0 7750--7759, 4 2019.
\newblock \showISSN{10916490}.

\bibitem{van_Laake_2022}
Lucas~C. van Laake, Jelle de~Vries, Sevda~Malek Kani, and Johannes~T.B.
  Overvelde.
\newblock A fluidic relaxation oscillator for reprogrammable sequential
  actuation in soft robots.
\newblock {\em Matter}, pages 2898--2917, 9 2022.
\newblock \showISSN{2590-2385}.

\bibitem{Choe2023}
Jun~Kyu Choe, Junsoo Kim, Hyeonseo Song, Joonbum Bae, and Jiyun Kim.
\newblock A soft, self-sensing tensile valve for perceptive soft robots.
\newblock {\em Nature Communications 2023 14:1}, 14:\penalty0 1--10, 7 2023.
\newblock \showISSN{2041-1723}.

\bibitem{osorio_manta_2023}
Juan~C. Osorio, Chelsea Tinsley, Kendal Tinsley, and Andres~F. Arrieta.
\newblock Manta {Ray} inspired multistable soft robot.
\newblock In {\em 2023 {IEEE} {International} {Conference} on {Soft} {Robotics}
  ({RoboSoft})}, pages 1--6. Institute of Electrical and Electronics Engineers
  Inc., April 2023.
\newblock ISSN: 2769-4534.

\bibitem{Gorissen2020}
Benjamin Gorissen, David Melancon, Nikolaos Vasios, Mehdi Torbati, and Katia
  Bertoldi.
\newblock Inflatable soft jumper inspired by shell snapping.
\newblock {\em Science Robotics}, 5:\penalty0 1967, 5 2020.
\newblock \showISSN{24709476}.

\bibitem{Tang2020}
Yichao Tang, Yinding Chi, Jiefeng Sun, Tzu~Hao Huang, Omid~H. Maghsoudi, Andrew
  Spence, Jianguo Zhao, Hao Su, and Jie Yin.
\newblock Leveraging elastic instabilities for amplified performance:
  Spine-inspired high-speed and high-force soft robots.
\newblock {\em Science Advances}, 6, 5 2020.
\newblock \showISSN{23752548}.

\bibitem{Chen2018}
Tian Chen, Osama~R. Bilal, Kristina Shea, and Chiara Daraio.
\newblock Harnessing bistability for directional propulsion of soft, untethered
  robots.
\newblock {\em Proceedings of the National Academy of Sciences of the United
  States of America}, 115:\penalty0 5698--5702, 5 2018.
\newblock \showISSN{10916490}.

\bibitem{conrad_3d-printed_2024}
S.~Conrad, J.~Teichmann, P.~Auth, N.~Knorr, K.~Ulrich, D.~Bellin, T.~Speck, and
  F.~J. Tauber.
\newblock {3D}-printed digital pneumatic logic for the control of soft robotic
  actuators.
\newblock {\em Science Robotics}, 9\penalty0 (86):\penalty0 eadh4060, January
  2024.
\newblock Publisher: American Association for the Advancement of Science.

\bibitem{peretz_underactuated_2020}
Ofek Peretz, Anand~K. Mishra, Robert~F. Shepherd, and Amir~D. Gat.
\newblock Underactuated fluidic control of a continuous multistable membrane.
\newblock {\em Proceedings of the National Academy of Sciences}, 117\penalty0
  (10):\penalty0 5217--5221, March 2020.
\newblock Publisher: Proceedings of the National Academy of Sciences.

\bibitem{Melancon2022}
David Melancon, Antonio~Elia Forte, Leon~M Kamp, Benjamin Gorissen, Katia
  Bertoldi, D~Melancon, L~M Kamp, J~A Paulson, A~E Forte, and K~Bertoldi.
\newblock Inflatable origami: Multimodal deformation via multistability.
\newblock {\em Advanced Functional Materials}, 32:\penalty0 2201891, 8 2022.
\newblock \showISSN{1616-3028}.

\bibitem{Raemdonck2023}
Bert~Van Raemdonck, Edoardo Milana, Michael~De Volder, Dominiek Reynaerts, and
  Benjamin Gorissen.
\newblock Nonlinear inflatable actuators for distributed control in soft
  robots.
\newblock {\em Advanced Materials}, 35:\penalty0 2301487, 9 2023.
\newblock \showISSN{1521-4095}.

\bibitem{Brinkmeyer2012}
A.~Brinkmeyer, M.~Santer, A.~Pirrera, and P.~M. Weaver.
\newblock Pseudo-bistable self-actuated domes for morphing applications.
\newblock {\em International Journal of Solids and Structures}, 49:\penalty0
  1077--1087, 5 2012.
\newblock \showISSN{0020-7683}.

\bibitem{mosadegh_pneumatic_2014}
Bobak Mosadegh, Panagiotis Polygerinos, Christoph Keplinger, Sophia Wennstedt,
  Robert~F. Shepherd, Unmukt Gupta, Jongmin Shim, Katia Bertoldi, Conor~J.
  Walsh, and George~M. Whitesides.
\newblock Pneumatic {Networks} for {Soft} {Robotics} that {Actuate} {Rapidly}.
\newblock {\em Advanced Functional Materials}, 24\penalty0 (15):\penalty0
  2163--2170, 2014.
\newblock \showISSN{1616-3028}.

\bibitem{Osorio2022ProgrammableGrippers}
Juan~C. Osorio, Harith Morgan, and Andres~F. Arrieta.
\newblock {Programmable Multistable Soft Grippers}.
\newblock {\em 2022 IEEE 5th International Conference on Soft Robotics,
  RoboSoft 2022}, pages 525--530, 2022.
\newblock \showISBN{9781665408288}.

\bibitem{chen_spatiotemporally_2022}
Yuzhen Chen, Tianzhen Liu, and Lihua Jin.
\newblock Spatiotemporally {Programmable} {Surfaces} via {Viscoelastic} {Shell}
  {Snapping}.
\newblock {\em Advanced Intelligent Systems}, 4\penalty0 (9):\penalty0 2100270,
  2022.
\newblock \showISSN{2640-4567}.

\bibitem{liu_effect_2021}
Tianzhen Liu, Yuzhen Chen, Liwu Liu, Yanju Liu, Jinsong Leng, and Lihua Jin.
\newblock Effect of imperfections on pseudo-bistability of viscoelastic domes.
\newblock {\em Extreme Mechanics Letters}, 49:\penalty0 101477, November 2021.
\newblock \showISSN{2352-4316}.

\bibitem{Meaud2020}
Julien Meaud.
\newblock Multistable two-dimensional spring-mass lattices with tunable band
  gaps and wave directionality.
\newblock {\em Journal of Sound and Vibration}, 434:\penalty0 44--62, 2018.
\newblock \showISSN{0022-460X}.

\bibitem{guyon_gene_2002}
Isabelle Guyon, Jason Weston, Stephen Barnhill, and Vladimir Vapnik.
\newblock Gene {Selection} for {Cancer} {Classification} using {Support}
  {Vector} {Machines}.
\newblock {\em Machine Learning}, 46\penalty0 (1):\penalty0 389--422, January
  2002.
\newblock \showISSN{1573-0565}.

\bibitem{shahriari_taking_2016}
Bobak Shahriari, Kevin Swersky, Ziyu Wang, Ryan~P. Adams, and Nando de~Freitas.
\newblock Taking the human out of the loop: A review of bayesian optimization.
\newblock {\em Proceedings of the IEEE}, 104\penalty0 (1):\penalty0 148--175,
  2016.

\bibitem{NinjatekNinjaFlexFilament}
{Ninjatek}.
\newblock {NinjaFlex{\textregistered} 3D Printing Filament}.
\newblock \ifshowURL {\showURL
  \path|https://ninjatek.com/support/technical-specs/|}\fi.

\bibitem{NinjatekCheetahFilament}
{Ninjatek}.
\newblock {Cheetah{\textregistered} 3D Printing Filament}.
\newblock \ifshowURL {\showURL
  \path|https://ninjatek.com/support/technical-specs/|}\fi.

\bibitem{visco_1975}
Wilhelm Flügge.
\newblock Viscoelasticity.
\newblock 1975.
\newblock \showISBN{978-3-662-02278-8}.

\bibitem{ANDREJEVIC2022111607}
Jovana Andrejevic and Chris~H. Rycroft.
\newblock Simulation of crumpled sheets via alternating quasistatic and dynamic
  representations.
\newblock {\em Journal of Computational Physics}, 471:\penalty0 111607, 2022.
\newblock \showISSN{0021-9991}.

\bibitem{riley_sensors_IEEE}
Katherine~S. Riley and Andres~F. Arrieta.
\newblock Flexible mechanical sensors with time-dependent, viscoelastic
  responses.
\newblock In {\em 2023 IEEE International Conference on Flexible and Printable
  Sensors and Systems (FLEPS)}, pages 1--4, 2023.

\bibitem{Han1977}
S.~P. Han.
\newblock A globally convergent method for nonlinear programming.
\newblock {\em Journal of Optimization Theory and Applications}, 22:\penalty0
  297--309, 7 1977.
\newblock \showISSN{00223239}.

\bibitem{Seffen_2016}
Keith~A Seffen and Stefano Vidoli.
\newblock Eversion of bistable shells under magnetic actuation: a model of
  nonlinear shapes.
\newblock {\em Smart Materials and Structures}, 25\penalty0 (6):\penalty0
  065010, may 2016.

\bibitem{insect_walking_1999}
Control of walking in the stick insect: From behavior and physiology to
  modeling.
\newblock {\em Autonomous Robots}, 7\penalty0 (3):\penalty0 271--288, 1999.
\newblock \showISBN{1573-7527}.

\end{thebibliography}

\appendix

\renewcommand{\theequation}{S.\arabic{equation}}
\renewcommand{\thefigure}{S\arabic{figure}}
\renewcommand{\thetable}{S\arabic{table}}
\setcounter{figure}{0}
\setcounter{table}{0}

\section{Supplementary Material}\label{supplementary}

\subsection{Geometry}
\subsubsection{Unit Cell}

\begin{figure}[!h]
  \centering
  \includegraphics[width=\textwidth]{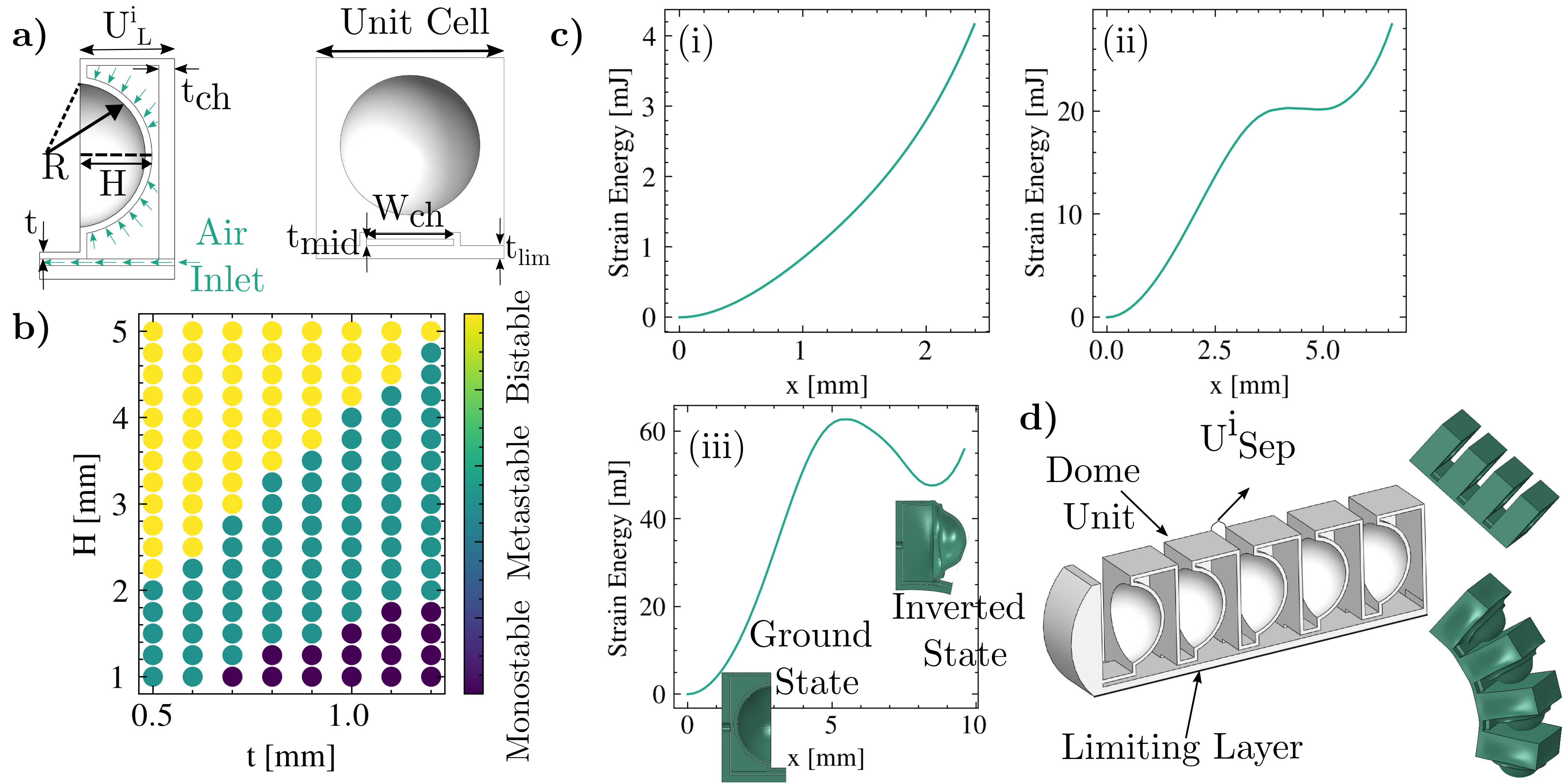}
  \caption{Dome unit geometry and mechanics. a) Unit cell geometry and parameters (Dome unit + air pressure chamber). b) Stability map dependent on dome thickness ($t$) and dome height ($H$) for DPF unit cell. c) Monostable (i), Metastable (ii), and Bistable (iii) dome unit behavior. d) Dome Phalanx finger geometry with two possible, stable states (Initial and fully actuated states).}
  \label{fig:si_Unit_Cell}
\end{figure}

The system's bistable constitutive units are composed of a dome-shaped structure encapsulated by a square chamber that allows for pneumatic actuation and reset by applying positive and negative pressure, respectably (see Figure~\ref{fig:si_Unit_Cell}a). Each dome unit can be geometrically tuned to exhibit a bistable behavior (Figure~\ref{fig:si_Unit_Cell}c (iii))~\cite{Faber2020Dome-PatternedSheets}, pseudo-bistable (metastable - Figure~\ref{fig:si_Unit_Cell}c (ii)) behavior where the is a snap-through instability, but the unit returns to its zero energy state~\cite{chen_spatiotemporally_2022,liu_effect_2021}, and a monostable behavior (Figure~\ref{fig:si_Unit_Cell}c (i)). These mechanical responses can be adjusted by modifying the height $H$ and thickness $t$ of the unit, as shown in Figure~\ref{fig:si_Unit_Cell}b.

\subsubsection{Dome Phalanx Finger (DPF)}
\begin{figure}[!t]
  \centering
  \includegraphics[width=0.9\textwidth]{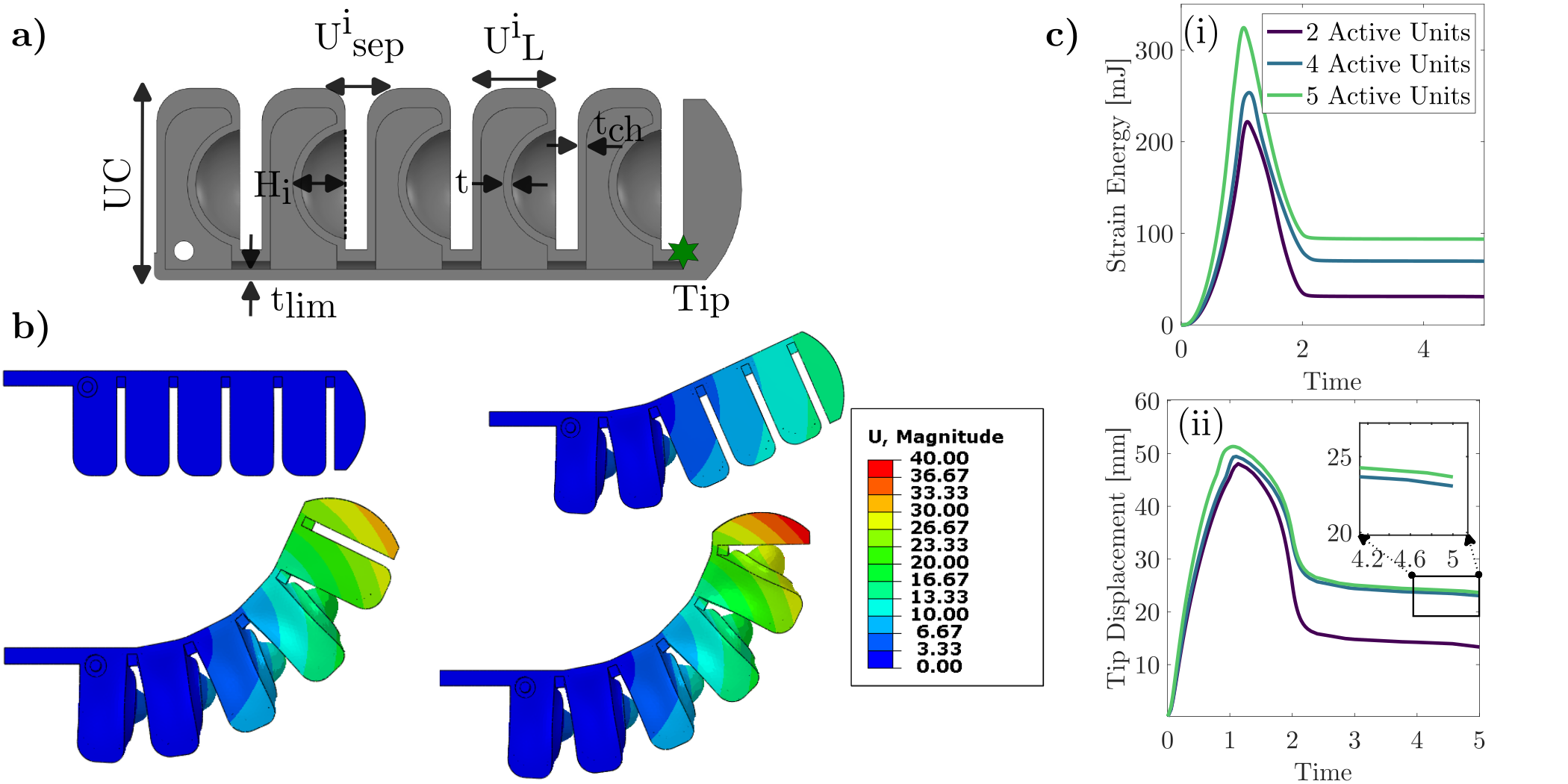}
  \caption{Dome Phalanx Finger geometry and behavior. a) Geometrical parameters that determine the final position and strain energy of the DPF. b) Different stable states of the DPF are encoded by tunning the dome height of each unit cell. c) Strain energy (i) and Tip displacement (ii) as a function of the number of active units. Three different sets of points are encoded into the DPF.}
  \label{fig:si_dome_phalanx}
\end{figure}

The DPF is multistable soft system with sequential dome units (Figure~\ref{fig:si_Unit_Cell}d). The finger derives its multistable behavior from domed-shaped shell elements (Figure~\ref{fig:si_Unit_Cell}a), and it can be geometrically tuned to reach and retain different final shapes after dome inversion.  The DPF performance and stable kinematic configurations can be tuned by changing the dome height ($H$), Unit Cell size (UC), strain limiting thickness ($t_{\text{lim}}$), air channel dimensions ($W_{ch}$ and $t_{\text{mid}}$), spacing between adjacent cells ($\text{U}^i_{\text{sep}}$), unit cell length ($\text{U}^i_{\text{L}}$), chamber thickness ($t_{ch}$) and dome thickness. Once fully inverted, contact between adjacent units and the strain-limiting layer induces the system's global curvature. The domes on each finger segment support programmable deflections as the domes' final positions dictate the global kinematic configuration (see Figure~\ref{fig:si_dome_phalanx}b). The contact between the dome tip and the adjacent unit's chamber provides an additional interaction, contributing to the final curvature after all units are activated and the pressure is removed. 

\begin{figure}[!h]
  \centering
  \includegraphics[width=0.95\textwidth]{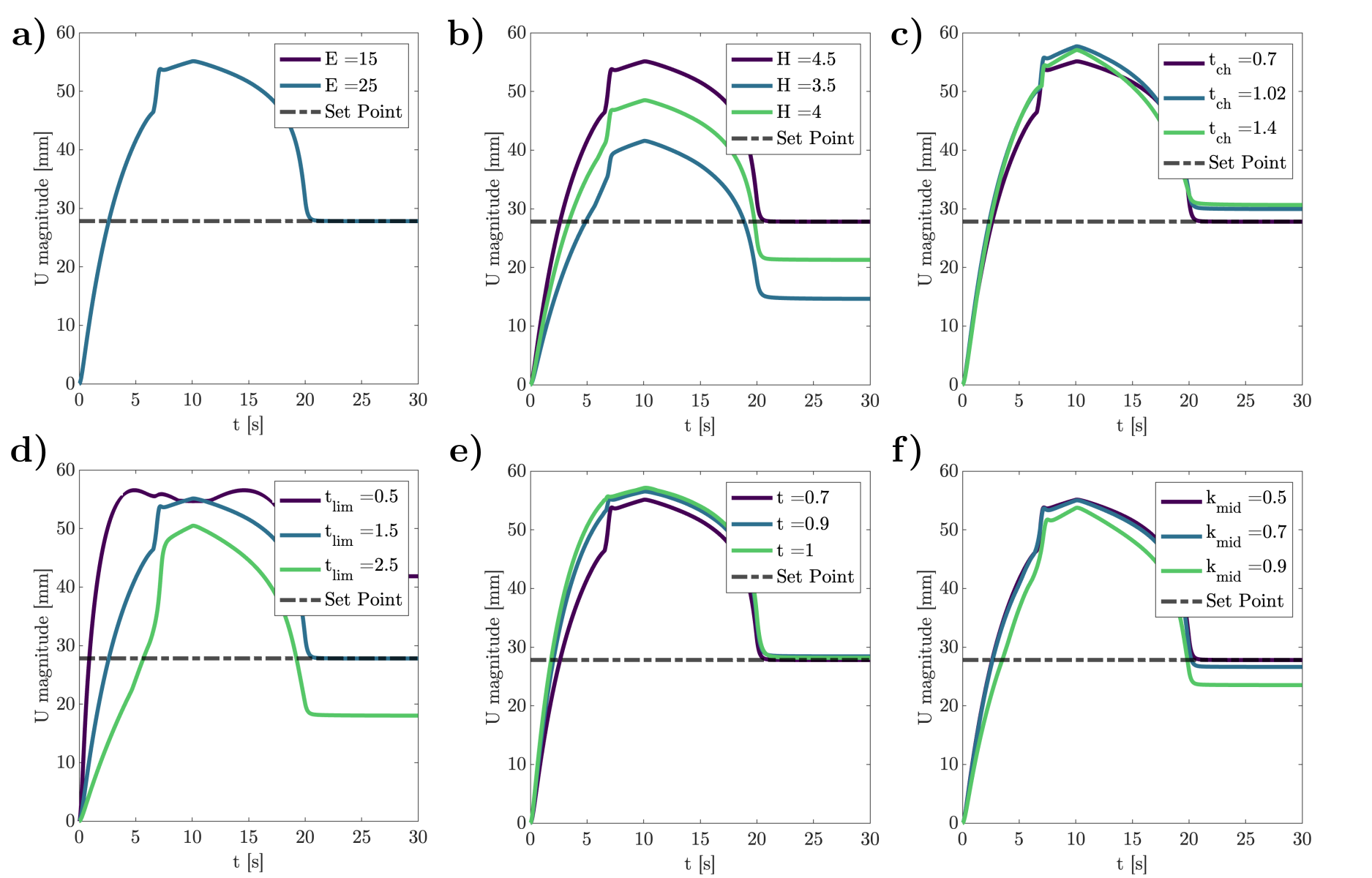}
  \caption{Geometric parameter effect on the time response of DPF. Setpoint, overshoot, and dynamic response can be tuned by combining different parameters. a) Elastic Modulus (E), b) Dome unit Height (H), c) Chamber thickness $t_{ch}$, d) Limiting layer thickness $t_{\text{lim}}$, e) Dome unit thickness ($t$), f) Channel with $W_{\text{ch}} = k_{\text{mid}}UC$}
  \label{fig:si_dp_param_effect}
\end{figure}

The strain energy stored in the gripper topology in its activated states is the summation of the individual contributions from the inverted domes (see Figure~\ref{fig:si_dome_phalanx}c). The magnitude of the interaction between neighboring domes is also affected by dome height ($H$) and chamber thickness ($t_{ch}$). Together, dome height and chamber thickness determine the degree of interaction between neighboring segments after dome inversion and yield the system's global curvature (Figure~\ref{fig:si_dome_phalanx}b) and the final tip position of the system (Figure~\ref{fig:si_dome_phalanx}c). Consequently, various dynamic behaviors can be achieved by adjusting the geometrical parameters, thereby enhancing tunability and control. Specifically, the dome height ($H$) and thickness ($t$), along with the limiting layer thickness ($t_{\text{lim}}$), influence the final stable position or set point. Meanwhile, the chamber thickness ($t_{ch}$) and the air channel dimensions ($W_{ch}$ and $t_{\text{mid}}$) govern the overshoot required to reach a target position (see Figure~\ref{fig:si_dp_param_effect}). This tunability enables the gripper’s response to be tailored by simply modifying the DPF’s geometric characteristics.

\newpage
\subsection{Finite Element Simulations}\label{sec:si_FE_simulation}
Finite element simulations are performed using Abaqus and Python scripting to iterate over different geometrical configurations. Given the dominance of the geometric phenomena in the system, all simulations are done using linear elastic material properties.

\subsubsection{Unit Cell Simulation}
\begin{figure}[!h]
  \centering
  \includegraphics[width=0.9\textwidth]{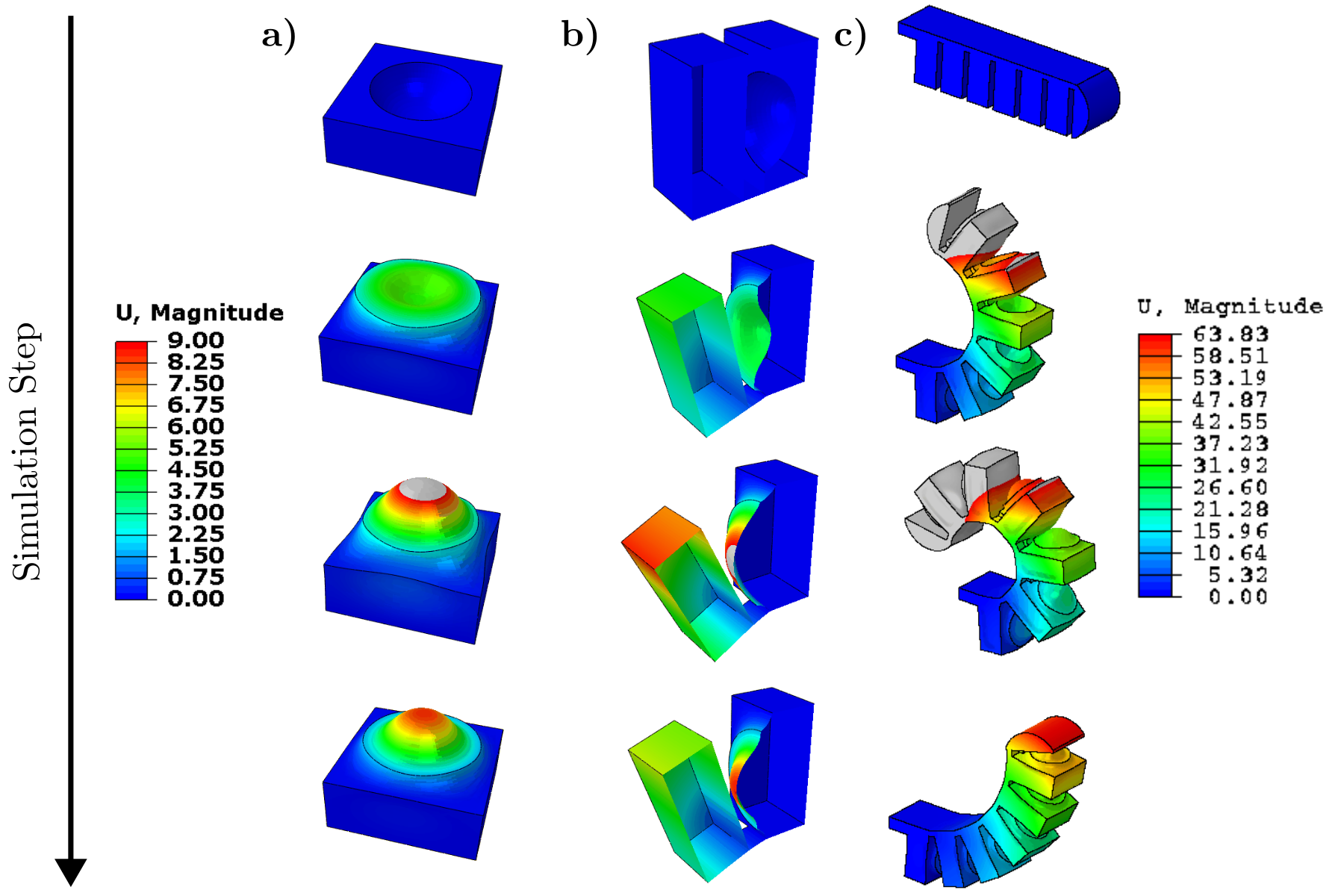}
  \caption{Finite Element (FE) simulations for model parameter tuning and validation. (a) Dome unit inversion sequence used to determine $k_b$ and $\alpha$. (b) Interaction between dome units to determine $d$. (c) 3D simulation for validating the model’s predictions against FE data.}
  \label{fig:si_FE_section}
\end{figure}

The unit cells are modeled using S3R and S4R shell elements. The mesh is structured to capture the dome symmetry and its appropriate post-buckling behavior. A dynamic implicit quasi-static approach is utilized to capture the structure’s instabilities. Geometric nonlinear analysis (Nlgeom) is used, and snap-through is triggered using a displacement control method on the dome while the edges are pinned. The unit cell is initially modeled in the stress-free state. Simulations were run for different material thicknesses ($t$), dome heights ($H$), dome radius ($R_b$), and Elastic Modulus ($E$). Strain energy vs dome tip displacement is extracted for every step of the simulation (see Figure~\ref{fig:si_FE_section}a) which are used to tune the lattice model constant ($k_b$ and $\alpha$ in Equation~\ref{eq:Bistable_Spring}).

\subsubsection{Dome Unit interaction}
The interaction between dome units is modeled by simulating two consecutive units (see Figure~\ref{fig:si_FE_section}b). This approach allows us to assess the influence of one unit on the final position of the adjacent dome tip. The interaction is captured using general contact between the surfaces while varying key parameters such as material thickness ($t$), dome height ($H$), dome radius ($r_b$), and elastic modulus (E). Tip displacement is recorded over time, and the final displacement values, as a function of the test parameters, are used to calibrate the parameter $d$ (Equation~\ref{eq:Bistable_Spring}) in the lattice model.

\subsubsection{DPF 3D Simulation}

The complete geometry of the DPF is modeled using C3D10 3D elements to capture the detailed deformation behavior. Snap-through is initiated by applying a pressure load to each inner wall of the air chambers. 3D simulations are chosen to capture the dynamic behavior under internal pressure loading. A fixed boundary condition is applied to the first unit to prevent rigid body motion. After dome inversion, a series of relaxation steps are performed to allow the structure to reach its final stable state (see Figure~\ref{fig:si_FE_section} c). To validate the lattice model, simulations are conducted for various geometric parameters, including the number of units ($N_{\text{Units}}$), material thickness ($t$), chamber wall thickness ($t_{\text{ch}}$), elastic modulus (E), dome height ($H$), and base radius ($r_b$). The final deformed shape is extracted for comparison with experimental results. The entire DPF geometry and simulation workflow is implemented using Python scripting in Abaqus, ensuring efficient parameter exploration and model validation.

\newpage
\subsection{Energy-based model and constant derivation}\label{sec:model_and_constants}

\subsubsection{Spring Lattice array}\label{sec:lattice_array}

Predicting the behavior of multistable DPFs requires consideration of both the unit cells' and the global geometric parameters. By characterizing the contribution of each DPF subcomponent and their interactions, an energy landscape can be built where each minimum corresponds to a stable state of the system. The resulting strain from the energy minimization process dictates the programmed stable shapes with their geometrical and stiffness characteristics. We represent the DPR as a lattice comprising nonlinear~\cite{Meaud2020}, linear, and torsional springs (Figure~\ref{fig:si_param_behavior}a). The springs' stiffness and connectivity allow us to map local extensions and rotations to defined energy contributions. We use nonlinear springs featuring a ground (unstressed) state and an inverted (stressed) to capture the influence of the bistable domes on the system. To appropriately represent the bistable dome behavior, we position the nonlinear springs so that the path of extension coincides with the dome's tip position between the ground and inverted states. Linear springs capture connections between the nonlinear units modeled as struts. In the DPF's case, the strain these segments experience is negligible, behaving as a strain-limiting layer. Consequently, we model these connections as nearly rigid. While the axial strains experienced by this strain-limiting layer are negligible, the layer is sufficiently thick to display bending resistance. We capture the influence of this bending stiffness by introducing torsional springs at the nodes coincident with the strain-limiting layer. The angular displacement experienced by the torsional springs is dictated by the change in the angle $\vartheta$ formed by the segments (see Figure~\ref{fig:si_param_behavior}a).We define $k_l=\frac{Et_{\text{lim}}UC}{U_L}$ and $k_t=\frac{Et^3_{lim}}{12(1-\nu^2)}\frac{UC}{U_{\text{sep}}}$ . Where $E$ is the young modulus, $\nu$ is the Poisson ratio, and $t_{lim}$ is the thickness of the limiting layer shown in Figure~\ref{fig:si_dome_phalanx}a. 

\vspace{0.5cm}
Minimizing the system's total energy, $E_{tot}=\sum_{i=1}^n E_{L}+E_{NL}+E_T$ where $n$ is the number of units, constrains the space of possible interactions for the lattice elements into a discrete set of stable states. The rich configuration space of our DPF requires establishing a rational method for providing the initial guess and initializing the minimization process. To this end, we implement a geometric base model to generate initial guesses for the optimization algorithm so that the obtained states are in the neighborhood of physically feasible configurations with improved computational time. The geometric model for the dome phalanx finger assumes a fixed extension of the nonlinear spring and pure rotation from segment to segment. Here, the activated states of the finger (i.e., with units with inverted domes) are determined by extending the length of the linkage corresponding to the bistable dome structure such that it reflects the distance along the central axis between the tip of the dome to the base of the dome chamber. The now extended segment yields an angle $\vartheta$ between the present and initial position of the front segment (see Figure~\ref{fig:si_param_behavior} a). The model uses this $\vartheta$ to calculate the rotation of the subsequent segment, and the process is repeated until the number of segments is exhausted. In this model, the displacement of the overall fingertip can be calculated as a summation of sines and cosines. This purely geometric model does not account for the internal force balancing characteristic of each stable state, which ultimately leads to errors in the configuration predictions. Nevertheless, predictions of a purely geometric model are a powerful tool in finding solutions within the hyperdimensional energy landscape of the spring lattice when we use the outputs of the geometric model to form the initial guess in our search for local extrema. The stable configurations that result from balancing the force and energy contribution of coupled springs differ from the activated configurations determined by the geometric model, but the discrepancy is one easily overcome by the energy minimization process.

\subsubsection{Material viscoelastic response}\label{sec:visco-elastic}
Material time response is captured using a variation of the Kelvin-Voigt viscoelastic constitutive model\cite{visco_1975}. The model consists of two components in parallel: a spring (representing the elastic behavior) and a dashpot (representing the viscous behavior) see Figure~\ref{fig:DPG_Model}a. 

\vspace{0.5cm}

\textbf{Nonlinear Spring (Elastic Element):} The spring represents the material's dome behavior, as explained in the previous section. This spring follows Equation~\ref{eq:Bistable_Spring}.

\vspace{0.5cm}

\textbf{Dashpot (Viscous Element):} The dashpot represents the material's viscous behavior, meaning it resists deformation by dissipating energy as heat. This behavior is modeled by considering two different damping forces, isotropic damping, and internal damping~\cite{ANDREJEVIC2022111607}. The isotropic damping for numerical stability on a node $i$ can be written as:

\begin{equation}
    F_d^{iso}(\dot{x}_i) = \eta_{iso}\dot{x}_i
\end{equation}

where $x_i$ represents the position and node $i$ and $\dot{x}_i$ represents its velocity. Internal damping is selected to be proportional to the spring coefficients as~\cite{ANDREJEVIC2022111607}:

\begin{equation}
F_d^{\text {int }}\left(x_{ij},\dot{x}_i,\dot{x}_j\right) =\eta_{\text {int }}\left(\left(1-\frac{s_{i j}}{\left\|x_{i j}\right\|}\right)\left(\dot{x}_i-\dot{x}_j\right)+\frac{s_{i j}}{\left\|x_{i j}\right\|^3}\left[x_{i j} \cdot\left(\dot{x}_i-\dot{x}_j\right)\right] x_{i j}\right)
\end{equation}

where $s_{ij}$ is the rest length of the spring, and $x_{ij}= x_{i} - x_{j}$. $\eta_{iso}$ is selected to guarantee numerical stability, and $\eta_{\text {int }}$ coefficient is obtained by fitting the dynamic behavior to the experiments for different loading conditions. We obtained $\eta = 0.05$ Ton s$^{-1}$ by utilizing the Prony series characterized in~\cite{riley_sensors_IEEE}.

\subsubsection{Parameter Tuning}\label{sec:parameter_tuning}

\begin{figure}[!t]
  \centering
  \includegraphics[width=\textwidth]{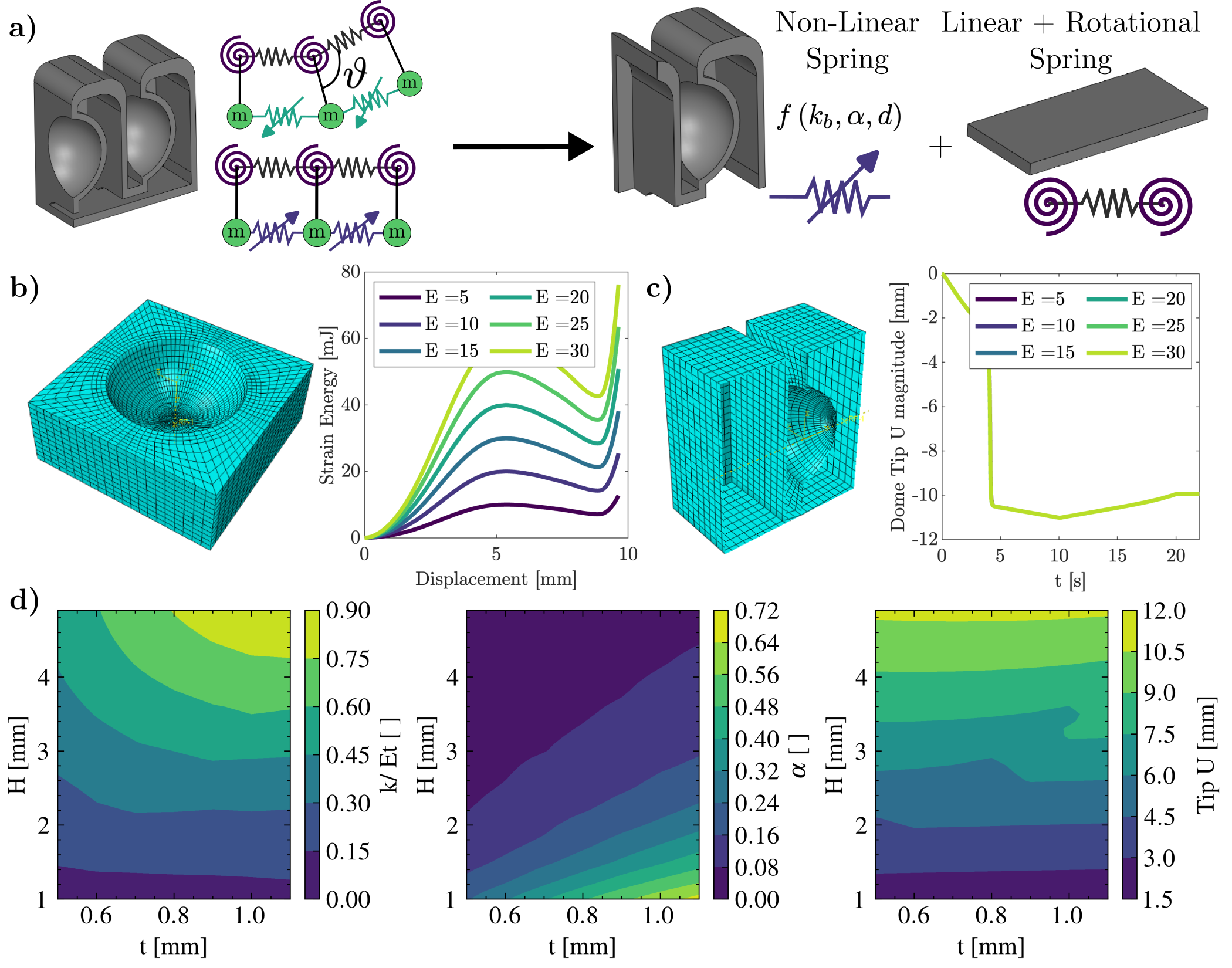}
  \caption{a) DPF fundamental unit cell isolated into different lumped parameter spring elements. (nonlinear + linear + rotational springs) b) nonlinear spring parameter behavior over different dome heights $H$ and dome thickness $t$.}
  \label{fig:si_param_behavior}
\end{figure}

\begin{figure}[!t]
  \centering
  \includegraphics[width=0.9\textwidth]{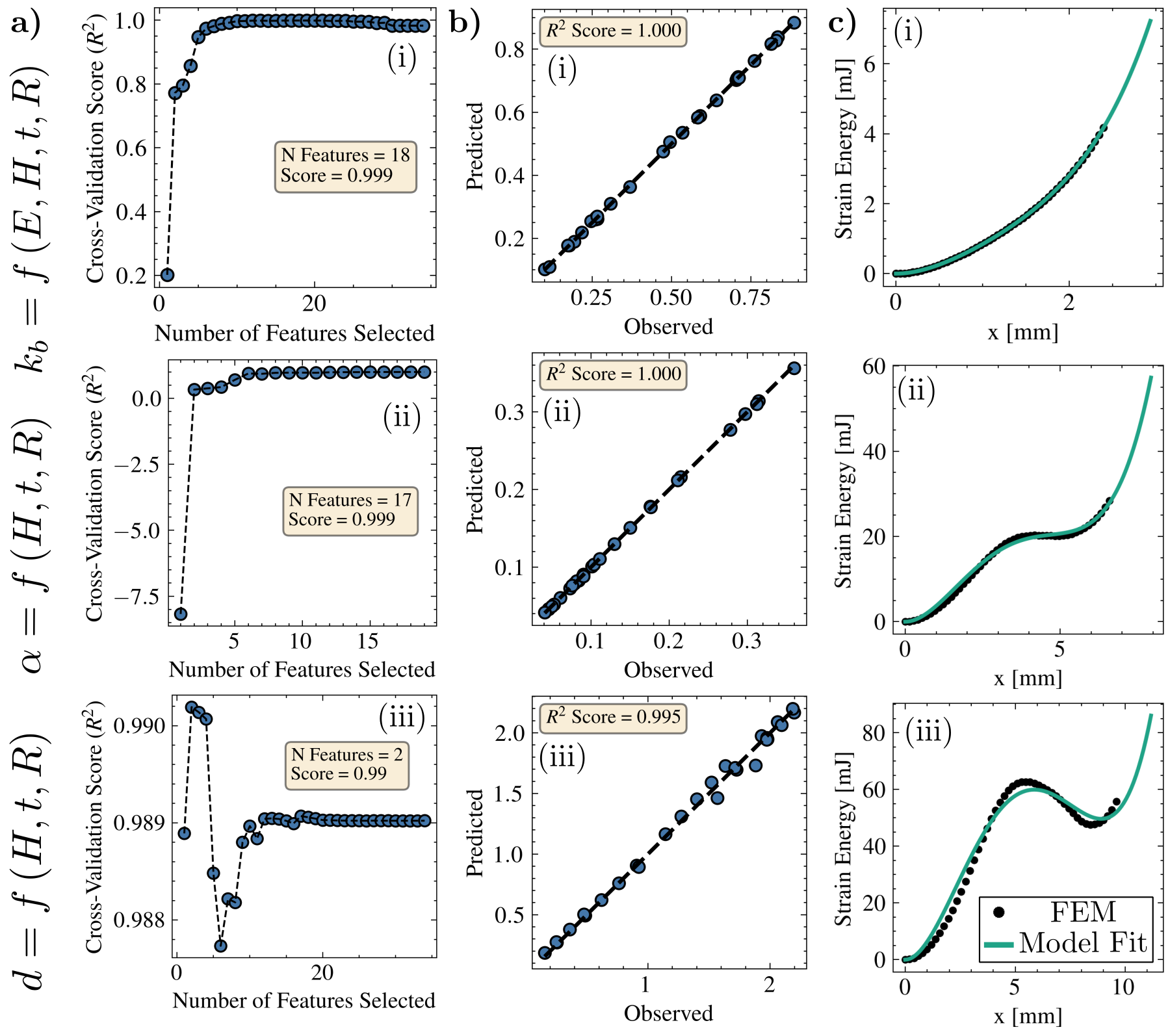}
  \caption{Parameter-fitting results for $\alpha$, $k_b$, $d$ as a function of geometric parameters. a) RFE with ridge regression cross-validation vs the number of features (training data) for  $k_b$ (i), $\alpha$ (ii), and $d$ (iii); b) Prediction vs observation plot on test data for the reduced model for $k_b$ (i), $\alpha$ (ii), and $d$ (iii); c) Model fit for energy as a function of geometric parameters for monostable (i), metastable (ii) and bistable (iii) cases.}
  \label{fig:si_param_tuning}
\end{figure}

To determine each of our lumped model's parameters, the DPF's fundamental unit cell is isolated into two different parts (see Figure~\ref{fig:si_param_behavior}a). The dome unit is fitted as a nonlinear spring following Equation~\ref{eq:Bistable_Spring} and the strain limiting layer is represented by a liner spring to capture stretching and two rotational springs to capture the bending rigidity. Each of the constants for these springs needs to be tuned for different geometrical cases to produce an accurate model for inverse co-design. Given Equation~\ref{eq:Bistable_Spring}, the parameters to be determined for the nonlinear spring are $k_b$, $\alpha$, and $d$. These parameters are functions of the dome unit's mechanical behavior (stiffness and energy barrier) as well as the interaction between the dome and the wall of the air chamber of the subsequent unit cell. The procedure to find each constant as a function of the geometrical parameters of the dome unit cell is as follows. First, different finite element simulations of the unit cell are utilized to determine the best-fitted values for $k_b$, $\alpha$, and $d$ (Figure~\ref{fig:si_param_behavior}b) for different geometrical cases ($H$ and $t$, same as shown in Figure~\ref{fig:si_Unit_Cell}). Different parameters, such as elastic modulus, are examined to establish their influence on the strain energy and tip displacement (Figure~\ref{fig:si_param_behavior}b) and establish a better understanding of the governing parameters to be considered to represent the nonlinear spring accurately. Equation~\ref{eq:Bistable_Spring} is fitted to the FE data using the nonlinear least square method to obtain the overall behavior of these parameters in terms of $H$ and $t$. Results for these fitted parameters can be observed in Figure~\ref{fig:si_param_behavior}c, and the energy-displacement curve with the fitted parameters can be observed in Figure~\ref{fig:si_param_tuning}a.

Once the values are obtained, we use Recursive Feature Elimination (RFE)~\cite{guyon_gene_2002} combined ridge regression to determine an expression for $k_b$, $\alpha$, and $d$ as functions of the dome height ($H$), dome thickness ($t$), and dome curvature ($R$). By using RFE, we can automatically determine the relevance of each feature in our model. Given the main task of inverse design of multistable actuators, a general and scalable model is needed to cover different design cases. We establish the model by considering three different non-dimensional relations:

\begin{equation}\label{eq:shell_numbers}
    \pi_1 = \frac{t}{H}\hspace{0.5cm}\pi_2 = \frac{t}{R}\hspace{0.5cm}\pi_3 = \frac{H}{R}
\end{equation}

where $\pi_1$ represents the load-carrying capacity of the shell, $\pi_2$ is the curvature-to-thickness ratio, and $\pi_3$ quantifies the dome shallowness. Using these three non-dimensional quantities, the following model is fitted:

\begin{equation}\label{param_relations}
\mathbf{\alpha} = \mathbf{C}(\pi_1,\pi_2,\pi_3)\xi_{\alpha} \hspace{0.5cm} \mathbf{k_b} = \mathbf{C}(\pi_1,\pi_2,\pi_3)\xi_{k}\hspace{0.5cm} \mathbf{d} = \mathbf{C}(\pi_1,\pi_2,\pi_3)\xi_{d}
\end{equation}

$\mathbf{C}(\pi_1,\pi_2,\pi_3)$ is a matrix of all the possible candidates and interactions of the non-dimensional relations. The $N\times M$ matrix can be written as:

\begin{equation}\label{eq:c_matrix}
\mathbf{C}(\pi_1,\pi_2,\pi_3) = 
\begin{bmatrix}
\pi_i & \pi_i^2 & \hdots & \pi^n_i & \pi_i\pi_j & \hdots & \pi^n_i\pi^m_j
\end{bmatrix}
\end{equation}

where superscripts $n$ and $m$ represent the maximum degree of the polynomial use for the regression and subscripts $i$ and $j$ represent each of the non-dimensional numbers. For the parameter $\alpha$ we can fully represent this regression as:

\begin{equation}
\begin{split}
&\begin{bmatrix}
\alpha(H_1,t_1,R_1)\\
\alpha_2(H_2,t_2,R_2)\\
\vdots\\
\alpha_k(H_k,t_k,R_k)
\end{bmatrix}=\\
&
\begin{bmatrix}
\pi_i(H_1,t_1,R_1)& \hdots & \pi^n_i(H_1,t_1,R_1) & \pi_i\pi_j(H_1,t_1,R_1) & \hdots & \pi^n_i\pi^m_j(H_1,t_1,R_1)\\
\pi_i(H_2,t_2,R_2)  & \hdots & \pi^n_i(H_2,t_2,R_2) & \pi_i\pi_j(H_2,t_2,R_2) & \hdots & \pi^n_i\pi^m_j(H_2,t_2,R_2)\\
\vdots\\
\pi_i(H_k,t_k,R_k) & \hdots & \pi^n_i(H_k,t_k,R_k) & \pi_i\pi_j(H_k,t_k,R_k) & \hdots & \pi^n_i\pi^m_j(H_k,t_k,R_k)
\end{bmatrix}
\begin{bmatrix}
\xi_1\\
\xi_2\\
\vdots\\
\xi_k
\end{bmatrix}
\end{split}
\end{equation}

Note that we assume that the library of terms used to obtain  $\mathbf{C}(\pi_1,\pi_2,\pi_3)$ has a sufficiently rich column space that the behavior will be represented by Equation~\ref{param_relations}, and it can be written as a linear combination of the weights $\xi$. The data is split into 70\% train and 30\% test data sets to calculate the accuracy of our model. Observation vs prediction plot and estimation error for the three different variables can be observed in Figure~\ref{fig:si_param_tuning}b (ii), where a good fit between the proposed model and the test data is observed. Moreover, we use RFE with cross-validation (cv) on the test data to eliminate features and retest the model to determine its accuracy. $R^2$ for the cross-validation data as a function of the number of features can be observed in Figure~\ref{fig:si_param_tuning}b(i). This gives a reduced model with similar prediction capabilities and the most relevant geometric relations of the problem, and yields an accurate representation of the behavior (see Figure~\ref{fig:si_param_tuning}b (iii)).

The final expressions for each of the nonlinear spring parameters can be written as:

\begin{equation}\label{eq:kb_model}
\begin{aligned}
k_b = f\left(E, \pi_1, \pi_2, \pi_3\right) &= \frac{E}{R^2}\Biggl(-\frac{1.97 R t^5}{H^3}+\frac{7.4 H^2 t^3}{R^2}+\frac{3.5 R t^4}{H^2}+\frac{0.37 H^2 t^2}{R}+\frac{42.2 t^5}{H^2} \\
&\quad -\frac{35.8 H t^4}{R^2}+\frac{71.8 t^5}{H R}-\frac{3.4 R t^3}{H}-\frac{67.7 t^4}{H}+4.2 R t^2+11.1 t^3\Biggr)
\end{aligned}
\end{equation}

\begin{equation}\label{eq:alpha_model}
\begin{aligned}
\alpha = f\left(\pi_1, \pi_2, \pi_3\right) &= \frac{0.4 H^3}{R^3}-\frac{0.6 t^3}{H^3}-\frac{2.9 H^2 t}{R^3}-\frac{0.9 H^2}{R^2}-\frac{9.0 t^3}{H^2 R}+\frac{1.5 t^2}{H^2} \\
&\quad -\frac{3.9 t^3}{H R^2}+\frac{9.2 H t}{R^2}+\frac{17.4 t^2}{H R}+\frac{0.6 H}{R} \\
&\quad +\frac{17.8 t^3}{R^3}-\frac{19.5 t^2}{R^2}-\frac{6.1 t}{R}.
\end{aligned}
\end{equation}

\begin{equation}\label{eq:d_model}
d = f\left(\pi_1,\pi_2,\pi_3\right)= 2.14 H + 0.25 t-H-\text{U}_{\text{sep}}-t_{ch}/2-t/2
\end{equation}

where $H$, $t$ and $R$ are dome unit parameters, $E$ is the elastic modulus of the material,$\text{U}_{\text{sep}}$ and $t_{ch}$ is the separation between units (see Figure~\ref{fig:si_dome_phalanx}).

\subsubsection{Derivation of Jacobian terms}\label{sec:derivation}

Here we detail the derivation for the energy terms of the linear. non-linear and rotational springs.

\vspace{0.5cm}
\textbf{Linear} \textbf{Springs:} The force on a node $i$ at position $x_i$ interacting with and adjacent node $j$ at a position $x_j$ is given by:

\begin{equation}
    F_L(x_i,x_j) = \nabla_{x_i} E_L = \frac{1}{2}k_l\nabla_{x_i}\left(||x_{ij}||-s_{ij}\right)^2 = k_l\left(\frac{s_{ij}}{||x_{ij}||}-1\right)x_{ij}
\end{equation}

where $x_{ij}=x_i - x_j$ and $s_{ij}$ is the rest length of the spring.

\vspace{0.5cm}
\textbf{Nonlinear} \textbf{Springs:} The term term of adjacent nodes $i$ and $j$ connected by a bistable spring can be written as:

\begin{equation}
    E_{NL} = \frac{1}{2}k_b\Bar{x}^2\left(1+(1-\alpha)\left(\frac{\Bar{x}^2}{d^2}-2\frac{\Bar{x}}{d}\right)\right)
\end{equation}

where $\Bar{x}=||x_{ij}||-s_{ij}$. Given this, the non-linear spring forces can be written as:

\begin{equation}
    F_{NL}(x_i,x_j) = \nabla_{x_i}E_{NL} = k_b\Bar{x}\left(1+(1-\alpha)\left(2\frac{\Bar{x}^2}{d^2}-3\frac{\Bar{x}}{d}\right)\right)\frac{x_{ij}}{||x_{ij}||}
\end{equation}

\vspace{0.5cm}
\textbf{Rotational} \textbf{Springs:}  The force on a node $i$ due to bending at the limiting layer is given by

\begin{equation}
    F_T = \nabla_{x_i} E_T(\vartheta) = k_{\vartheta}\left(\vartheta - \vartheta_0\right)\nabla_{x_i}\vartheta
\end{equation}

This force is an interaction between three nodes, thus the gradient $\nabla_{x_i}$ is computed with respect to $x_i\in {x_0,x_1,x_2}$. The relevant vectors can be written as

\begin{equation}
    e_0 = x_1 - x_0
\end{equation}
\begin{equation}
    e_1 = x_2 - x_0
\end{equation}

with $\hat{e}_i = e_i/||e_i||$ denotating the normal vector. We can calculate the gradient of $\vartheta$ as

\begin{equation}
    \nabla_{x_i}\vartheta = \nabla_{x_i}\left(\cos^{-1}\left(\hat{e}_0\cdot\hat{e}_1\right)\right) = -\frac{\nabla_{x_i}\left(\hat{e}_0\cdot\hat{e}_1\right)}{\sin\vartheta}
\end{equation}

The gradients with respect to each component of $x_i$ may be expressed as

\begin{equation}
    \nabla_{x_0}\left(\hat{e}_0\cdot\hat{e}_1\right) = -\frac{e_0 + e_1}{||e_0||||e_1||} + \left(e_0\cdot e_1\right)\left(\frac{e_0}{||e_0||^3||e_1||}+\frac{e_1}{||e_0||||e_1||^3}\right)
\end{equation}

\begin{equation}
    \nabla_{x_1}\left(\hat{e}_0\cdot\hat{e}_1\right) = \frac{1}{||e_0||||e_1||}\left(e_1-\frac{e_0\cdot e_1}{||e_0||^2}e_0\right)
\end{equation}

\begin{equation}
    \nabla_{x_2}\left(\hat{e}_0\cdot\hat{e}_1\right) = \frac{1}{||e_0||||e_1||}\left(e_0-\frac{e_0\cdot e_1}{||e_1||^2}e_1\right)
\end{equation}

Using this expression, we can calculate $\nabla\left(E_L(x_{ij}) + E_{NL}(x_{ij}) +E_T(x_{ij})\right) = F_L + F_{NL} + F_T$.

\newpage
\subsection{DPF Static and Dynamic Response}\label{sec:static_dynamic_model}

Using the parameters from the fitting process, we can predict stable state configurations for the DPF with the resulting model. Given that the nonlinear elements of the gripper are independently stable, the expected number of states for a given dome phalanx finger topology follows $2^n$, where $n$ is the number of bistable segments on the finger. The model can determine the behavior of a given design in orders of magnitude less time than the FE numerical simulations, which makes it feasible for iterating through potential configurations and optimizing for the best design according to a given task (e.g., position and grasping force).

\begin{figure}[!h]
  \centering
  \includegraphics[width=0.9\textwidth]{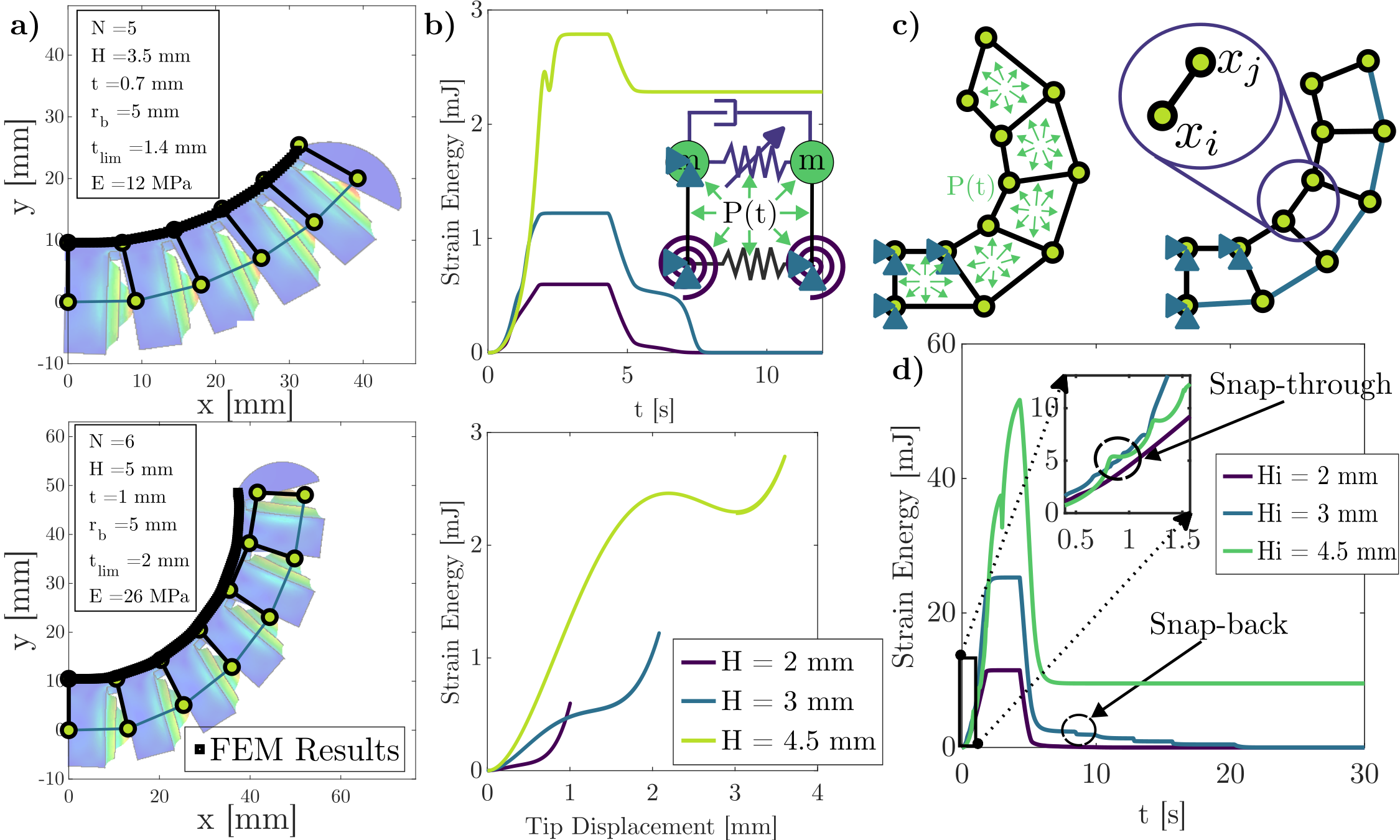}
  \caption{Static and dynamic response of the DPF. a) Stable states are found by minimizing the strain energy. A comparison with FE simulations for five domes of the same height. b) Dynamic response of lattice model with viscoelastic dampers c) Schematic representation of applied load and final DPF position. d) Five-segment DPF dynamic response for different dome H.}
  \label{fig:si_static_dynamic}
\end{figure}

\vspace{0.2cm}
\textbf{Static response:}
The static response of the finger is predicted by minimizing the total energy of the geometry. We performed this process using Matlab's fmincon sqp algorithm~\cite{Han1977}. Our algorithm can predict different final shapes depending on the initial guess geometry. Results for the model are compared with FE simulations, where various points of the limiting layer are compared (see Figure~\ref{fig:si_static_dynamic}a). Results show that our model can predict the final shape of the DPF for different geometric parameters with an error below 3\% compared with FE simulations (see Table~\ref{tab:val_FE} for an extended comparison).

\vspace{0.2cm}
\textbf{Dynamic response:}
The model can be further expanded by including point masses on each of the segments to represent the total mass of the system and dampers to capture the viscoelastic response of the material. By incorporating these elements and determining its jacobian terms (see section \ref{sec:derivation}), a dynamic model can be constructed to capture the time response of the DPF. The equations of motion for the system can be written as

\begin{equation}
    [M]\ddot{x}_i + F_d^{int}(x_{ij},\dot{x}_i,\dot{x}_j) + F_d^{iso}(\dot{x}_i,\dot{x}_j) + \nabla\left(E_L(x_{ij}) + E_{NL}(x_{ij}) +E_T(x_{ij})\right) = F_{ext}(x_{ij},t) 
\end{equation}

where $[M]$ is the mass matrix of the system, $F_d^{int}(x_{ij},\dot{x}_i,\dot{x}_j)$ is the internal damping, which can be modeled as a dashpot in parallel to each spring (see Figure~\ref{fig:DPG_Model}a) and $F_d^{iso}(\dot{x}_i)$ is the isotropic drag, which acts as an external force on each node proportional to velocity ($\dot{x}_i$)~\cite{ANDREJEVIC2022111607}. $F_{ext}(x_{i,j},t)$ is the external actuation pressure that can be calculated as $ F_{ext}(x_{ij},t) =A(x_{ij})P(t)$. Where $A(x_{ij},)$ is the area of the face where the pressure is applied. By substituting $\dot{x}_1=x_2$ and $\dot{x}_2=\ddot{x}_i$, we can rewrite the system as

\begin{equation}
\begin{bmatrix}
\dot{x}_1 \\
\dot{x}_2 
\end{bmatrix}
=
\begin{bmatrix}
x_2 \\
-[M]^{-1}\left( F_d^{int}(x_{ij},\dot{x}_i,\dot{x}_j) + F_d^{iso}(\dot{x}_i,\dot{x}_j) + \nabla\left(E_L(x_{ij}) + E_{NL}(x_{ij}) +E_T(x_{ij})\right) - F_{ext}(x_{ij},t) \right) 
\end{bmatrix}
\end{equation}

\begin{figure}[!h]
  \centering
  \includegraphics[width=\textwidth]{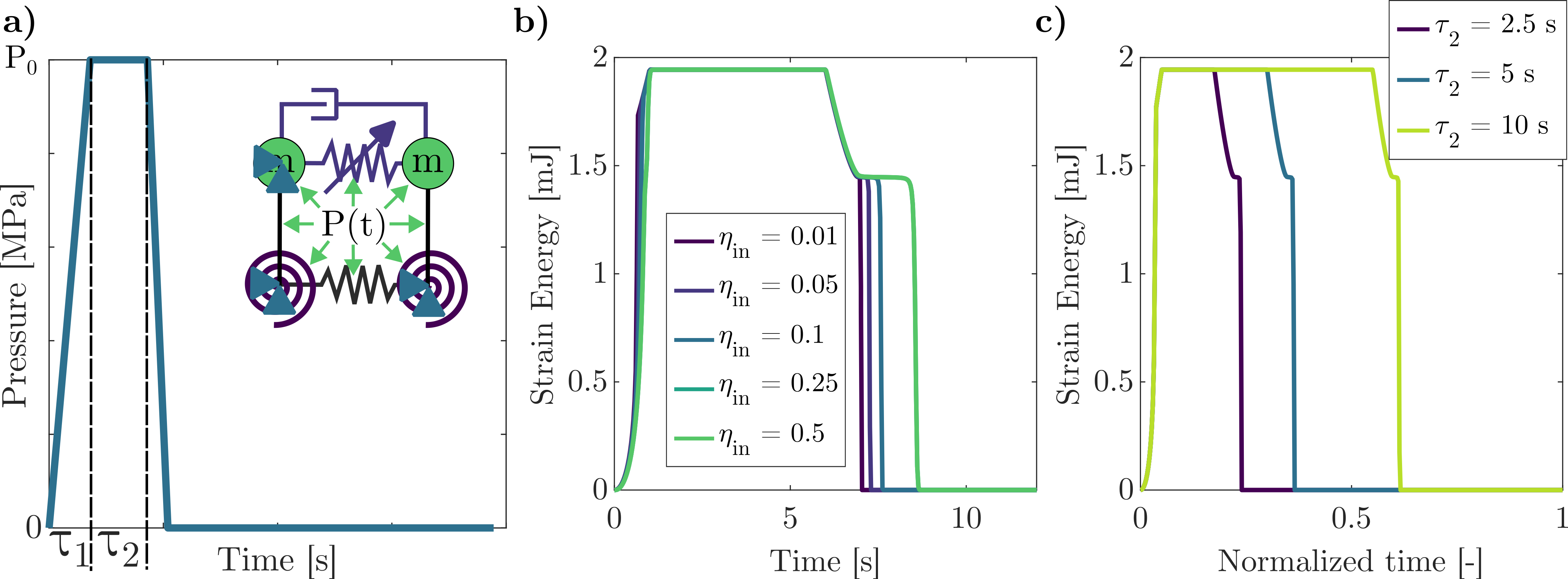}
  \caption{Lattice model dynamic response for dome unit cell. a) Internal pressure over time $P(t)$. Loading time $\tau_1$ and applied load time $\tau_2$. b) Effect of internal damping on dynamic response. c) Effect of applied load time ($\tau_2$) on metastable unit dome reseating time.}
  \label{fig:si_Static_Unit_Cell}
\end{figure}

\begin{figure}[!h]
  \centering
  \includegraphics[width=\textwidth]{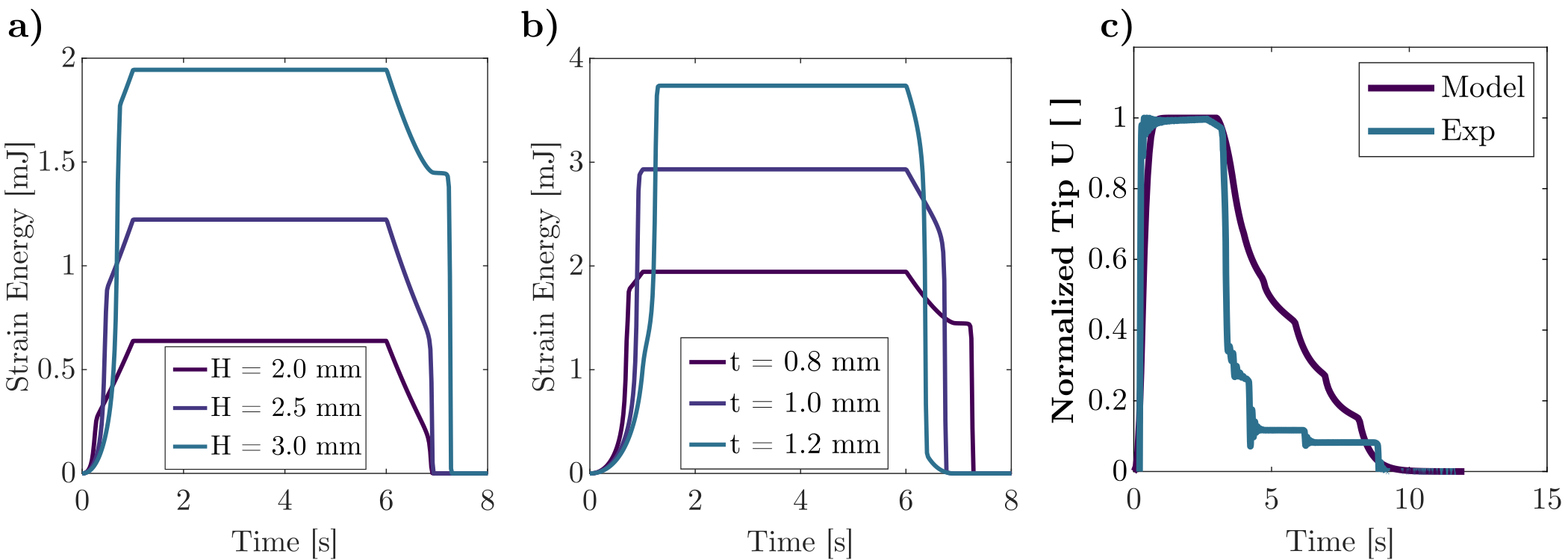}
  \caption{Metastable dome unit analysis. a) Effect of dome height in resetting time of the units (t = 0.8 mm). b) Effect of dome thickness in resetting time of the units (H = 3.0 mm). c) Comparison between experimentally recorded data shown in Figure~\ref{fig:DPG_Geometry}c and our proposed model.}
  \label{fig:si_metaUnit_Cell}
\end{figure}

The system is solved using the Runge-Kutta time integration method for loading and unloading external force conditions. We explore the dynamic behavior of a five-segment DPF by applying the internal pressure as shown in Figure~\ref{fig:si_static_dynamic}c, where we obtain the final stable state as shown and the dynamic response shown in Figure~\ref{fig:si_static_dynamic}d. As expected, the model can capture the dynamic response of a bistable, metastable, and monostable DPF, where snap-through and snap-back phenomena are observed. Furthermore, we analyze the temporal response of the metastable units to assess how geometrical parameters and material damping influence the resetting time. Our results show that adjusting the dome unit’s thickness modifies the resetting time (see Figure~\ref{fig:si_metaUnit_Cell}b). However, an even more pronounced effect is observed when varying the loading duration (see Figure~\ref{fig:si_Static_Unit_Cell}c). This suggests that after manufacturing the DPF, additional tunability can be achieved by controlling the applied pressure (see Movie 6). The resetting time response is validated against experimental data to assess its predictive accuracy. As shown in Figure~\ref{fig:si_metaUnit_Cell}c, the model effectively captures the time scales of a fully metastable DPF. However, due to the inherent randomness in the resetting process when all units are metastable, the tip displacement is not precisely predicted. Nevertheless, this tool enables the design of both the static and dynamic responses by programming the resetting time scales of the metastable domes (see Movie 2 for experimental comparison).

\newpage
\subsection{Model Validation and Simulation time}\label{sec:val_and_sim_time}
The lattice model is evaluated under different geometric configurations as shown in Table~\ref{tab:val_FE}. 

\begin{table}[h]
\centering
\begin{tabular}{cccccccc}
\hline    & N Segments & $R_b$ & $H/R_b$ & t    & E  & $t_{\text{lim}}$ & Error \% \\ \hline
1  & 5          & 5   & 0.7   & 0.7  & 5  & 1.4  & 1.37    \\
2  & 5          & 5   & 0.8   & 0.8  & 5  & 1.6  & 0.62    \\
3  & 5          & 5   & 1     & 1    & 26 & 2    & 0.51    \\
4  & 5          & 8   & 0.7   & 0.75 & 5  & 1.5  & 0.51    \\
5  & 5          & 8   & 0.8   & 1    & 26 & 1.6  & 2.73    \\
6  & 6          & 5   & 0.7   & 0.7  & 5  & 1.4  & 1.42    \\
7  & 6          & 5   & 0.8   & 0.8  & 5  & 1.6  & 0.6     \\
8  & 6          & 5   & 1     & 1    & 26 & 2    & 0.4     \\
9  & 6          & 8   & 0.7   & 0.75 & 5  & 1.5  & 0.44    \\
10 & 6          & 8   & 0.8   & 1    & 26 & 1.6  & 2.47\\ \hline   
\end{tabular}
\caption{List of validation cases for spring model. Error \% reference to the difference between model and FE analysis.}
\label{tab:val_FE}
\end{table}

\begin{figure}[!h]
  \centering
  \includegraphics[width=\textwidth]{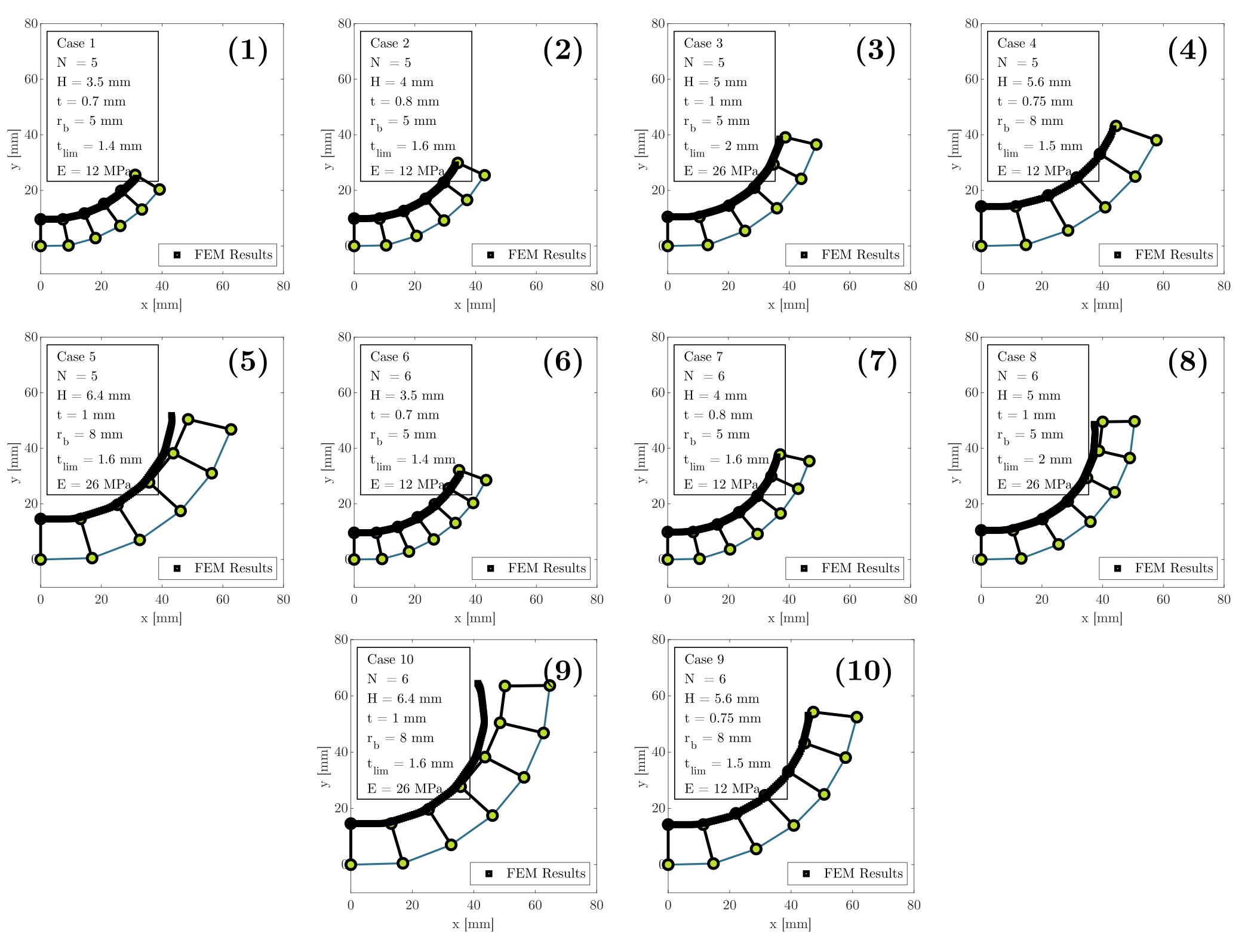}
  \caption{Model comparison with FE method. (1)-(10) Case numbers parameters and error can be observed in Table~\ref{tab:val_FE}.}
  \label{fig:si_model_validation}
\end{figure}

Different numbers of segments, elastic modulus, and limiting layer thickness are included to demonstrate the versatility of our approach. The error between the model and the simulation is calculated by comparing the difference between the final position of the limiting layer after all units are activated (see Fig~\ref{fig:si_model_validation}).

Moreover, our model computational cost is evaluated by comparing the total simulation time simulation time with FE simulations. Results show that the model can yield the final shape of the DPF in less than 1 s. More importantly, the model does not exhibit a coupling between the number of segments and computational time. (see Table \ref{tab:sim_time}).

\begin{table}[h]
\centering
\begin{tabular}{cccc}
\hline N segments  & FE Time [s] & FE Time [h] & Model Time [s]\\ \hline 
2 & 3094  & 0.86 & 0.66 \\
3 & 5960 & 1.66 & 0.58 \\
4  & 12072  & 3.35 & 0.64 \\
5 & 15859 & 4.41 & 0.76 \\
6 & 21455 & 5.96 & 0.67 \\ \hline       
\end{tabular}
\caption{Simulation time comparison between FE analysis and spring model.}
\label{tab:sim_time}
\end{table}

\newpage
\subsection{Geometry Optimization}\label{sec:geo_opti}

\subsubsection{Design space for inverse design}\label{sec:design_space}

We explored the design space of our model by calculating the tip displacement, curvature, and stiffness as a function of the input parameters. Given the width range of parameters, we utilize a Latin Hypercube sampling technique to uniformly sample all of the parameter space ($H_i$, $\text{U}^i_{\text{sep}}$, $\text{U}^i_{\text{L}}$, $t$ and $t_{\text{lim}}$) of the DPF to observe its response after all the units are activated. 1500 samples are taken to guarantee the full exploration of the design space. Analysis for different numbers of units (2 - 7 units) is performed, and the design variables are kept in the ranges listed in Table \ref{tab:var_ranges}.

\begin{table}[h]
\centering
\begin{tabular}{cccc}
\hline 
  & min & max \\ \hline
$H_i$ & 3.0  & 5.0 \\
$\text{U}^i_{\text{sep}}$ & 1.0  & 5.0 \\
$\text{U}^i_{\text{L}}$ & 5.5  & 10.0 \\
$t$ & 0.5  & 1.0 \\
$t_{\text{lim}}$ & 1.0  & 2.0 \\ \hline       
\end{tabular}
\caption{Optimization variables ranges.}
\label{tab:var_ranges}
\end{table}

\begin{figure}[!h]
  \centering
  \includegraphics[width=\textwidth]{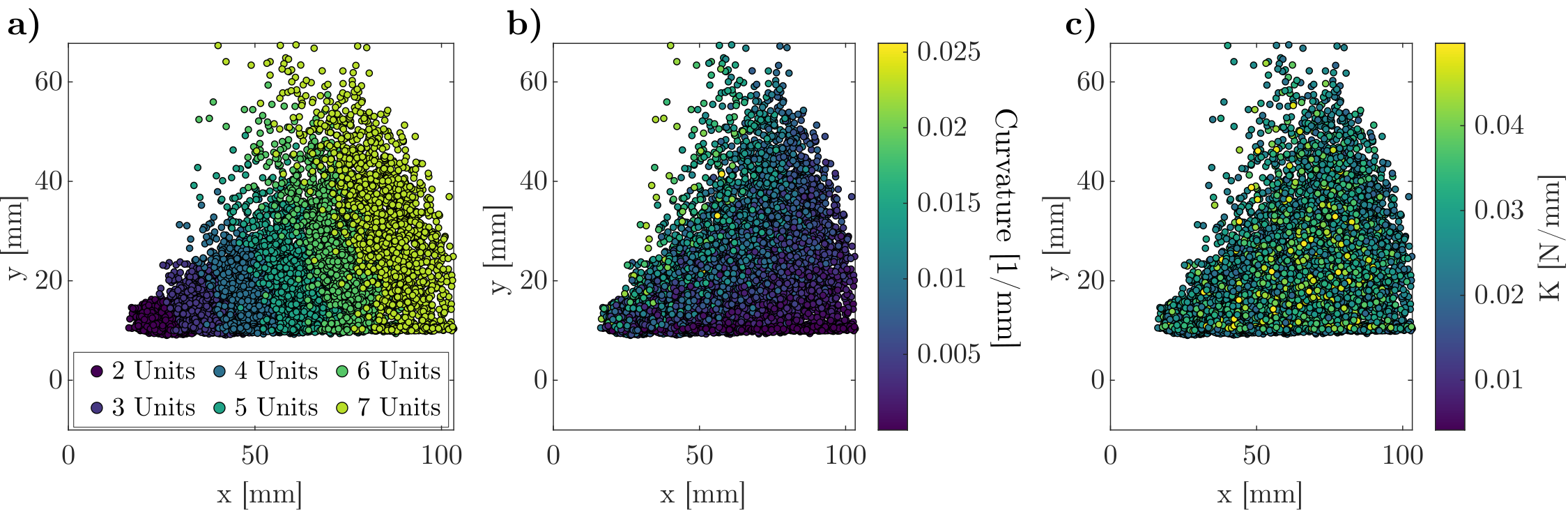}
  \caption{Exploration of the design space. We use our spring lattice model to characterize the deformation modes of structures made of different numbers of units with varying input parameters (Distributed uniformly). a) Tip position design space. b) Curvature design space. c) Stiffness design space.}
  \label{fig:si_design_space}
\end{figure}

\subsubsection{Inverse problem $\rightarrow$ Position Optimization}\label{sec:Position_Opti}

Given a target tip coordinate [x,y] ($\text{Target}_{xy}$), the inverse problem objective function can be written as:

\begin{align} \label{eq:opti_pos}
    &\min_{H_i, t, \text{U}^i_L, \text{U}^i_\text{sep}} \left(\text{Target}_{xy} - \text{Tip}_{\text{dis}} \right)^2\\
    &\text{s.t.} \quad H_{i+1} \leq H_i,\nonumber,\quad i = 1, \dots, N-1\\
    &\hspace{0.5cm}\quad U^{i+1}_{\text{sep}} = U^{i}_{\text{sep}},\quad i = 1, \dots, N\nonumber\\
    &\hspace{0.5cm}\quad U^{i+1}_{\text{L}} = U^{i}_{\text{L}},\quad i = 1, \dots, N\nonumber
\end{align}

\begin{figure}[!h]
  \centering
  \includegraphics[width=0.9\textwidth]{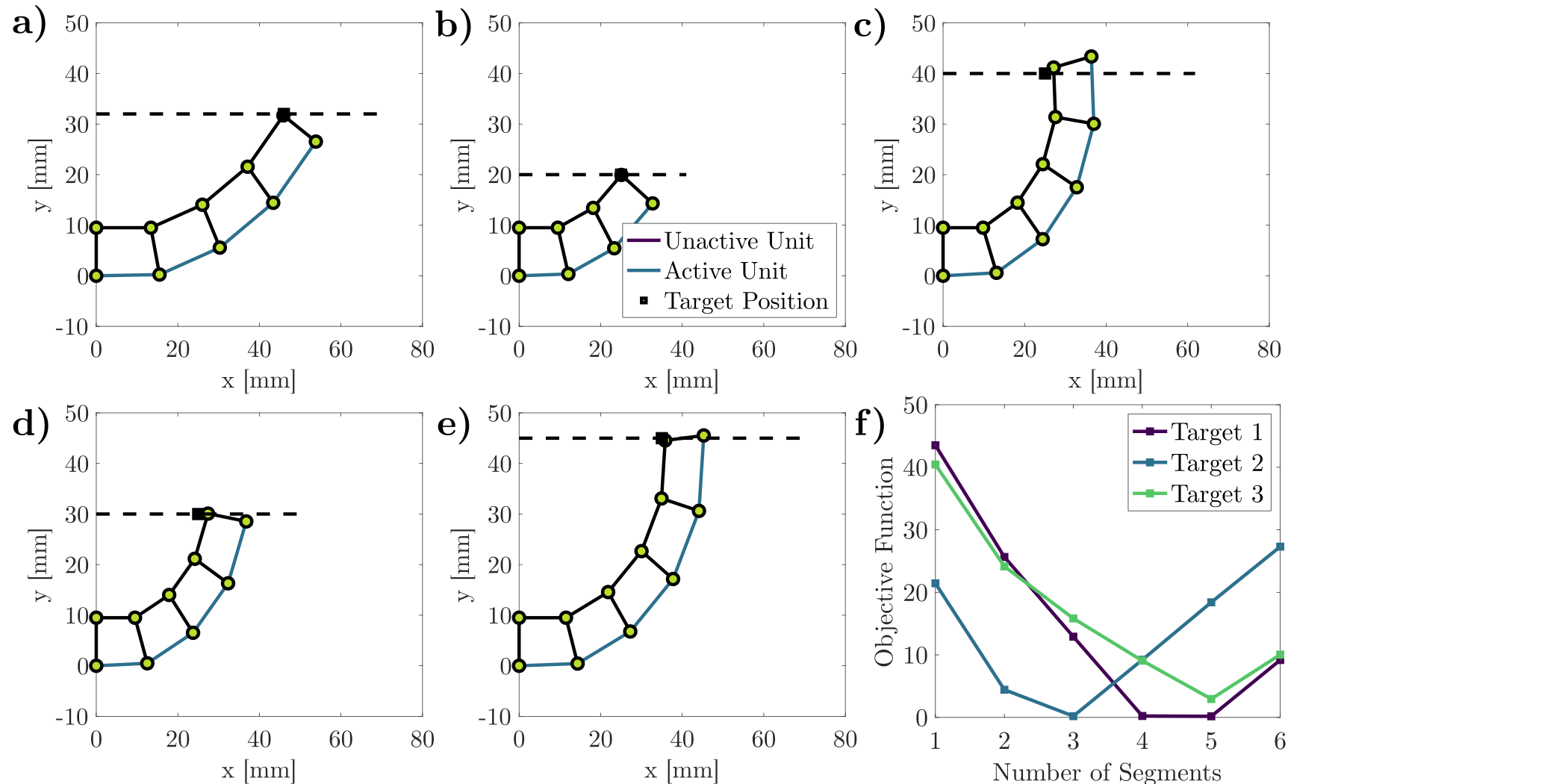}
  \caption{Final results for inverse design problem for tip position as an objective. a) Target 1, b) Target 2, c) Target 3, d) Target 4, e) Target 5, f) Objective function vs Number of segments for the first three target positions.}
  \label{fig:si_inverse_position}
\end{figure}

The final results for the position optimization can be observed in Figure~\ref{fig:si_inverse_position}, where the five target positions and the final shape predicted by the model are shown. The final geometry and objective function of the optimization algorithm are summarized in Table~\ref{tab:position_results_Ninja} and~\ref{tab:position_results_Cheetah}, where we can observe that for targets 1,2 and 5, the desired position is attained with almost perfect precision.

\begin{table}[!h]
\caption{Inverse design problem results for five different targeted positions (NinjaFlex 85A).}
\label{tab:position_results_Ninja}
\centering
\begin{tabular}{cccccc}\hline
             & \textbf{Target 1} & \textbf{Target 2} & \textbf{Target 3} & \textbf{Target 4} & \textbf{Target 5} \\
             & {[}46 32{]}       & {[}25 20{]}       & {[}25 40{]}       & {[}25 30{]}       & {[}35 45{]}    \\\hline
H1           & 4.5               & 4.1               & 4.8               & 4.5               & 4.35              \\
H2           & 4.7               & 4.3               & 4.8               & 4.7               & 4.35              \\
H3           & 4.8               & 4.9               & 4.9               & 5.0               & 4.72              \\
H4           & 5.0               &                   & 4.9               & 5.0               & 4.87              \\
H5           &                   &                   & 5.0               &                   & 4.95              \\
Unit S       & 2.0               & 1.0               & 1.0               & 1.0               & 1.01              \\
Unit L       & 9.9               & 7.0               & 7.3               & 7.0               & 8.99              \\
t       &       0.6         &        0.62        &      0.68          &      0.62         &        0.75       \\
$t_{\text{lim}}$       &       1.5         &      1.3          &         1.2       &      1.4         &         1.25      \\
UC           & 15.0              & 15.0              & 15.0              & 15.0              & 15.0             \\
$K_{\text{dis}}$       &       0.009         &         0.025       &        0.01        &       0.018        &       0.007        \\\hline
\textbf{Obj} & 0.2               & 0.2               & 2.9               & 0.8               & 0.4               \\  
\end{tabular}
\end{table}

\begin{table}[!h]
\caption{Inverse design problem results for five different targeted positions (Cheetah 95A).}
\label{tab:position_results_Cheetah}
\centering
\begin{tabular}{cccccc}\hline
             & \textbf{Target 1} & \textbf{Target 2} & \textbf{Target 3} & \textbf{Target 4} & \textbf{Target 5} \\
             & {[}46 32{]}       & {[}25 20{]}       & {[}25 40{]}       & {[}25 30{]}       & {[}35 45{]}    \\\hline
H1           &       4.87         &      4.52          &        4.89         &      4.96          &   4.9            \\
H2           &        4.8       &        3.93        &       4.72         &        4.52        &     4.87          \\
H3           &        4.67       &        3.63        &       4.52         &        4.08        &    4.72           \\
H4           &         3.98       &                   &        4.33        &        3.99        &      4.35         \\
H5           &                   &                   &       4.02         &                   &     4.3          \\
Unit S       &        2.31        &       1.10         &        1.17        &       1.08         &    1.52           \\
Unit L       &        9.83        &      7.28          &       6.95         &      6.93         &    8.8           \\
t       &       0.79         &        0.83        &       0.65         &      0.76         &         0.63      \\
$t_{\text{lim}}$       &      1.63          &         1.59       &       1.19         &       1.07        &       1.72        \\
UC           & 15.0              & 15.0              & 15.0              & 15.0              & 15.0             \\
$K_{\text{dis}}$       &      0.093          &        0.341        &         0.079       &       0.112        &        0.063       \\\hline
\textbf{Obj} &     0.775           &      0.056          &       0.55         &   0.22             &       0.32         \\ 
\end{tabular}
\end{table}

\subsubsection{Inverse problem $\rightarrow$  Position + Stiffness Optimization}\label{sec:SI_Stiff_pos_Inv}

We further expand our model to include stiffness optimization in our objective function. The finger’s stiffness is calculated by using the dynamic model described in section~\ref{sec:static_dynamic_model}, by perturbing the structure from a programmed stable configuration (i.e., set point). We determine the system’s total internal force by solving the dynamic equation:

\begin{align}\label{eq:stiff_dynamics}
    &[M]\ddot{x_i} + F_d^{int}(x_{ij},\dot{x_i},\dot{x_j}) + F_d^{iso}(\dot{x_i},\dot{x_j}) + F_{\text{in}} = F_{ext}(x_{i,j},t)\\
    & F_{\text{in}} = \nabla\left(E_L(x_{ij}) + E_{NL}(x_{ij}) + E_T(x_{ij})\right)\nonumber
\end{align}

where  $F_{\text{ext}}(x_{i,j}, t)$  is a follower force applied at the tip of the DPF (see Figure~\ref{fig:Inv_design}d) to perturb the DPF from its equilibrium position. The gradient of the internal forces in the structure, $F_{\text{in}}$, is obtained through the analytical expression described in Section~\ref{sec:model_derivation}. The force is applied at a constant load rate of  5 mm/min, ensuring a quasi-static regime. The stiffness of the DPF in the desired state is calculated by determining the internal force ($F_{\text{in}}$) and the corresponding tip displacement over time (see Figure~\ref{fig:Inv_design}). A linear fit is then applied to the resulting data. Using this methodology, the DPF's geometry can be optimized to achieve a specific target position with maximum stiffness in the vicinity of that stable state (see Figure~\ref{fig:Inv_design}e).

By incorporating finger stiffness, grasping force is increased by changing the geometric parameters. Results for the optimization algorithm, where the same number of segments as in Table~\ref{tab:position_results_Ninja} and~\ref{tab:position_results_Cheetah} is maintained, are shown in Table~\ref{tab:stiff_results_Ninja} and~\ref{tab:stiff_results_Cheetah}.

\begin{table}[!h]
\caption{Inverse design problem results for five different targeted positions with maximum stiffness (NinjaFlex 85A).}
\label{tab:stiff_results_Ninja}
\centering
\begin{tabular}{cccccc}\hline
             & \textbf{Target 1} & \textbf{Target 2} & \textbf{Target 3} & \textbf{Target 4} & \textbf{Target 5} \\
             & {[}46 32{]}       & {[}25 20{]}       & {[}25 40{]}       & {[}25 30{]}       & {[}35 45{]}    \\\hline
H1     & 4.63        & 4.10        & 4.74        & 4.81        & 4.41        \\
H2     & 4.67        & 4.27        & 4.83        & 4.82        & 4.58        \\
H3     & 4.90        & 4.49        & 4.85        & 4.87        & 4.71        \\
H4     & 4.98        &             & 4.86        & 4.92        & 4.81        \\
H5     &             &             & 4.97        &             & 4.92        \\
Unit S & 2.04        & 1.00        & 1.08        & 1.02        & 1.07        \\
Unit L & 9.94        & 7.00        & 7.07        & 7.03        & 8.85        \\
t       &     0.83           &       0.86         &       0.81         &     0.85          &        0.83       \\
$t_{\text{lim}}$       &        1.5        &        1.36        &       1.45         &     1.35          &         1.4      \\
UC           & 15.0              & 15.0              & 15.0              & 15.0              & 15.0             \\
$K_{\text{st}}$       &      0.011          &        0.0283        &        0.0146        &     0.0204          &        0.009       \\\hline
Improvement       &     1.22          &        1.13        &        1.46        &      1.13         &     1.3          \\\hline
\end{tabular}
\end{table}

\begin{table}[!h]
\caption{Inverse design problem results for five different targeted positions with maximum stiffness (Cheetah 95A).}
\label{tab:stiff_results_Cheetah}
\centering
\begin{tabular}{cccccc}\hline
             & \textbf{Target 1} & \textbf{Target 2} & \textbf{Target 3} & \textbf{Target 4} & \textbf{Target 5} \\
             & {[}46 32{]}       & {[}25 20{]}       & {[}25 40{]}       & {[}25 30{]}       & {[}35 45{]}    \\\hline
H1           &       4.99         &       4.66         &       4.96         &       5.0         &        4.9       \\
H2           &       4.88        &         4.41       &        4.9        &       4.78         &        4.687       \\
H3           &       4.86       &        4.31        &        4.72        &         4.69       &         4.57      \\
H4           &        4.15        &                   &       4.4         &        4.66        &         4.225      \\
H5           &                   &                   &       3.92         &                   &         4.2      \\
Unit S       &        2.26        &       1.4         &       1.11         &        1.01        &       1.03        \\
Unit L       &        9.98        &        6.85        &        7.11        &        6.89       &        8.92       \\
t       &       0.9         &      0.91          &       0.88         &      0.99         &      0.99         \\
$t_{\text{lim}}$       &       1.89         &        1.95        &       1.24         &      1.66         &    1.94           \\
UC           & 15.0              & 15.0              & 15.0              & 15.0              & 15.0             \\
$K_{\text{st}}$       &      0.093          &        0.341        &        0.079        &      0.112         &     0.063          \\\hline
Improvement       &     1.51          &        1.61        &        1.48        &      2.7         &     2.37          \\\hline
\end{tabular}
\end{table}

\newpage
\subsection{Experimental Test}\label{si:exp_tests}

\subsubsection{Pressure and displacements Measurements}\label{si:P_d_measure}

\begin{figure}[h]
  \centering
  \includegraphics[width=0.8\textwidth]{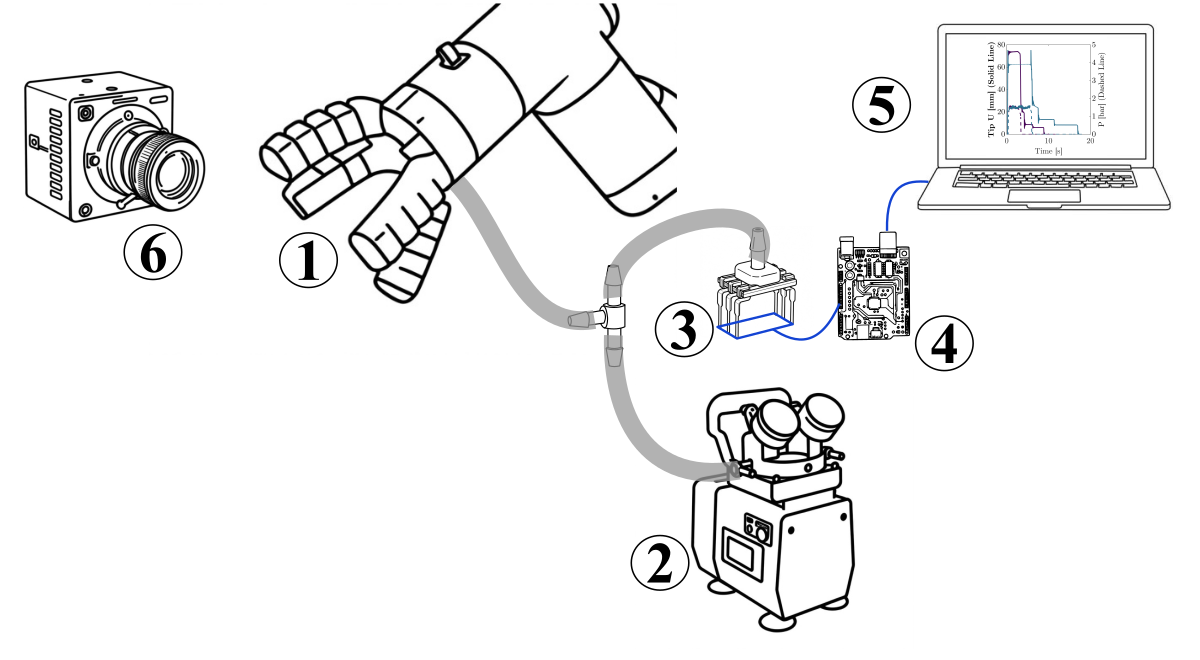}
  \caption{Experimental setup for pressure and displacement measurement. Schematic of the test setup used to record the displacement and pressure over time of the DPF and DFR. (1) DPR or DPF (2) Air Pump. (3) Pressure sensor. (4) Arduino Uno. (5) Laptop. (6) High Photron Speed camera.}
  \label{fig:si_pressure_system}
\end{figure}

As shown in Figure~\ref{fig:si_pressure_system}, an air compressor (Pump DOA-P704-AA) is used to supply air to the DPF. The pressure is measured using a HONEYWELL ABPDANN010BG2A3 pressure sensor (0-10 Bar range), which is connected to an Arduino Uno for data acquisition. The Arduino records pressure readings at 500ms intervals to ensure accurate monitoring. The pressure is gradually increased to a specified value before being released to invert the dome units. The dynamic behavior of the DPF is captured in detail using a Photron Fastcam Mini UX100 high-speed camera, as illustrated in Movie 1. Tip displacement over time is then analyzed using the Tracker Video Analysis and Modeling Tool, providing precise measurements of the system’s response.

\subsubsection{Inverse problem experimental validation}
\begin{figure}[h]
  \centering
  \includegraphics[width=\textwidth]{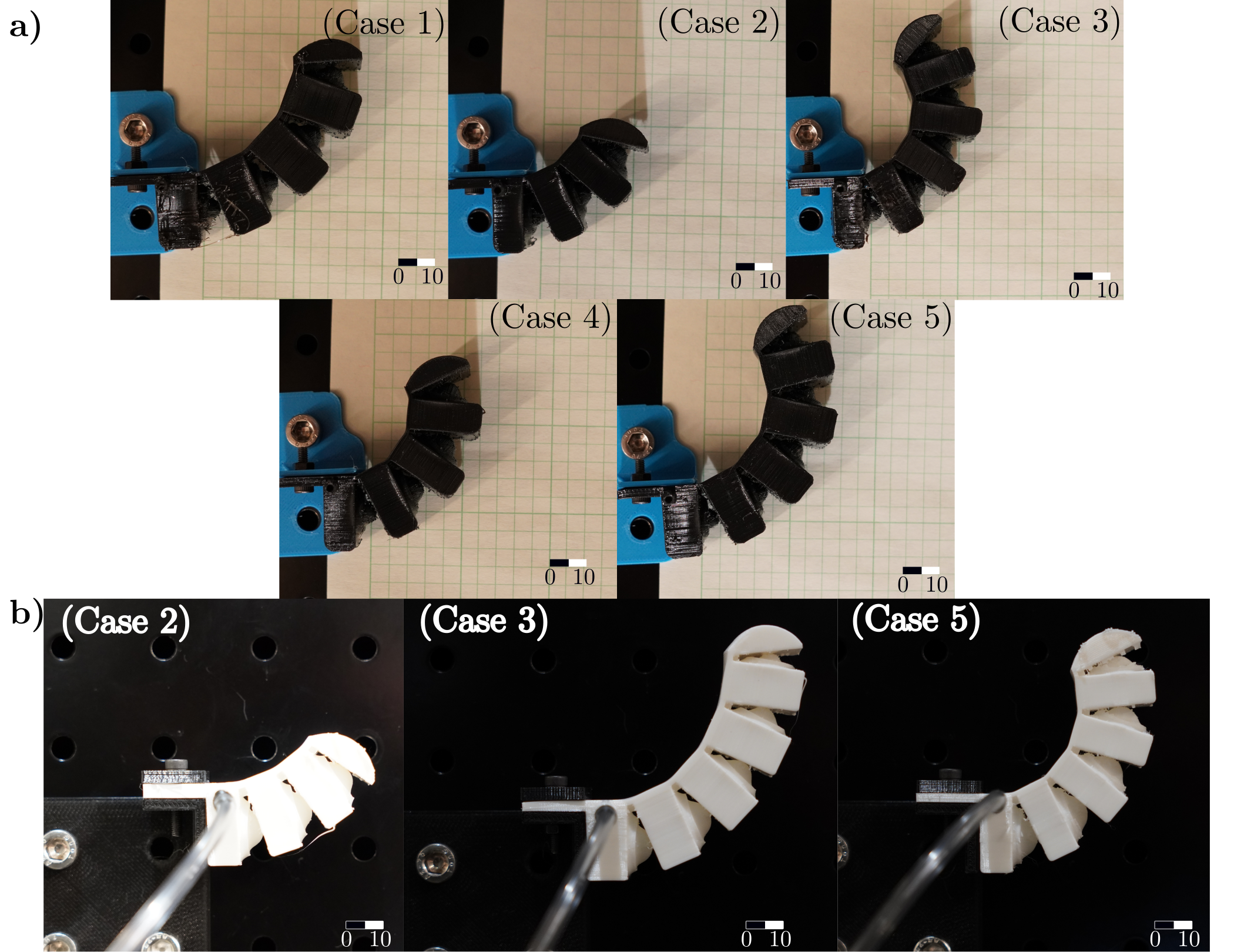}
  \caption{Experimental results for different DPF geometries. a) Printed samples for inverse design target positions (i) - (v) for Ninjaflex 85A b) Printed samples for inverse design target positions (i) - (iii) for Cheetah 95A.}
  \label{fig:si_exp_validation}
\end{figure}

Five geometries are tested by activating all bistable unit cells and measuring the final tip position. The final position of the finger is measured by using image processing. Results can be observed in Figure~\ref{fig:si_exp_validation}a (i-v) for NinjaFlex 85A and Figure~\ref{fig:si_exp_validation}b (i-iii) for Cheetah 95A, and the target coordinates, model prediction, and experimental measurements are reported in Table~\ref{tab:experimental_val}.

\begin{table}[!h]
\centering
\begin{tabular}{cccc}\hline
\textbf{Target Coordinates}      & \textbf{Model}  & \textbf{Experiments} & \textbf{Error}\\\hline
{[}46 , 32{]} & {[}46.85 , 31.22{]} & {[}42.9 , 37.7{]}   & 13\%\\
{[}25 , 20{]} & {[}25.34 , 20.34{]} & {[}23.53 , 22.93{]} & 10\%\\
{[}25 , 40{]} & {[}30.10 , 40.70{]} & {[}29.4 , 43.5{]}   & 6\%\\
{[}25 , 30{]} & {[}28.76 , 29.87{]} & {[}23.0 , 34.26{]}  & 17\%\\
{[}35 , 45{]} & {[}37.98 , 44.36{]} & {[}36.8 , 50.77{]}  & 11\%\\\hline
\end{tabular}
\caption{Experimental measurements for five different DPFs optimized for a specific target position (NinjaFlex 85A).}
\label{tab:experimental_val}
\end{table}

\begin{table}[!h]
\centering
\begin{tabular}{cccc}\hline
\textbf{Target Coordinates}      & \textbf{Model}  & \textbf{Experiments} & \textbf{Error}\\\hline
{[}46 , 32{]} & {[} 45.6, 31.8 {]} & {[}43.31 , 28.38{]}   & 6.9\%\\
{[}25 , 20{]} & {[} 24.7, 19.5{]} & {[} 25,17, 16.55{]}   & 4.3\%\\
{[}25 , 40{]} & {[} 25.15, 39.3{]} & {[} 23.03, 39.29{]}   & 2.5\%\\
{[}25 , 30{]} & {[}25.2, 29.22{]} & {[} 22.9, 31.6{]}   & 1.1\%\\
{[}35 , 45{]} & {[} 33.97, 45.3{]} & {[} 33.35, 44.9{]}   & 1.2\%\\\hline
\end{tabular}
\caption{Experimental measurements for five different DPFs optimized for a specific target position (Cheetah 95A).}
\label{tab:experimental_val}
\end{table}

\newpage
\subsubsection{Experimental validation stiffness optimization}

The stiffness of each DPF is measured using an Instron universal testing machine and applying a perpendicular force to the tip of the actuator (see Figure~\ref{fig:si_exp_validation}b). Linear behavior is observed in the neighborhood of each stable state (0, 2, 4, and 5 active units in Figure~\ref{fig:si_exp_validation}b), with a different stiffness for every state. Finally, the inverse design problem results are tested by comparing the stiffness of two geometries with the same target position but two distinct objective functions (Equation~\ref{eq:opti_pos} and Equation~\ref{eq:opti_pos_stiff}).

\begin{figure}[h]
  \centering
  \includegraphics[width=0.9\textwidth]{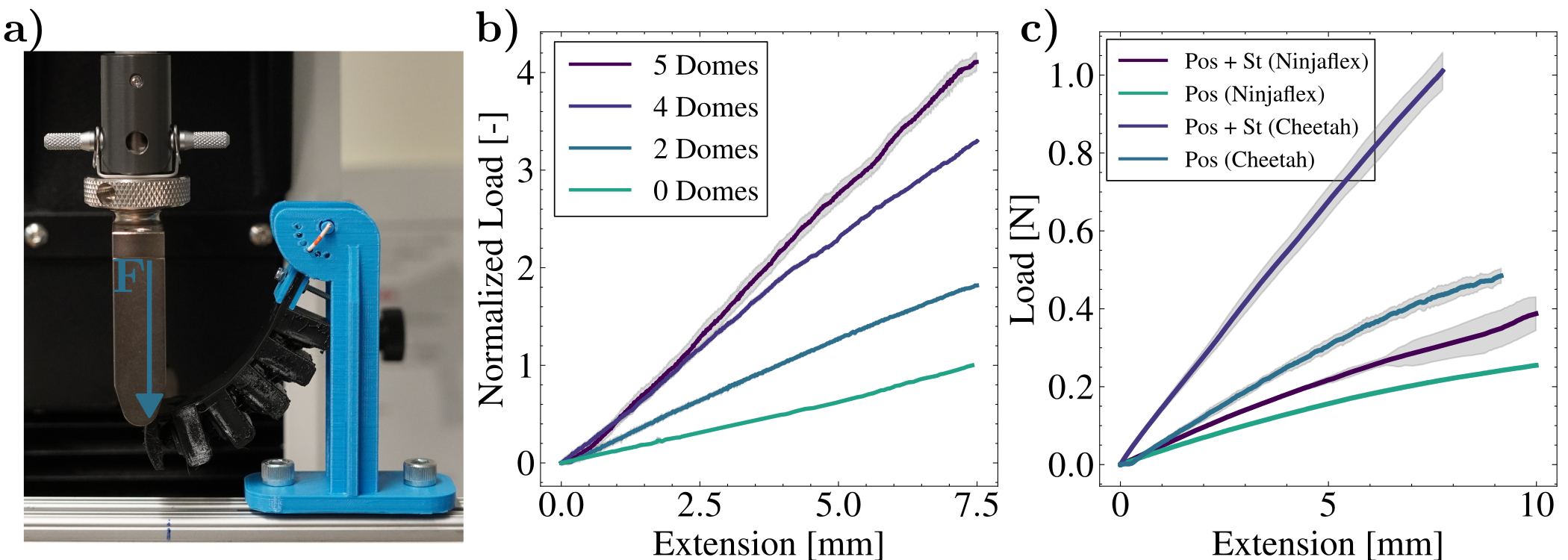}
  \caption{Force vs displacement plot for different DPF. a) Experimental setup. b) Effect of number of active domes on DPF stiffness. c) Comparison between DPF geometries for position inverse design and position+stiffness inverse design for target 5 (Cheetah 95A and NinjaFlex 85A).}
  \label{fig:SI_Intron}
\end{figure}

\subsubsection{Durability Test}

The robustness of the DPF design is tested by performing 100 cycles over three weeks to show the geometry's repeatability and effect. Results show that after three weeks and 100 cycles per week, the tip position stays within 5\% of the target tip displacement.

\begin{figure}[!h]
  \centering
  \includegraphics[width=\textwidth]{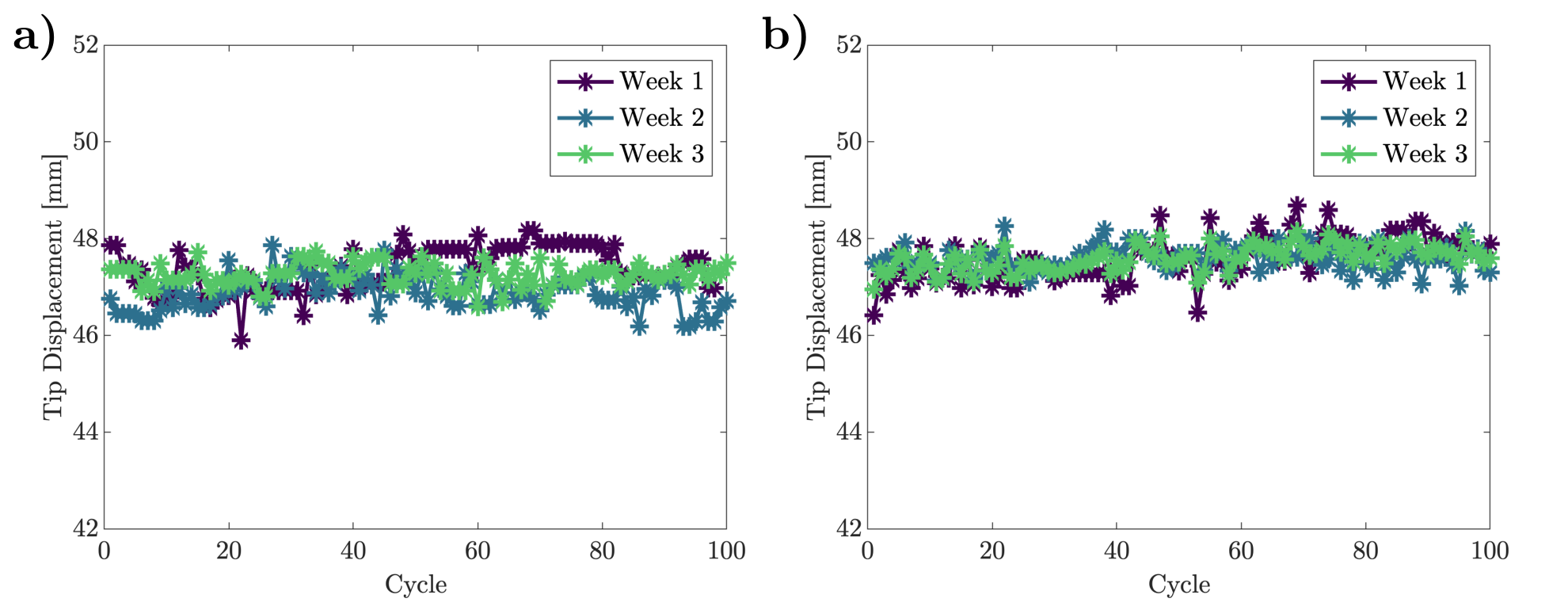}
  \caption{Durability test and time decay response for a) NinjaTek NinjaFlex b) NinjaTek Cheetah (95A).}
  \label{fig:SI_durability}
\end{figure}

Notice that the characteristics of multistability reduce the potential effects of material variability by making geometry (i.e., morphology) the dominant driver of our soft robot’s response.

\subsubsection{Carrying Capacity}
\begin{figure}[!t]
  \centering
  \includegraphics[width=\textwidth]{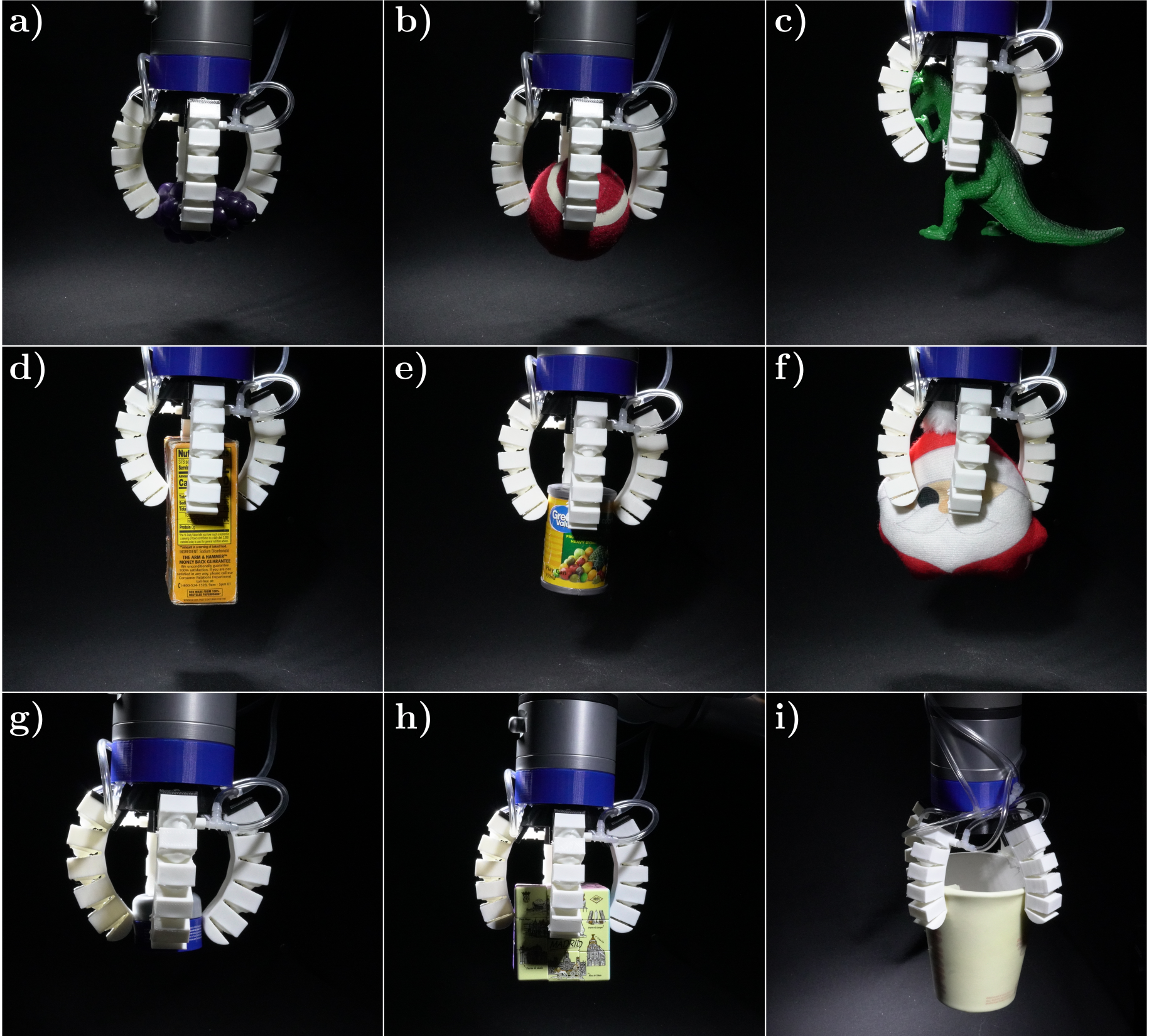}
  \caption{DPF grasping capabilities are shown with six different objects. a) Blackberry Replica, b) Tennis Ball, c) Toy Dinosaur, d) Baking Soda pack, e) Mini Fruit Can. f) Plush Toy g) Cylindrical Cream. h) Rubik's Cube. i) Coffee Cup.}
  \label{fig:SI_carrying_capacity}
\end{figure}

Payload capacity and object grasping tests were conducted to evaluate the performance of our DPR, demonstrate its grasping capabilities, and determine the maximum carrying weight of the system. The designed global states enable adjustments in output force and position based on the number of inverted domes (see Movie 5). This adaptability allows the DPR to grasp objects of various sizes and topologies by removing the actuation pressure. Figure~\ref{fig:SI_carrying_capacity} illustrates the range of objects successfully grasped by the DPR using only the energy stored in the structure due to dome inversion. As depicted, diverse object types are effectively grasped, highlighting the system’s versatility. Additionally, to characterize the payload capacity, weights were incrementally added to the object shown in Figure~\ref{fig:SI_carrying_capacity} (i) until the structure could no longer grasp it. A maximum load of 738 g was recorded, resulting in a maximum load-to-weight ratio of 10.25 (See Movie 5).

\subsubsection{Scalability}

As the mechanical response of the DPF is derived from the dome geometry and its stability, the DPF and robot can be scaled to increase to large dimensions just by considering the energy stored during the dome inversion. Following the analysis by Seffen and Vidoli~\cite{Seffen_2016}, we can derive different additional relations for scaling, given the energy required for inverting a single dome. This energy is proportional to $\frac{Et^3}{1-\nu^2}\left(\frac{H}{R}\right)^2$ where E, $\nu$, t, and R are the elastic modulus, Poisson’s ratio, thickness, height, and radius of the domes respectively. We can utilize the dome shallowness $\frac{H}{R}$ and the curvature-to-thickness ratio $\frac{t}{R}$ to either scale up or down the DPF geometry. It should be mentioned that our dome unit bistable behavior can be geometrically scaled; however, for metastable units, it is necessary to consider the effect of the viscoelastic response on a larger system as this response is a combination of the geometry and material response. To test this, we 3D printed two different geometries with the same energetics but a 2x difference.

\begin{figure}[t!]
  \centering
  \includegraphics[width=\textwidth]{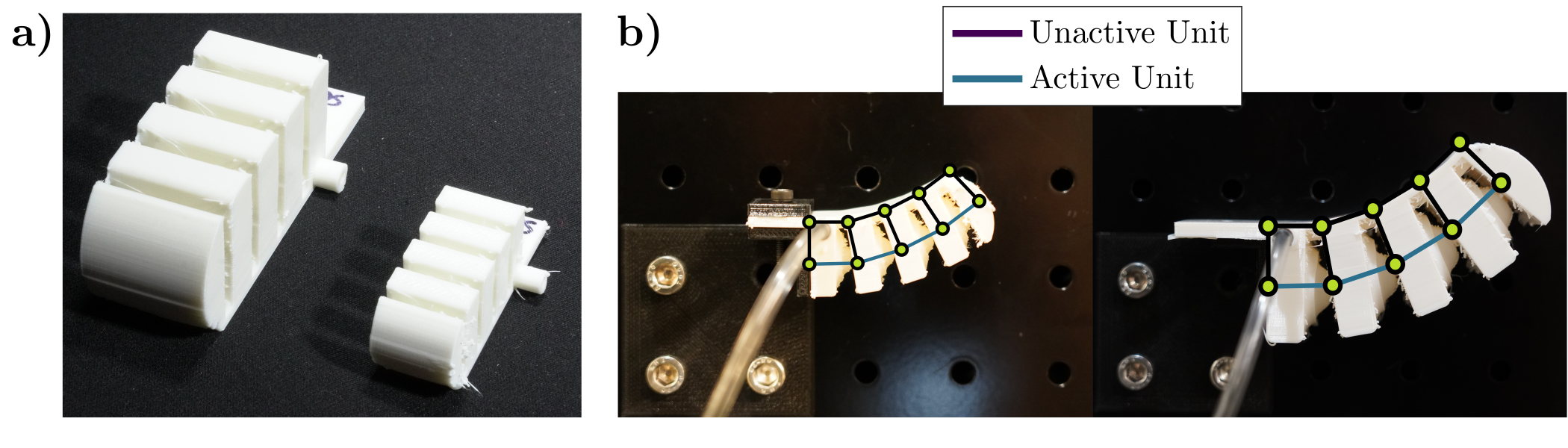}
  \caption{DPF scalability test. a) Two different DPF specimens scale using adimensional relationships. b) Stable states with all dome units active and a comparison with our lattice model.}
  \label{fig:SI_scalability}
\end{figure}

\newpage
\subsection{Dynamic response for Pick and Place Application}\label{sec:DPG_pick_place}

\begin{figure}[!h]
  \centering
  \includegraphics[width=0.95\textwidth]{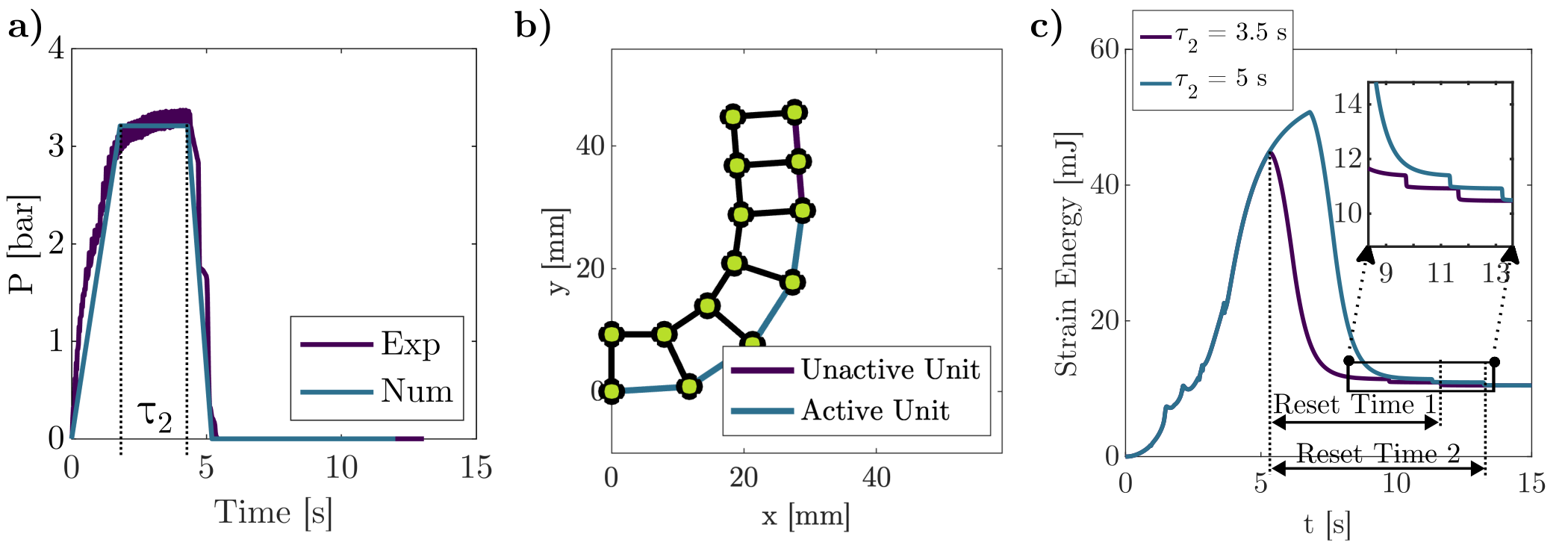}
  \caption{Pick and place task planning set points and model inputs. a) Pressure profile for task planning. b) Setpoint (stable state) for the first active three bistable units. c) Dynamic response for the bistable + metastable actuator.}
  \label{fig:si_inverse_meta}
\end{figure}

For our pick and place, we create a pressure profile that accurately matches the experimental measurements for our DPF (see Figure \ref{fig:si_inverse_meta}a). At the same time, the optimization algorithm is utilized to determine the dome heights of the last two units and the remaining geometric parameters ($t$).

\begin{figure}[!h]
  \centering
  \includegraphics[width=0.95\textwidth]{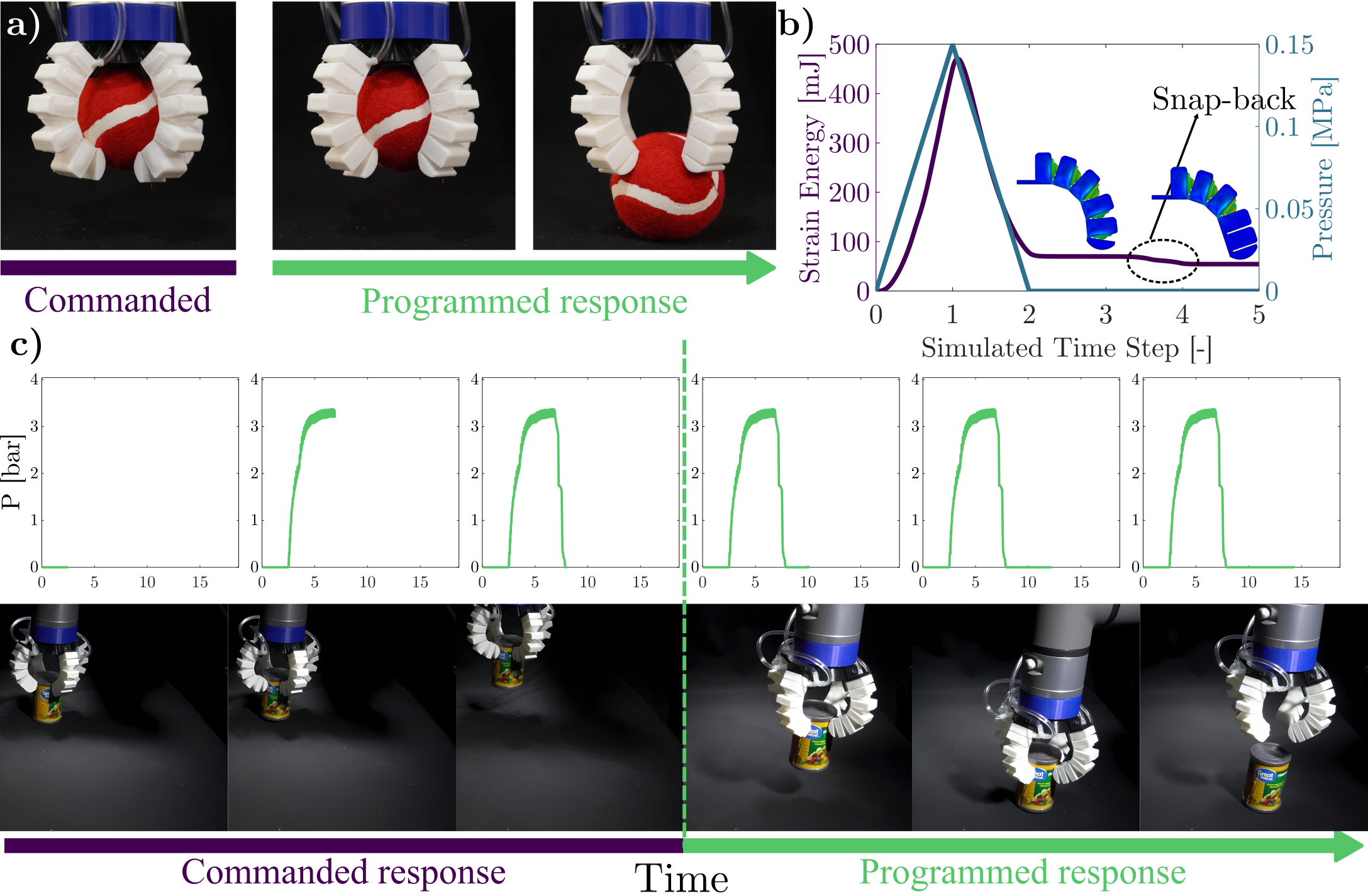}
  \caption{Programmed Pick and Place application for the Dome Phalanx Gripper a) Close up of commanded response when picking a tennis ball.  b) Bistable + Metastable finger behavior: Strain energy and pressure vs. time response. c) Snapshots of embodied pick-and place tasks: A single pressure input is given for the robot to pick the object (Commanded response) and then release it depending on the viscoelastic material response (Programmed response)(see Movie 7).}
  \label{fig:si_pick_place}
\end{figure}

\begin{table}[h]\centering
\caption{Pick and Place DPF final geometry}
\label{tab:pick_place_geo}
\begin{tabular}{cc}
\hline 
H1           &        4.0\\
H2           &        4.5\\
H3           &        5.0\\
H4           &        5.0\\
H5           &        3.0\\
H6           &        3.0\\
Unit S      &        1.0\\
Unit L      &        6.0\\
t      		&        0.83\\
$t_{\text{lim}}$  &      1.0        \\
UC           &      15.00  \\
\hline       
\end{tabular}
\end{table}

Given this, a DPG architecture is built that utilizes the dynamic programmed response (see Figure \ref{fig:si_pick_place}a) to perform a simple pick-and-place application after the pressure is released. This behavior is achieved as the metastable domes reset due to visco-elastic effects, showing a snap-back phenomenon on a specific time scale (see Figure \ref{fig:si_pick_place}). It should be mentioned that the metastability can be utilized in either stable state, meaning this embodied task can be performed for different gripper apertures (see Movie 6). 

\newpage
\subsubsection{Embodied Classification Task}\label{sec:Multi_robot}

\begin{figure}[!h]
  \centering
  \includegraphics[width=\textwidth]{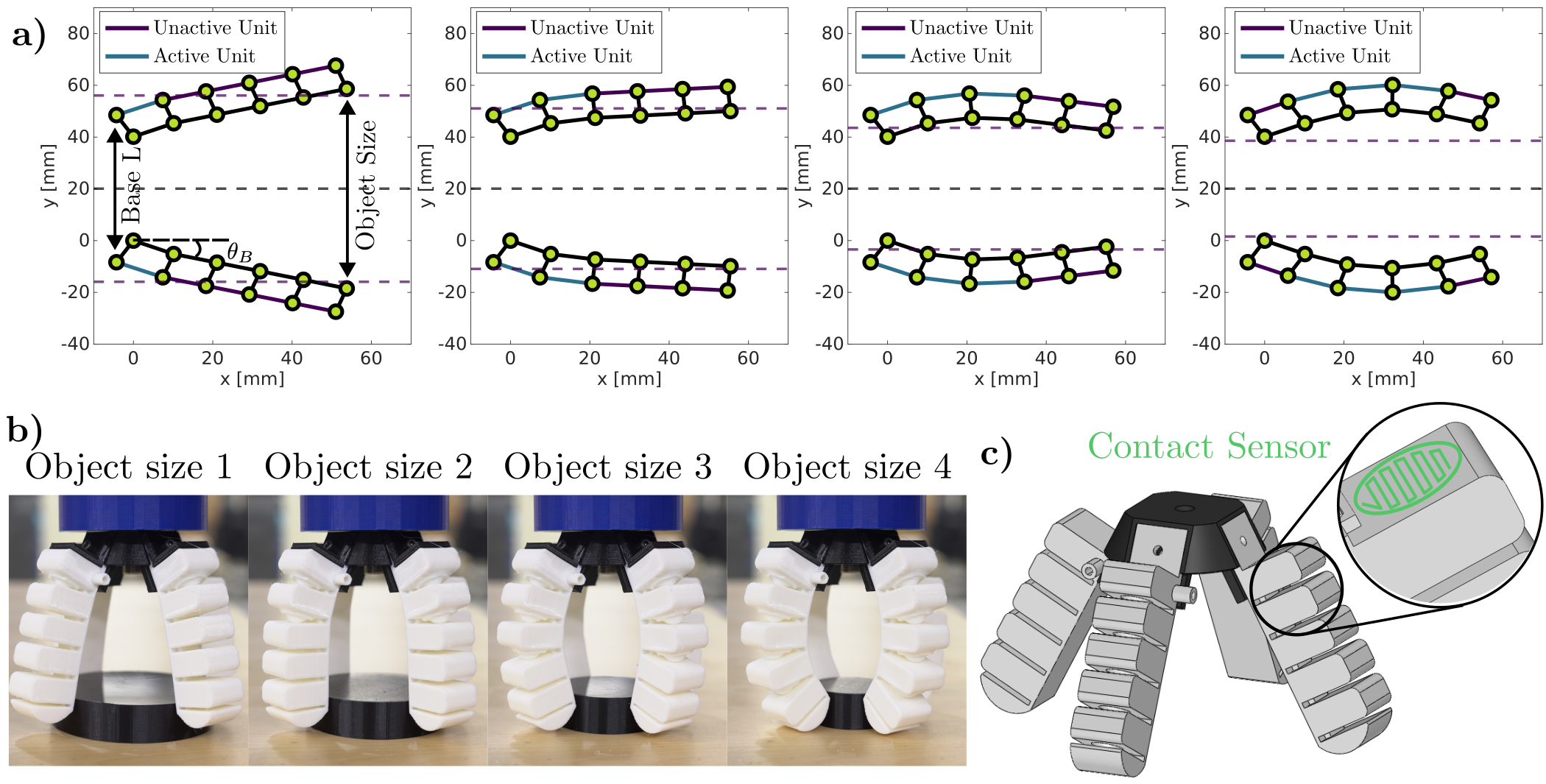}
  \caption{Dome Phalanx Gripper lattice design. a) Lattice model results for the four optimized stable states. b) 3D printed DPG on its four different programmed stable states. Black disks represent the designed object size. c) Contact sensor position placement to detect perturbations from the pre-programmed stable state. }
 \label{fig:si_multi_obj}
\end{figure}

We leverage the multistability of our DPF to develop a multistable gripper (DPG) capable of achieving multiple aperture configurations. This enables the robot to approximate and adapt to various object sizes. Representative object sizes, corresponding target positions, and optimization results are presented in Table~\ref{tab:object_size_targets}, while a visual comparison with the 3D-printed prototypes is shown in Figure~\ref{fig:si_multi_obj}b. Due to the gripper's symmetry, the aperture is modeled considering only two fingers. By targeting a specific set of aperture values, we encode object size information directly into the gripper's geometry and design. This allows for discrimination between objects with dimensions near the predefined targets. To decode this morphological information, we integrate a contact sensor into one of the finger units (Figure~\ref{fig:si_multi_obj}c), which detects when the gripper deviates from a stable state. We use a commercial RP-C10-ST pressure sensor (0.2–2~kg), although other sensors capable of detecting perturbations—such as curvature or strain sensors—can also be employed. The sensor threshold is calibrated by observing the sensor reading when the dome behind it is inverted, guaranteeing that all the stable configurations would not be detected as a contact. By utilizing the feedback from the contact sensor and the robot's morphology, we can create a classification loop (see Movie 7) that can classify between different object sizes and weights.

\begin{table}[h]\centering
\caption{Object size and target coordinate for multistable DPR}
\label{tab:object_size_targets}
\begin{tabular}{cc}
\hline 
Object size [mm]   & Model\\ \hline
35  & 35.02 \\
45  & 45.07 \\
60  & 60.023 \\
70  & 73.15 \\
\hline       
\end{tabular}
\end{table}

\newpage
Results for the optimization algorithm for different object sizes can be observed in Table~\ref{tab:classification_geo}.

\begin{table}[h]\centering
\caption{Object classification optimization results}
\label{tab:classification_geo}
\begin{tabular}{cc}
\hline 
H1           &        4.82\\
H2           &        4.52\\
H3           &        4.15\\
H4           &        3.18\\
H5           &        2.5\\
Unit S      &        1.82\\
Unit L      &        8.57\\
t      		&        0.83\\
$t_{\text{lim}}$  &      1.05        \\
UC           &      15.00  \\
Base L & 32.4\\
$\theta_B$ & 28$^\circ$\\
\hline       
\end{tabular}
\end{table}

\newpage
\subsection{Dome Phalanx Walker}\label{sec:DPW_Walker}
\begin{figure}[!h]
  \centering
  \includegraphics[width=\textwidth]{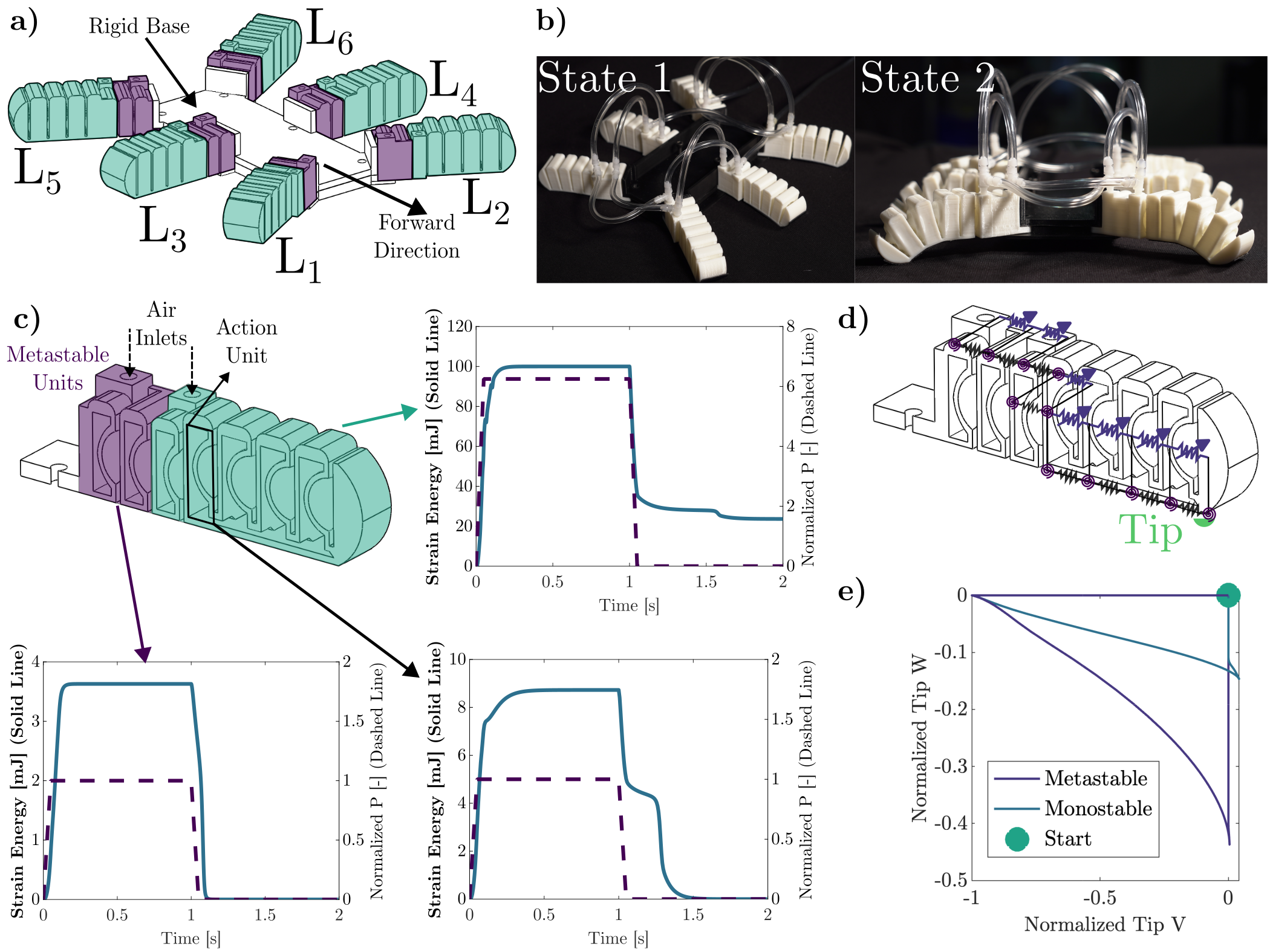}
  \caption{Dome Phalanx Walker design with lattice model. a) DPW architecture. b) Flat state of the walker (State 1) and arc-like state (State 2) by activating the bistable units. c) The DPW leg and its dynamic behavior are divided by zones and units. The metastable action unit exhibits a higher resetting time than the metastable units in the purple region.}
  \label{fig:si_walker_1}
\end{figure}

The dome phalanx walker (Figure~\ref{fig:si_walker_1}a) is designed by combining six legs, each composed of different arrangements of dome units (see Figure~\ref{fig:si_walker_1}a). The front and back legs are angled at 30$^\circ$, inspired by leg arrangements observed in insects~\cite{insect_walking_1999}, to enhance both stability and locomotion speed. Each leg has two functional zones (Purple and green zone in Figure~\ref{fig:si_walker_1}c), each composed of a combination of different dome unit geometries which can exhibit monostable, metastable, and bistable behaviors. Each leg exhibits two stable states (Figure~\ref{fig:si_walker_1}b), enabling the robot to support greater loads while remaining compatible with monolithic 3D printing in its initial flat configuration. To study the leg's motion and design the resetting times of the metastable units, we employ our lattice model to generate a discrete representation of the leg (see Figure~\ref{fig:si_walker_1}d). Our central modeling assumption is that units within different planes and zones do not interact, allowing for independent design of each zone while maintaining the model's simplicity. Using this approach, we determine both the metastable units' resetting times and each dome unit's design parameters (Figure~\ref{fig:si_walker_1}c) to program the timing of each locomotion phase. Additionally, the model allows us to track the tip displacement in the y-z plane over time (Figure~\ref{fig:si_walker_1}e), revealing how metastability amplifies leg movement through the time delays introduced by the viscoelastic response.

\begin{figure}[!h]
  \centering
  \includegraphics[width=\textwidth]{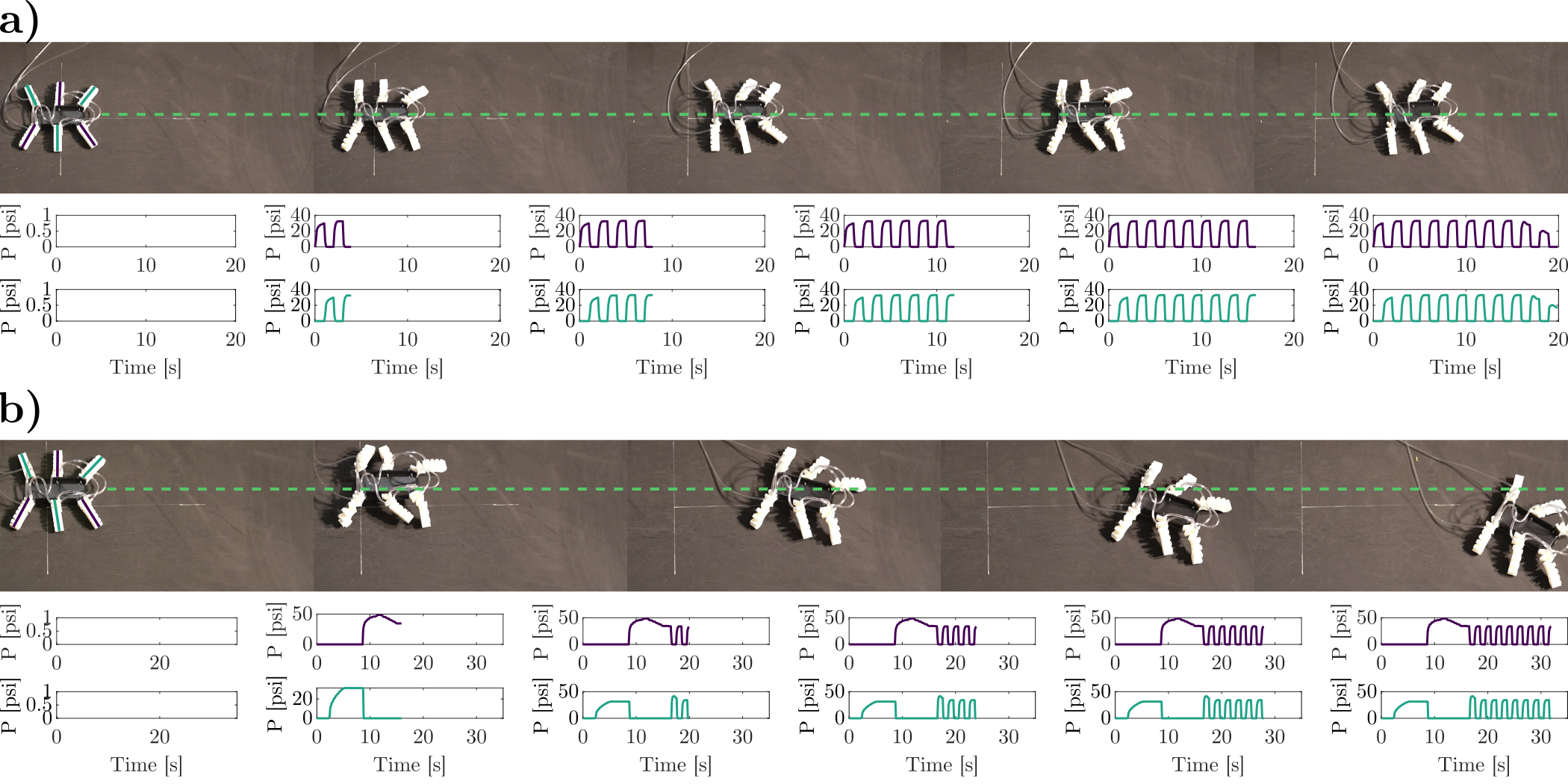}
  \caption{Dome Phalanx Walker programmed response and pressure measurements a) Snapshots of a straight walking direction given a uniform alternating pressure phase. b) Snapshots of the turning maneuver result from an asymmetry created by the pressure increase on the first phase cycle.}
  \label{fig:si_walker_2}
\end{figure}

\newpage
The design unit (see Figure~\ref{fig:si_walker_1}c) is tuned individually for each leg, enabling versatility and programmability in the walker's behavior. We employ this unit to achieve directional walking by controlling only the actuation pressure (see Figure~\ref{fig:si_walker_2}a and b). This is accomplished by introducing a design unit with metastable behavior in the front legs, characterized by a longer resetting time and a higher inversion pressure than the metastable unit used in zone 1 (purple zone).This configuration ensures that the front-leg units do not prematurely revert to their initial state, as actuation pressure is applied before their resetting time elapses. This creates a sustained asymmetry in leg behavior, driving directional movement until the pressure is released or increased on the opposite phase (see Figure~\ref{fig:si_walker_2}). It is important to note that the dome geometry directly affects the resetting time—longer resetting times are associated with higher inversion pressures. Accordingly, we designed the front-leg units for a longer resetting time, resulting in higher inversion pressure.

\end{document}